\documentclass[review, 3p]{elsarticle}
\usepackage{float}
\usepackage{threeparttable}
\usepackage{subcaption}
\usepackage[font=small,skip=0pt]{caption}
\usepackage[mediumspace,mediumqspace,Grey,squaren]{SIunits}
\usepackage{multirow}
\usepackage{inputenc}
\usepackage{lipsum}
\usepackage{booktabs}

\usepackage{lineno}
\usepackage{hyperref}
\hypersetup{
	colorlinks=true,
	linkcolor=black,
	citecolor=black,
	filecolor=black,
	urlcolor=black,
}
\usepackage{cleveref}
\modulolinenumbers[1]

\makeatletter
\usepackage{etoolbox}
\patchcmd\H@refstepcounter{\protected@edef}{\protected@xdef}{}{}
\makeatother

\journal{ISPRS Journal of Photogrammetry and Remote Sensing}

\biboptions{authoryear}

\begin{document}

\begin{frontmatter}

\title{30m resolution Global Annual Burned Area Mapping based on Landsat images and Google Earth Engine}

\author[fn1,fn2]{Tengfei~Long}
\author[fn1,fn2]{Zhaoming~Zhang\corref{cor1}}
\author[fn1,fn2]{Guojin~He\corref{cor1}}
\author[fn1,fn2]{Weili~Jiao}
\author[fn3]{Chao~Tang}
\author[fn1]{Bingfang~Wu}
\author[fn1,fn2]{Xiaomei~Zhang}
\author[fn1,fn2]{Guizhou~Wang}
\author[fn1]{Ranyu~Yin}

\cortext[cor1]{
	Corresponding author. Tel.: +86-01082178188.\newline
	E-mail address:~\href{mailto:hegj@radi.ac.cn}{hegj@radi.ac.cn} ~\href{mailto:zhangzm@radi.ac.cn}{zhangzm@radi.ac.cn}}
\address[fn1]{Institute of Remote Sensing and Digital Earth, Chinese Academy of Sciences, Beijing, China. 100094}
\address[fn2]{Hainan Key Laboratory for Earth Observation, Sanya, Hainan Province, China. 572029}
\address[fn3]{China University of Mining and Technology, Xuzhou, Jiangsu, China. 221116}

\begin{abstract}
Heretofore, global burned area (BA) products are only available at coarse spatial resolution, since most of the current global BA products are produced with the help of active fire detection or dense time-series change analysis, which requires very high temporal resolution. In this study, however, we focus on automated global burned area mapping approach based on Landsat images. By utilizing the huge catalog of satellite imagery as well as the high-performance computing capacity of Google Earth Engine, we proposed an automated pipeline for generating 30-meter resolution global-scale annual burned area map from time-series of Landsat images, and a novel 30-meter resolution global annual burned area map of 2015 (GABAM 2015) is released.	GABAM 2015 consists of spatial extent of fires that occurred during 2015 and not of fires that occurred in previous years. Cross-comparison with recent Fire\_cci version 5.0 BA product found a similar spatial distribution and a strong correlation ($R^2=0.74$) between the burned areas from the two products, although differences were found in specific land cover categories (particularly in agriculture land). Preliminary global validation showed the commission and omission error of GABAM 2015 are 13.17\% and 30.13\%, respectively.
\end{abstract}


\end{frontmatter}



\section{Introduction}
\par Accurate and complete data on fire locations and burned areas (BA) are important for a variety of applications including quantifying trends and patterns of fire occurrence and assessing the impacts of fires on a range of natural and social systems, e.g. simulating carbon emissions from biomass burning~\citep{Chuvieco_2016}. Remotely sensed satellite imagery has been widely used to generate burned area products. Burned area products at global scale using satellite images have been mostly based on coarse spatial resolution data such as Advanced Very High Resolution Radiometer (AVHRR), Geostationary Operational Environmental Satellite (GOES), VEGETATION or Moderate Resolution Imaging Spectroradiometer (MODIS) images. Main global burned area products include GBS ~\citep{Carmona_Moreno_2005}, Global Burned Area 2000 (GBA2000)~\citep{Tansey_2004}, GLOBSCAR~\citep{Simon_2004}, GlobCarbon~\citep{Plummer_2006}, L3JRC~\citep{Tansey_2008}, MCD45~\citep{Roy_2005}, GFED~\citep{Giglio_2010}, MCD64~\citep{Giglio_2016}, and Fire\_cci~\citep{Chuvieco_2016}. 

The recently released Fire\_cci product is produced based on MODIS images and has the highest spatial resolution (250m) of all the existing global burned area products~\citep{Chuvieco_2016,Pettinari2018}. However, the requisites of the climate modelling community are not yet met with the current global burned area products, as these products do not provide enough spatial detail~\citep{Bastarrika_2011}. Imagery collected by the family of Landsat sensors is useful and appropriate for monitoring the extent of area burned and provide spatial and temporal resolutions ideal for science and management applications. Landsat sensors can provide a longer temporal record (from 1970s until now) of burned area relative to existing global burned area products and potentially with increased accuracy and spatial detail in most areas on the earth~\citep{Stroppiana_2012}. Great importance has been attached to developing burned area products based on Landsat data in the past 10 years~\citep{Bastarrika_2011,Stroppiana_2012,Hawbaker_2017}. Up to now, there is no Landsat based global burned area product, however, some regional Landsat burned area products have been publicly released in recent years. Australia released its Fire Scars (AFS) products derived from all available Landsat 5, 7 and 8 images using time series change detection technique~\citep{Goodwin_2014}. Fire scars are automatically detected and mapped using dense time series of Landsat imagery acquired over the period 1987 – 2015 and the AFS product only covers the state of Queensland, Australia. Monitoring Trends in Burn Severity (MTBS) project, sponsored by the Wildland Fire Leadership Council (WFLC) provides consistent, 30-meter resolution burn severity data and fire perimeters across all lands of the United States from 1984-2015 (only fires larger than 200 ha in the eastern US and 400 ha in the western US are mapped)~\citep{Eidenshink_2007}. MTBS products are generated based on the difference of Normalized Burned Ratio (NBR) calculated from pre-fire and post-fire images, in which the burned area boundary is delineated by on-screen interpretation and the process of developing a categorical burn severity product is subjective and dependent on analyst interpretation. The Burned Area Essential Climate Variable (BAECV), developed by the U.S. Geological Survey (USGS), produced Landsat derived burned area products across the conterminous United States (CONUS) from 1984-2015, and its products have been released in April 2017 \citep{Hawbaker_2017}. The main differences between the MTBS and BAECV is the BAECV products are automatically generated based on all available Landsat images.

In summary, global burned area products are only available at coarse spatial resolution while 30-meter resolution burned area products are limited to specific regions. The majority of coarse spatial resolution algorithms developed to produce global burned area products use a multi-temporal change detection technique, because such satellite data have very high temporal resolution and are capable of monitoring fire-affected land cover changes. For example, the algorithm of MODIS burned area product (MCD45) is developed from the bi-directional reflectance model-based expectation change detection approach~\citep{Roy_2005}. One of the difficulties to produce Landsat based burned area products is that the traditional approaches successfully applied to extract global burned area from MODIS, VEGETATION, etc. don’t work well due to the limited temporal resolution of the Landsat sensors. Moreover, the analysis of post-fire reflectance may be easily contaminated by clouds or weakened by quick vegetation recovery, particularly in Tropical regions~\citep{Alonso_Canas_2015}. Another difficulty is that global 30-meter resolution annual burned areas mapping needs to utilize dense time-series Landsat images, and the required datasets can be hundreds of thousands of Landsat scenes, resulting impractical processing time. Although some researches have been addressed to detect burned area regionally from Landsat time series~\citep{Goodwin_2014,Hawbaker_2017_dense,Liu_2018}, results of global-scale have not been reported. However, thanks to Google Earth Engine (GEE), a new generation of cloud computing platforms with access to a huge catalog of satellite imagery and global-scale analysis capabilities~\citep{Gorelick_2017}, it is now possible to perform global-scale geospatial analysis efficiently without caring about pre-processing of satellite images. In this study, we focus on an automated approach to generate global-scale high resolution burned area map using dense time-series of Landsat images on GEE, and a novel 30-meter resolution global annual burned area map of 2015 (GABAM 2015) is released.

\section{Methodology}
\subsection{Sampling design}\label{subsec:sampling}
\par The spectral characteristics of burned areas vary in complex ways for different ecosystems, fire regimes and climatic conditions. In terms of guaranteeing the accuracy of global burned area map and also the completeness of quality assessment, a stratified random sampling method\citep{Padilla_2015, Boschetti_2016, Padilla_2017} was used to generate two sets of sites for classifier training and the validation of GABAM 2015, respectively. The training and validation sites were chosen randomly based on stratifications of both fire frequency and type of land cover.

\par Firstly, the Earth's Land Surface was partitioned based on the 14 land cover classes according to the MCD12C1 product~\citep{MCD12C1_2015} of 2012 using University of Maryland (UMD) scheme. These types were then merged into 8 categories based on their similarities~\citep{chuvieco2011esa}, i.e. Broadleaved Evergreen, Broadleaved Deciduous, Coniferous, Mixed forest, Shrub, Rangeland, Agriculture and Others. Table~\ref{tbl:categories} shows the reclassification rule from UMD land cover types to new classifications. As ``Others'' category consists of the biomes less prone to fire, only other 7 land cover categories are considered to create the geographic stratifications in this work.
\begin{table}[!ht]
	\caption{Map between original UMD land cover types and new classifications for the geographic stratification.}
	\label{tbl:categories}
	\centering
	\begin{tabular}{ll}
		\hline 
		New classification	&  Original UMD type\\ 
		\hline 
		Broadleaved Evergreen	&  Evergreen Broadleaf forest\\ 
		Broadleaved Deciduous &  Deciduous Broadleaf forest\\ 
		Coniferous &  Evergreen Needleleaf forest\\ 
		&  Deciduous Needleleaf forest\\ 
		Mixed Forest &  Mixed forest\\ 
		Shrub &  Closed shrublands\\ 
		&  Open shrublands\\ 
		Rangeland &  Woody savannas\\ 
		&  Savannas\\ 
		&  Grasslands\\
		Agriculture &  Croplands\\
		Others &  Water\\ 
		&  Urban and built-up\\ 
		&  Barren or sparsely vegetated\\
		\hline 
	\end{tabular} 
\end{table}

\par Secondly, the globe was divided into 5 partitions based on the BA density in 2015 provided by the Global Fire Emissions Database (GFED) version 4.0 ~\citep{Giglio_2013}, the most widely used inventory in global biogeochemical and atmospheric modeling studies~\citep{Giglio_2016}. Specifically, GFED4 monthly products of 2015 were utilized to produce an annual composition (GFED4 2015), consisting of 720 rows and 1440 columns which correspond to the global $0.25^\circ \times 0.25^\circ$ GFED grid, and each pixel summed the total areas of BA  (BA density, km$^2$) occurred in the grid cell during the whole year. The BA density of GFED4 2015 was then divided into 5 equal-frequency intervals~\citep{chuvieco2011esa} with Quantile classification.

\par
\begin{figure}[!ht]
	\centering
	\includegraphics[width=0.99\columnwidth]{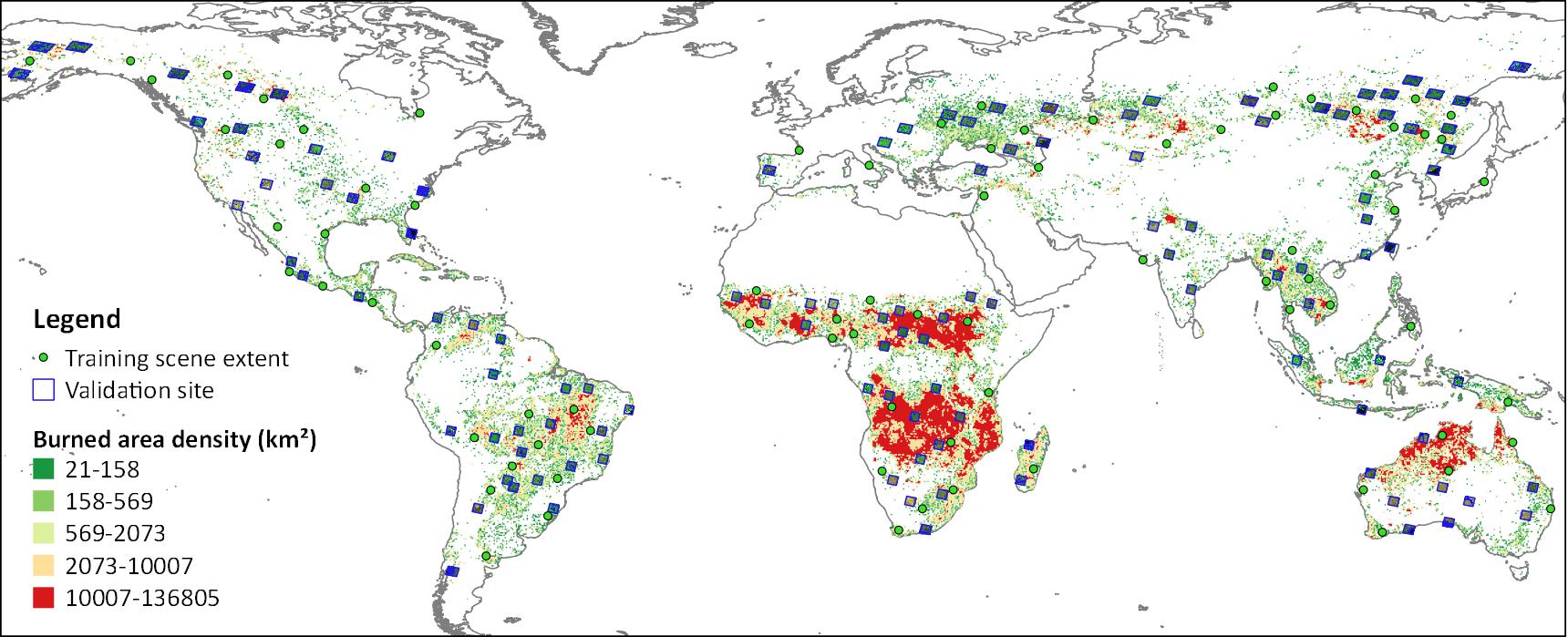}
	\caption[]{The distribution of 120 random Landsat image scenes and 80 validation sites over a map of BA density extracted from GFED4 2015.}
	\label{fig:scene_distribution}
\end{figure}

 By spatially intersecting the 7 land cover categories and 5 BA density levels, we obtained the final 35 strata with different fire frequencies and biomes. The samples were equally allocated to 5 BA density levels, but for different land cover categories, we also took into account the BA extent within each stratum with larger sample sizes allocated to strata with higher BA extent~\citep{Padilla_2014}. According to the strategy of stratified sampling, 120 samples (24 for each BA density level) were randomly selected to generate training dataset, and spatial dimension of sampling units was based on Landsat World Reference System II (WRS-II). Similarly, 80 validation sites (16 for each BA density level) were also created by randomly stratified sampling, but trying to keep a distance (at least 200km) from the training samples so as not to fall into the extent of training Landsat scenes. Figure~\ref{fig:scene_distribution} illustrates the distribution of 120 random Landsat image scenes and 80 validation sites over a map of BA density extracted from GFED4 2015, and Table~\ref{tbl:counter} shows the distribution of training and validation samples over the different land cover types.

\begin{table}[!ht]
	\caption{Distribution of training and validation samples over the different land cover types.}
	\label{tbl:counter}
	\centering
	\begin{tabular}{lcc}
		\hline 
		Land cover type	&  Training sample count & Validation sample count\\ 
		\hline 
		Broadleaved Evergreen	&  16	&  11\\ 
		Broadleaved Deciduous &  12	&  9\\ 
		Coniferous &  13	&  9\\ 
		Mixed Forest &  12	&  8\\ 
		Shrub &  18	&  12\\ 
		Rangeland &  25	&  15\\ 
		Agriculture &  24	&  16\\
		\hline 
	\end{tabular} 
\end{table}

\subsection{Training dataset}\label{subsec:training}
\par In terms of analyzing the characteristics of burned areas in Landsat images, 120 Landsat-8 image scenes were chosen according to the WRS-II frames generated by stratified random sampling in \cref{subsec:sampling}. All the Landsat-8 images used in this study were acquired from datasets of USGS Landsat-8 Surface Reflectance Tier 1 and Tier 2 in Google Earth Engine platform, whose ImageCollection IDs are ``LANDSAT/LC08/C01/T1\_SR'' and ``LANDSAT/LC08/C01/T2\_SR''. These data have been atmospherically corrected using LaSRC~\citep{Vermote_2016}, and include a cloud, shadow, water and snow mask produced using Fmask~\citep{Zhu_2014}, as well as a per-pixel saturation mask. For the purpose of burned area mapping, 6 bands of Landsat-8 image were used, i.e. three Visible bands (BLUE, 0.452-0.512~{\micro\metre}; GREEN, 0.533-0.590~{\micro\metre}; RED, 0.636-0.673~{\micro\metre}), Near Infrared band (NIR, 0.851-0.879~{\micro\metre}), and two Short Wave Infrared bands (SWIR1, 1.566-1.651~{\micro\metre}; SWIR2, 2.107-2.294~{\micro\metre}).
\par In this work, burned area mapping algorithm was implemented on GEE platform, and the maximum quantity of input samples are limited by GEE’s classifiers, thus average 90--100 sample points were collected by experienced experts from each Landsat-8 image, making the total quantity of sample points 12881 (6735 burned samples and 6146 unburned samples). Specifically, shortwave infrared (SWIR2), Near Infrared (NIR) and Green bands were composited in a Red, Green, Blue (RGB) combination in order to visualize burned areas better, and burned samples, including fire scars of different burn severity and of various biomass types, were extracted from the pixels showing magenta color~\citep{Koutsias_2000}. The unburned pixels were extracted randomly over the non fire affected areas covering vegetation, built-up land, bare land, topographic shadows, borders of lakes, etc. For those confusing pixels which were difficult to identify whether they were burned scars, further check was performed by examining the Landsat images on nearest date of the previous year or higher resolution images on nearest date in Google Earth software. To ensure only clearly burned pixels were selected, the burned samples were collected carefully to avoid pixels near the boundaries of burned scar~\citep{Bastarrika_2011}; and burned pixels located in burning flame or covered by smoke were also excluded to prevent potential contamination of burned samples. Land surface reflectance of the collected samples in BLUE, GREEN, RED, NIR, SWIR1, SWIR2 bands were extracted for further analysis. 

\subsection{Sensitive features for burned surfaces}\label{subsec:feature}
\begin{figure}[!ht]
	\centering
	\includegraphics[width=0.70\columnwidth]{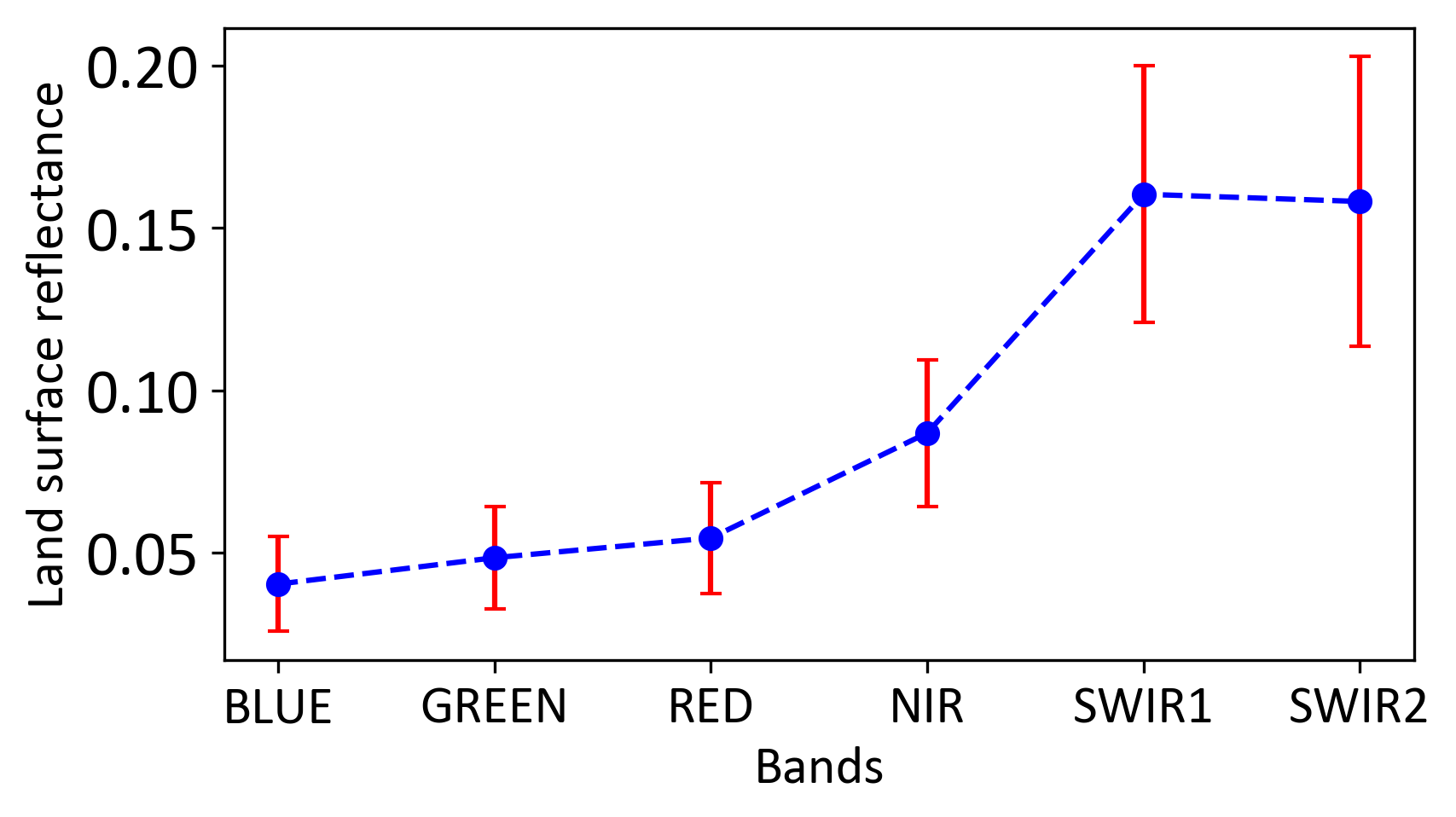}
	\caption[]{Means and standard deviations of land surface reflectance of burned Landsat-8 pixels in different bands.}
	\label{fig:reflectance}
\end{figure}

Figure~\ref{fig:reflectance} shows the statistical mean reflectance (with standard deviations) of burned samples in Landsat 8 bands.

\par
\begin{figure}[!ht]
	\centering
	\includegraphics[width=0.98\columnwidth, trim={0.8cm 0.1cm 1cm 0.8cm},clip]{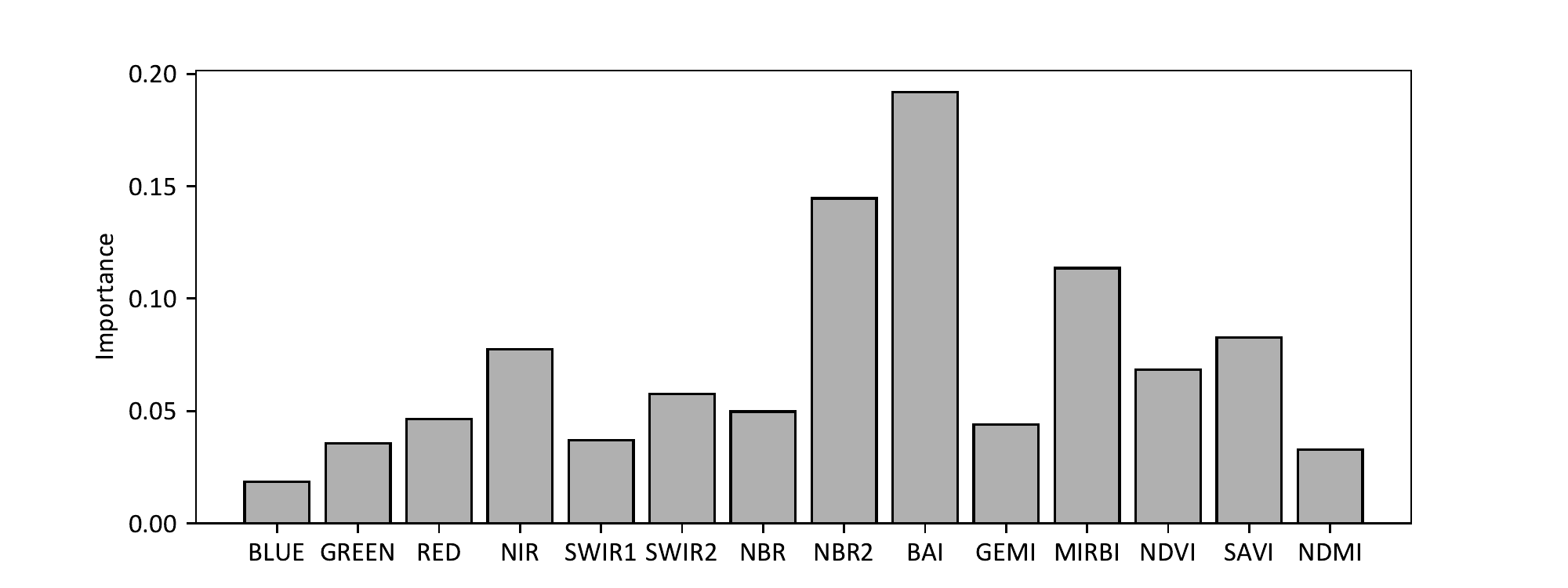}
	\caption[]{Relative importance of Landsat image features on burned areas classification evaluated by random forest algorithm.}
	\label{fig:importance}
\end{figure}

 Burned areas are characterized by deposits of charcoal, ash and fuel, and the reflectance of the burned pixels generally increases along with the wave length while the burned pixels have similar reflectance in SWIR1 and SWIR2 bands, which is greater than that in other bands. However, the spectral character of post-fire pixels varies greatly (standard deviations in Figure~\ref{fig:reflectance}) according to the type and condition of the vegetation prior to burning and the degree of combustion~\citep{Bastarrika_2014}, and none of existing spectral indices can be considered the best choice for identifying burned surfaces without misclassification with other targets in all environments or fire regimes~\citep{Boschetti_2010}. Consequently, in this work, we made use of most common spectral indices for Landsat image previously suggested in BA studies, and their formulas are summarized as Table~\ref{tbl:formula}. Some of these spectral indices were specifically developed for burn detection as they are sensitive to charcoal and ash deposition, such as normalized burned ratio (NBR~\citep{key1999normalized}), normalized burned ratio 2 (NBR2~\citep{lutes2006firemon}), burned area index (BAI~\citep{martin1998cartografia}), mid infrared burn index (MIRBI~\citep{Flasse_2001}). In addition, other indices that are not burn-specific may also be useful to map burned areas when cooperating with burn-specific indices. For instance, although normalized difference vegetation index (NDVI) is not the best index for burned area mapping, it is sensitive to vegetation greenness and therefore to the absence of vegetation in the case of burned areas~\citep{Stroppiana_2009}. The global environmental monitoring index (GEMI~\citep{Pinty_1992}) is an improved vegetation index, specifically designed to minimize problems of contamination of the vegetation signal by extraneous factors, and it is considered very important for the remote sensing of dark surfaces, such as recently burned areas~\citep{Pereira_1999}. The soil adjusted vegetation index (SAVI~\citep{Huete_1988}), which is originally designed for sparse vegetation and outperforms NDVI in environments with a single vegetation~\citep{Veraverbeke_2012}, is also helpful to improve separability of burns from soil and water~\citep{Stroppiana_2012}. The normalized difference moisture index (NDMI~\citep{Wilson_2002}), which is sensitive to the moisture levels in vegetation, is also relative to fuel levels in fire-prone areas. We also evaluated the relative importance of 14 Landsat features (8 spectral indices in Table~\ref{tbl:formula} and the surface reflectance in 6 bands of Landsat-8 image) when applied to classify burned areas using random forest algorithm~\citep{scikit-learn} (as shown in Figure~\ref{fig:importance}).

\begin{table}[!ht]
	\caption{The formulas of spectral indices that are sensitive to burned areas.}
	\label{tbl:formula}
	\centering
	\begin{threeparttable}
		\begin{tabular}{ll}
			\toprule
			\textbf{Name}	& \textbf{Formula}\\
			\midrule
			Normalized burned ratio		& $NBR=\frac{\rho_{NIR}-\rho_{SWIR2}}{\rho_{NIR}+\rho_{SWIR2}}$\\
			Normalized burned ratio 2		& $NBR2=\frac{\rho_{SWIR1}-\rho_{SWIR2}}{\rho_{SWIR1}+\rho_{SWIR2}}$\\
			Burned area index		& $BAI=\frac{1}{(\rho_{NIR}-0.06)^2+(\rho_{RED}-0.1)^2}$\\
			Mid infrared burn index		& $MIRBI=10\rho_{SWIR2}-0.98\rho_{SWIR1}+2$\\
			Normalized difference vegetation index		& $NDVI=\frac{\rho_{NIR}-\rho_{RED}}{\rho_{NIR}+\rho_{RED}}$\\
			Global environmental monitoring index		& $GEMI=\frac{\eta(1-0.25\eta)-(\rho_{RED}-0.125)}{1-\rho_{RED}}$,\\
			                                    		&$\eta=\frac{2(\rho_{NIR}^2-\rho_{RED}^2)+1.5\rho_{NIR}+0.5\rho_{RED}}{\rho_{NIR}+\rho_{RED}+0.5}$\\
			Soil adjusted vegetation index		& $SAVI=\frac{(1+L)(\rho_{NIR}-\rho_{RED})}{\rho_{NIR}+\rho_{RED}+L}$, $L=0.5$\\
			Normalized difference moisture index		& $NDMI=\frac{\rho_{NIR}-\rho_{SWIR1}}{\rho_{NIR}+\rho_{SWIR1}}$\\
			\bottomrule
		\end{tabular}
		\begin{tablenotes}[para,flushleft]
			$\rho_{RED}$ is the surface reflectance in RED, $\rho_{NIR}$ is the surface reflectance in NIR, $\rho_{SWIR1}$ is the surface reflectance in SWIR1 band, and $\rho_{SWIR2}$ is the surface reflectance in SWIR2 band.
		\end{tablenotes}
	\end{threeparttable}
\end{table}

\par Figure~\ref{fig:importance} shows that spectral indices were generally more important than the surface reflectance in most bands, except for NIR band and SWIR2 band, which are sensitive to removal of vegetation cover and deposits of char and ash~\citep{Pleniou_2013}. We also found that NBR2, BAI, MIRBI and SAVI had the greatest relative importance, whereas SAVI was not initially developed for burned area detection. However, considering the potential contribution of those features with relative low importance in distinguishing burned scars, all of 14 features were selected as sensitive features to perform global burned area mapping in this study.

\subsection{Burned area mapping via GEE} \label{subsec:gee}
\par
\begin{figure}[!ht]
	\centering
	\includegraphics[width=0.99\columnwidth, trim={0.6cm 0.6cm 0.6cm 0.6cm},clip]{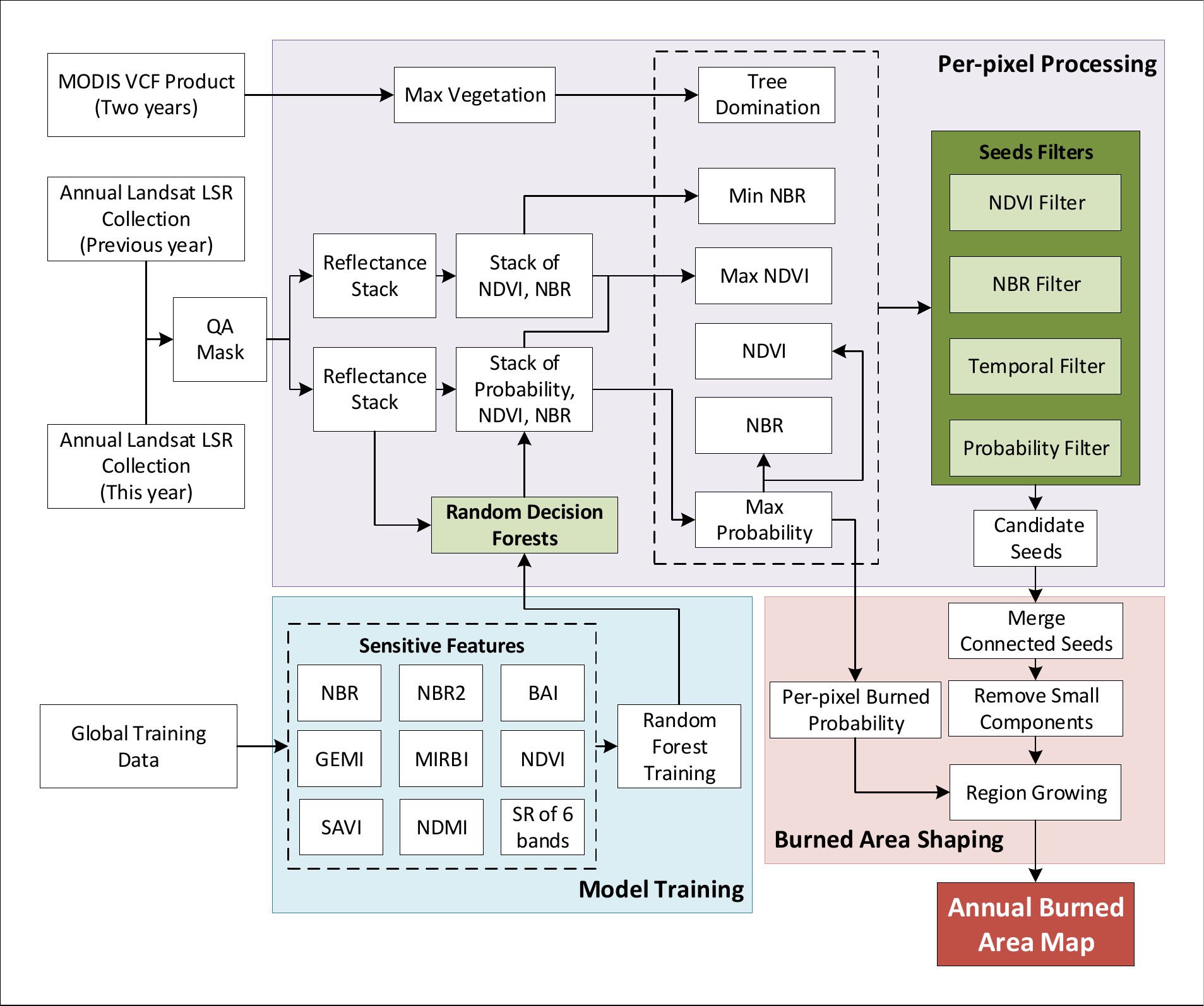}
	\caption[]{Workflow for annual burned area mapping using Google Earth Engine.}
	\label{fig:gee_workflow}
\end{figure}

 In this work, annual burned area map is defined as spatial extent of fires that occurs within a whole year and not of fires that occurred in previous years. Therefore, global 30-meter resolution annual burned areas mapping needs to utilize dense time-series Landsat images, and the pipeline of annual burned area mapping via GEE is described as Figure~\ref{fig:gee_workflow}.

\par As shown in Figure~\ref{fig:gee_workflow}, the pipeline mainly consists of three steps, model training, per-pixel processing and burned area shaping, and the following provides more details of each step.
\subsubsection{Model Training}\label{subsubsec:training}
\par The random forest (RF) algorithm provided by GEE were applied to train a decision forest classifier, and the global training data consisted of 6735 burned and 6146 unburned samples which were manually collected from 120 Landsat scenes generated by stratified random sampling (in \cref{subsec:sampling} and \cref{subsec:training}). Random forest classifier with higher number of decision trees usually provides better results, but also causes higher cost in computation time. Since the input features of the algorithm includes the surface reflectance (SR) in 6 bands of Landsat-8 image as well as 8 spectral indices that have high sensitivity to burned surface, we limited the number of decision trees in the forest to 100 for trade-off between accuracy and efficiency. Additionally, we chose ``probability'' mode for GEE's RF algorithm, in which the output is the probability that the classification is correct, and the probability would be further utilized to perform region growing in the step of burned area shaping.

\subsubsection{Per-pixel Processing}
\par In this step, Landsat surface reflectance collections from GEE, which consist of all the available Landsat scenes, were employed for dense time-series processing. At a pixel, the occurrence of a single Landsat satellite could be more than 20 or 40 times (considering the overlap between adjacent paths) within a year, and it would double when contemporary satellites (e.g. Landsat-7 and Landsat-8) were utilized. However, considering the failure of Scan Line Corrector (SLC) in the ETM+ instrument of Landsat-7 satellite, we only utilized USGS Landsat-8 Surface Reflectance collections (``LANDSAT/LC08/C01/T1\_SR'' and ``LANDSAT/LC08/C01/T2\_SR''). The quality assessment (QA) band of Landsat image, which was generated by FMask algorithm~\citep{Zhu_2014}, was used to perform QA masking. Pixels flagged as being clouds, cloud shadows, water, snow, ice, or filled/dropped pixels were excluded from Landsat scenes, and only clear land pixels remained after QA masking. At each pixel, the geometrically aligned dense time-series Landsat image scenes provided a reflectance stack of 6 bands, which was then split into two stacks by date filters, i.e. a stack of current year and that of the previous year. 
\par For the reflectance stack of current year, 8 spectral indices were computed at each time period, and then the trained decision forest classifier in \cref{subsubsec:training} produced a stack of burned probability using the 8 spectral indices and the reflectance of 6 bands. The maximum value of a probability stack indicates the probability that the pixel had ever appeared like burned scar during the whole year. Four quantities were noted for each pixel, i.e. the date on which the maximum probability was observed ($t_1$), as well as the burned probability ($p_{max}$), NDVI value ($NDVI_{1}$) and NBR value ($NBR_{1}$) on that date. However, it is not usually possible to unambiguously separate in a single image the spectral signature of burned areas from those caused by unrelated phenomena and disturbances such as shadows, flooding, snow melt, or agricultural harvesting~\citep{Boschetti_2015}; the burned scars which occurred in previous years but not yet recovered (particularly in boreal forests) should also be excluded from the annual BA map of current year. In this sense, we also concerned the summary statistics of current year and previous year: $NDVI_2$, the maximum NDVI value within the couple of years (current year and previous year); $t_2$, the date of $NDVI_2$; and $NBR_2$, the minimum NBR value within the previous year. Then most of the unreasonable tree-covered burned-like pixels would be excluded unless they met all the following constraints.

\begin{enumerate}
	\item $NDVI_2>T_{NDVI}$, the maximum NDVI value within the couple of years should be greater than a threshold~$T_{NDVI}$. We choose NDVI as it has been found to be a good identifier of vigorous vegetation, and this constraint is used to exclude areas that appear like burned but in fact were just lacking vegetation.
	\item $NDVI_2-NDVI_1>T_{dNDVI}$, the difference between the maximum NDVI and the NDVI when the pixel was most like burned scar should be greater than a threshold~$T_{dNDVI}$. This constraint ensures an evidence of vegetation decrease when burn happened.
	\item $NBR_2-NBR_1>T_{dNBR}$, the NBR value of a burned pixel should be less than the minimum NBR of the previous year, and the threshold $T_{dNBR}$ is the minimum acceptable decline of NBR. This constraint is useful to exclude false detections with periodic variation of NBR and NDVI, such as mountain shadows, burned-like soil in deciduous season, snow melting and flooding.
	\item $t_1>t_2$ or $t_2-t_1>T_{DAY}$, the most flourishing date of vegetation should be earlier than the burning date, or the lagged days should be less than a threshold~$T_{DAY}$. For tree-covered surface, it usually takes a long time for the vegetation to recover more flourishing than the previous year, thus the burn-like pixels with $t_1<=t_2$ are likely attributed to a false alarm. However, as the recovering of burned trees can be fast in tropic regions, high post-fire regrowth within a reasonable days is also acceptable.
\end{enumerate}

\par We named the first two constraints as ``NDVI filter'', the third and fourth ones as ``NBR filter'' and ``temporal filter'', respectively. In this work, the thresholds in above constraints were chosen empirically, $T_{NDVI}=0.2$, $T_{dNDVI}=0.2$, $T_{DAY}=100$(days) and $T_{dNBR}=0.1$. Determining a globally optimal NDVI threshold is not easy or even impossible for various types and conditions of the vegetation, and we chose a low threshold $T_{NDVI}=0.2$~\citep{Sobrino_2000}, not expecting to directly exclude all confusing surfaces never covered by vegetation. Actually, the second constraint would also help to exclude non-vegetation with high NDVI, because the decline of NDVI, in the absence of vegetation variation, commonly wouldn't meet the constraint. The change of NBR in pre-fire and post-fire images, defined as delta NBR or dNBR, has proved to be a good indicator of burn severity and vegetation regrowth (higher the severity, greater the dNBR)~\citep{Miller_2007,Lhermitte_2011}. It was suggested dNBR greater than 0.1 commonly indicates burn of low severity~\citep{lutes2006firemon}, thus we chose $T_{dNBR}=0.1$. Lastly, in the temporal filter, a fixed recovery cycle for all kinds of trees is also not available, and we just approximately chose an average time, 100 days. 

However, for herbaceous vegetation, we should use only the first two constraints, as grassland usually recovers very quickly and can be burned year after year. Accordingly, annual MODIS Vegetation Continuous Fields (VCF) 250 m Collection 5.1 (MOD44B) product~\citep{NASA_2015} of current and previous year, which contains the Tree-Cover Percent layer and Non-Tree Vegetation layer, were utilized to determine whether the pixel is dominated by tree or by herbaceous vegetation. Passing the filters of NDVI, NBR and temporal context, those pixels with annual burned probability greater than or equal to 0.95 (``probability filter'') were selected as seeds for region growing.

\subsubsection{Burned Area Shaping}
\par
\begin{figure}[!ht]%
	\centering
	\begin{subfigure}[t]{0.495\columnwidth}
		\centering
		\includegraphics[width=\textwidth]{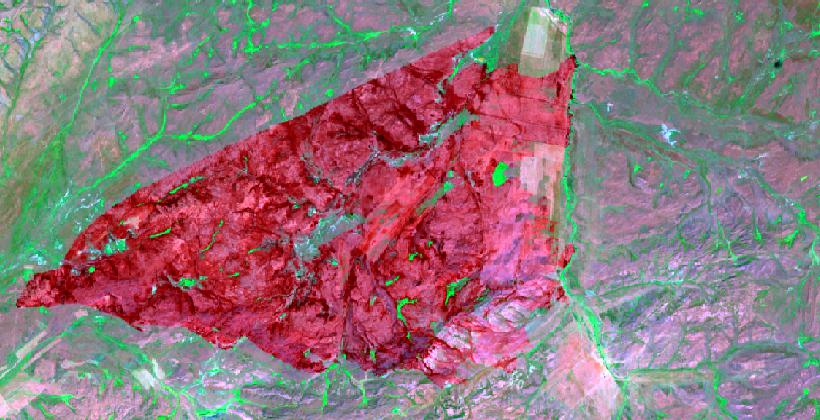}%
		\caption{}
		\label{fig:region_growing_image}%
	\end{subfigure}
	\begin{subfigure}[t]{0.495\columnwidth}
		\centering
		\includegraphics[width=\textwidth]{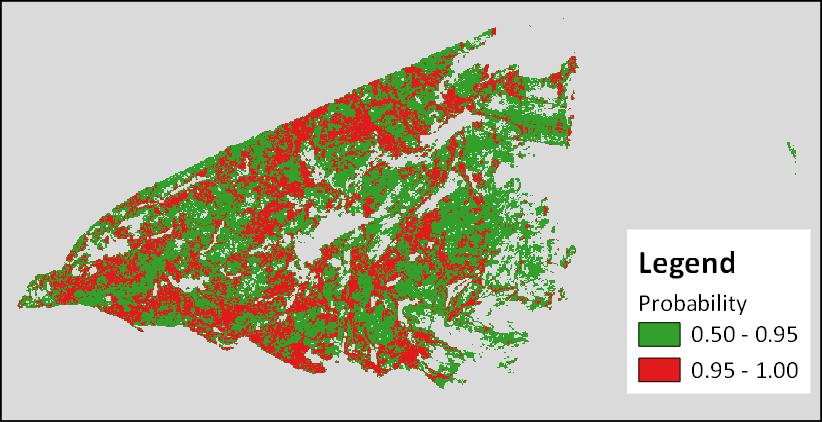}%
		\caption{}
		\label{fig:region_growing_probability}%
	\end{subfigure}
	\begin{subfigure}[t]{0.495\columnwidth}
		\centering
		\includegraphics[width=\textwidth]{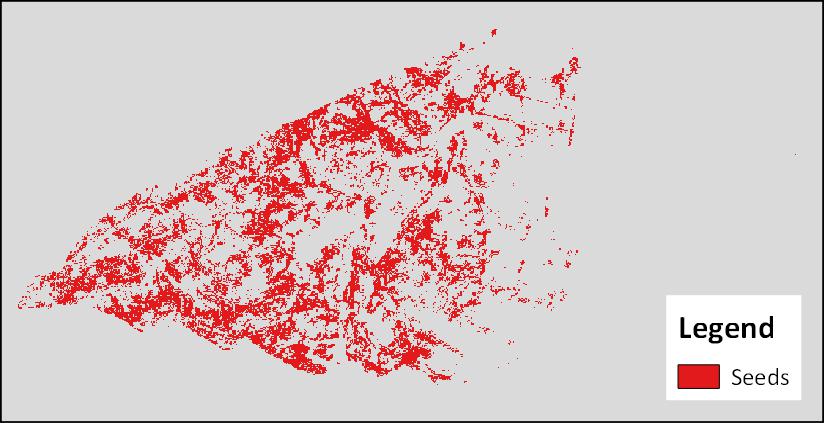}%
		\caption{}
		\label{fig:region_growing_seeds}%
	\end{subfigure}
	\begin{subfigure}[t]{0.495\columnwidth}
		\centering
		\includegraphics[width=\textwidth]{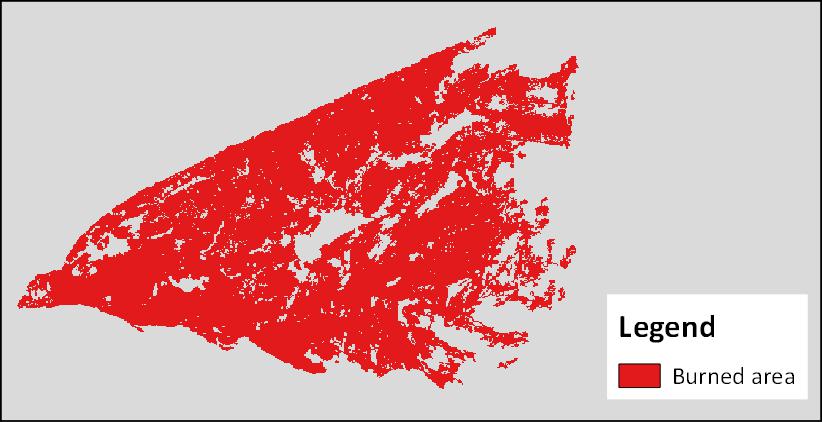}%
		\caption{}
		\label{fig:region_growing_burned}%
	\end{subfigure}%
	\caption[]{
		Example of region growing for burned area detection. ~\ref{fig:region_growing_image} is the Landsat-8 image displayed in false color composition (red: SWIR2 band, green: NIR band and blue: GREEN band), \ref{fig:region_growing_probability} is the map of burned probability generated by proposed method, \ref{fig:region_growing_seeds} is the candidate seeds of burned area, \ref{fig:region_growing_burned} shows the final burned area map after region growing.
	}\label{fig:region_growing}%
\end{figure}

 In this step, a region growing process was employed to shape the burned areas. Region growing has proved to be necessary for BA mapping in many studies~\citep{Bastarrika_2011, Stroppiana_2012, Hawbaker_2017_dense}, because spectral based methods sometimes give ambiguous evidence (i.e. spectral overlapping between burned areas and unrelated phenomena with similar spectral characteristics, such as cloud shadows, ephemeral water or dark soils~\citep{Stroppiana_2012}), and accepting all positive evidence can lead to confusion errors. Although candidate seeds were chosen with high confidence, false seed pixels were still frequently included in confusing surfaces, e.g. shadows, borders of lakes. Different from the candidate seeds in the actual burned scars, those falsely introduced seed pixels always distributed sparsely. Consequently, we aggregated the seed pixels into connected components using a kernel of 8-connected neighbors, and by ignoring small fires with areas less than 1 ha~\citep{Laris_2005}, those fragmentary components (smaller than 11 pixels), which included most false seed pixels, were removed. Finally, an iterative procedure of region growing were performed around each seed pixel. For each iteration, the 8-connected neighbors of the seed pixels were aggregated as burned pixels (new seeds) if their burned probabilities were greater than or equal to 0.5, and the iteration stopped when no more pixels can be aggregated as burned pixels. Figure~\ref{fig:region_growing} shows an example of region growing. One can see that only some pixels showing strong magenta color in the burned scars were chosen as seeds while those showing light magenta color were labeled as candidates for region growing, including some actual burned pixels as well as some false detections (right-middle in Figure~\ref{fig:region_growing_probability}). However, after the processes of small seeds removal and region growing, the false detections were excluded while those candidates near the seeds were aggregated to the final BA map.

\section{Results and analysis}
\subsection{Product description}

\par Employing the proposed approach, we produced the global annual burned area map of 2015 (GABAM 2015), which was projected in a Geographic (Lat/Long) projection at $0.00025^\circ$ (approximately 30 meters) resolution, with the WGS84 horizontal datum and the EGM96 vertical datum. The result consists of 10x10 degree tiles spanning the range 180W–180E and 80N–60S and can be freely downloaded from https://vapd.gitlab.io/post/gabam2015/. To make visualization of GABAM better, burned area density is used instead of directly drawing the burned pixels on a global map, and it is defined as the proportion of burned pixels in a $0.25^{\circ}\times0.25^{\circ}$ grid. An overview of global distribution of burned area density, derived from the 1 arc-second resolution GABAM 2015, is shown in Figure~\ref{fig:grid-gabam}, together with that of the Fire\_cci product in \cref{subsec:comparison}.

\begin{figure}[!ht]%
	\centering
	\begin{subfigure}[t]{0.32\columnwidth}
		\centering
		\includegraphics[width=\textwidth]{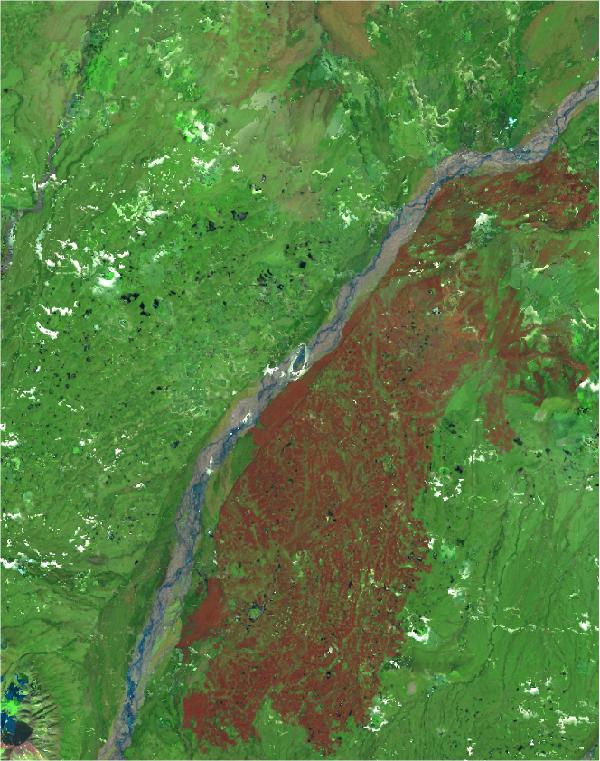}%
		\caption{2014}
		\label{fig:Canada_2014}%
	\end{subfigure}\hspace{2pt}
	\begin{subfigure}[t]{0.32\columnwidth}
		\centering
		\includegraphics[width=\textwidth]{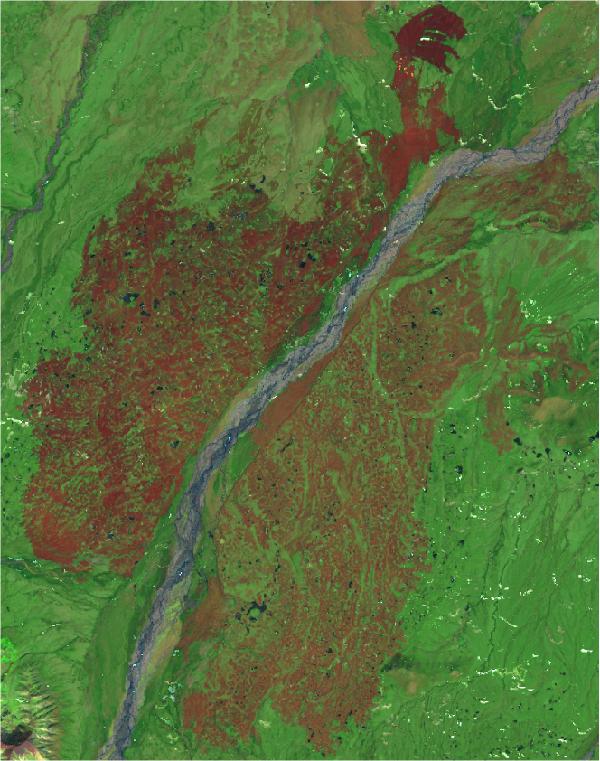}%
		\caption{2015}
		\label{fig:Canada_2015}%
	\end{subfigure}\hspace{2pt}
	\begin{subfigure}[t]{0.32\columnwidth}
		\centering
		\includegraphics[width=\textwidth]{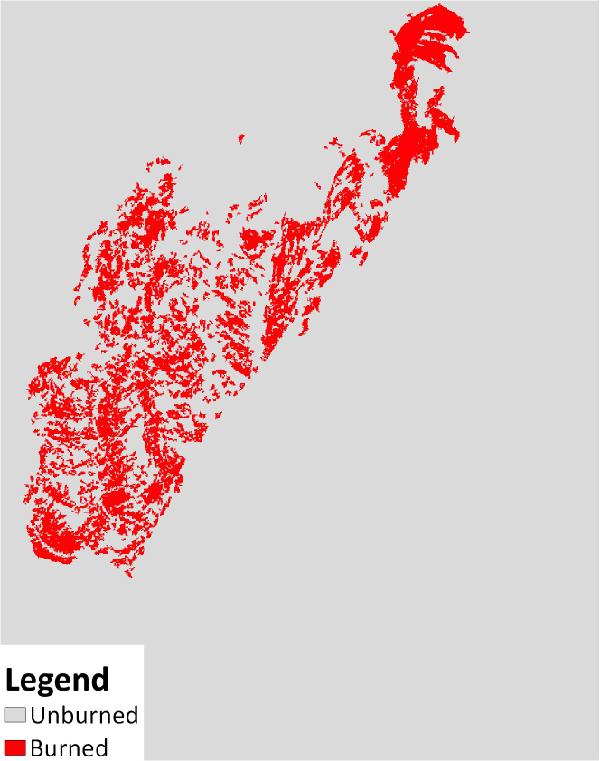}%
		\caption{BA}
		\label{fig:Canada_burned}%
	\end{subfigure}
	\caption[]{
		Burned area map example in Canada.  \ref{fig:Canada_2014} is the annually composited Landsat images of 2014 with the minimum NBR values; \ref{fig:Canada_2015} is the annually composited Landsat images of 2015; \ref{fig:Canada_burned} shows the detected burned scars occurred in 2015.
	}\label{fig:example2}%
\end{figure}

Figure~\ref{fig:example2} illustrates an examples of GABAM 2015 in Canada, and the annually composited Landsat reference images with minimum NBR values of 2015 and 2014 are also included. This region is located in high latitude zones, and the burned scars may not completely recover within a year. Consequently, when new burning occurs around the unrecovered burned scars, we must determine which burned scars come from this year. Owing to the temporal filters, GABAM succeeded to clear up such confusion. From Figure~\ref{fig:Canada_2015}, one can see that the burned scars mainly consist of two components, separated by the river. Figure~\ref{fig:Canada_2014}, however, shows that burned scars on the right side of the river can be observed in 2014, hence the result of GABAM 2015 only remains the component on the left side.

\subsection{Comparison with Fire\_cci product} \label{subsec:comparison}
\subsubsection{Data preparing}
\par As 30m resolution global burned area products are currently not available, we made a comparison between GABAM 2015 and the Fire\_cci version 5.0 products (spatial resolution is approximately 250 meters)~\citep{Pettinari2018}, which are based on MODIS on board the Terra satellite. 
The monthly Fire\_cci pixel BA products of 2015 were composited as an annual pixel BA product by labeling the pixels as burned ones once their values in Julian Day (the Date of the first detection) layer were valid (from 1 to 366) in any of the 12 monthly products. Additionally, in order to perform regression analysis between two products of different spatial resolution, we also produced an annual grid composition of BA within 2015 from the composited annual pixel BA product by computing the proportion of burned pixels in each $0.25^\circ \times 0.25^\circ$ grid. Note that the monthly grid BA products of Fire\_cci were not used to composite the annual grid product, because summing up the areas of BA for each grid in all monthly products might result in repetitive counting at those pixels burned more than once within the year. 
\subsubsection{Visually comparing}
\par
\begin{figure}[!ht]%
	\centering
	\begin{subfigure}[t]{0.40\columnwidth}
		\centering
		\includegraphics[width=\textwidth]{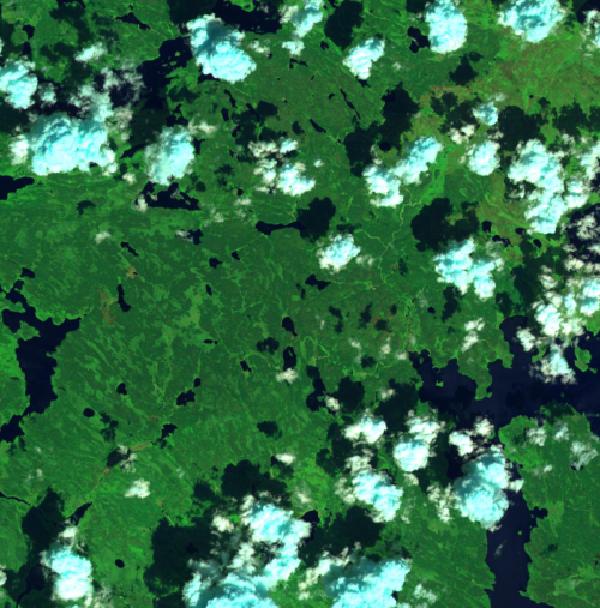}%
		\caption{}
		\label{fig:fcci-compare-landsat-0624}%
	\end{subfigure}
	\begin{subfigure}[t]{0.40\columnwidth}
		\centering
		\includegraphics[width=\textwidth]{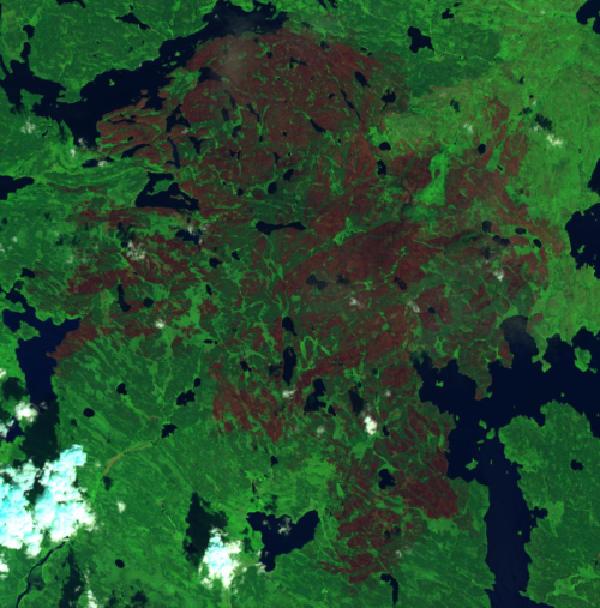}%
		\caption{}
		\label{fig:fcci-compare-landsat-0726}%
	\end{subfigure}
	\begin{subfigure}[t]{0.40\columnwidth}
		\centering
		\includegraphics[width=\textwidth]{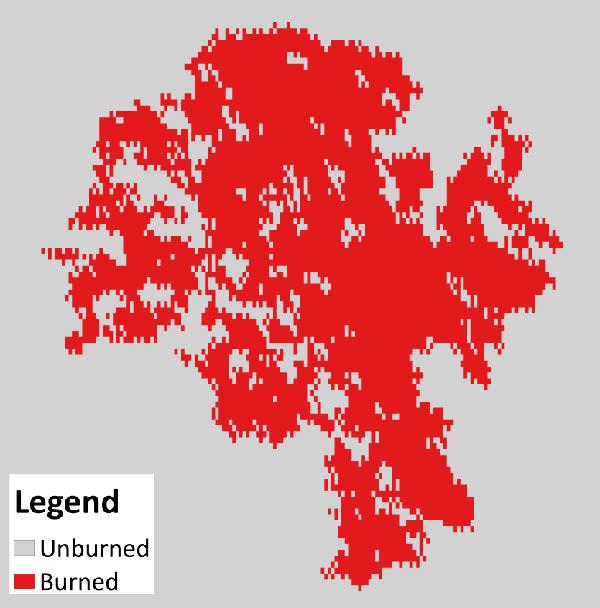}%
		\caption{}
		\label{fig:fcci-compare-fcci}%
	\end{subfigure}
	\begin{subfigure}[t]{0.40\columnwidth}
		\centering
		\includegraphics[width=\textwidth]{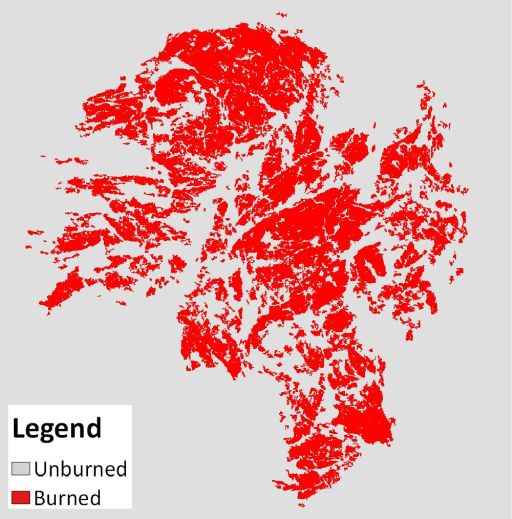}%
		\caption{}
		\label{fig:fcci-compare-gabam}%
	\end{subfigure}
	\caption[]{
		Comparison between Fire\_cci and GABAM. ~\ref{fig:fcci-compare-landsat-0624} and ~\ref{fig:fcci-compare-landsat-0726} are the Landsat-8 images before (June 24th, 2015) and after (July 26th, 2015) fire, respectively, displayed in false color composition (red: SWIR2 band, green: NIR band and blue: GREEN band); \ref{fig:fcci-compare-fcci} shows the burned areas of annually composited Fire\_cci product, and \ref{fig:fcci-compare-gabam} shows the burned areas generated by proposed method.
	}\label{fig:fcci-compare}%
\end{figure}

 Figure~\ref{fig:fcci-compare} shows an example of the two annual pixel BA products, and it can be seen that both products correctly detected the BAs in Landsat image (Figure~\ref{fig:fcci-compare-landsat-0726}), yet the BAs in Figure~\ref{fig:fcci-compare-fcci} occupy more pixels than those in Figure~\ref{fig:fcci-compare-gabam}. Due to the limitation in spatial resolution of the input sensor of the Fire\_cci BA product, some of the mixed pixels (consisting of burned and unburned pixels) may be classified as burned ones. On the other hand, the result of GABAM 2015 shows finer boundaries of BAs, compared with that of the Fire\_cci product.

\subsubsection{Global grid map}
\par
\begin{figure}[!ht]
	\centering
	\begin{subfigure}[t]{0.99\columnwidth}
		\centering
		\includegraphics[width=\textwidth]{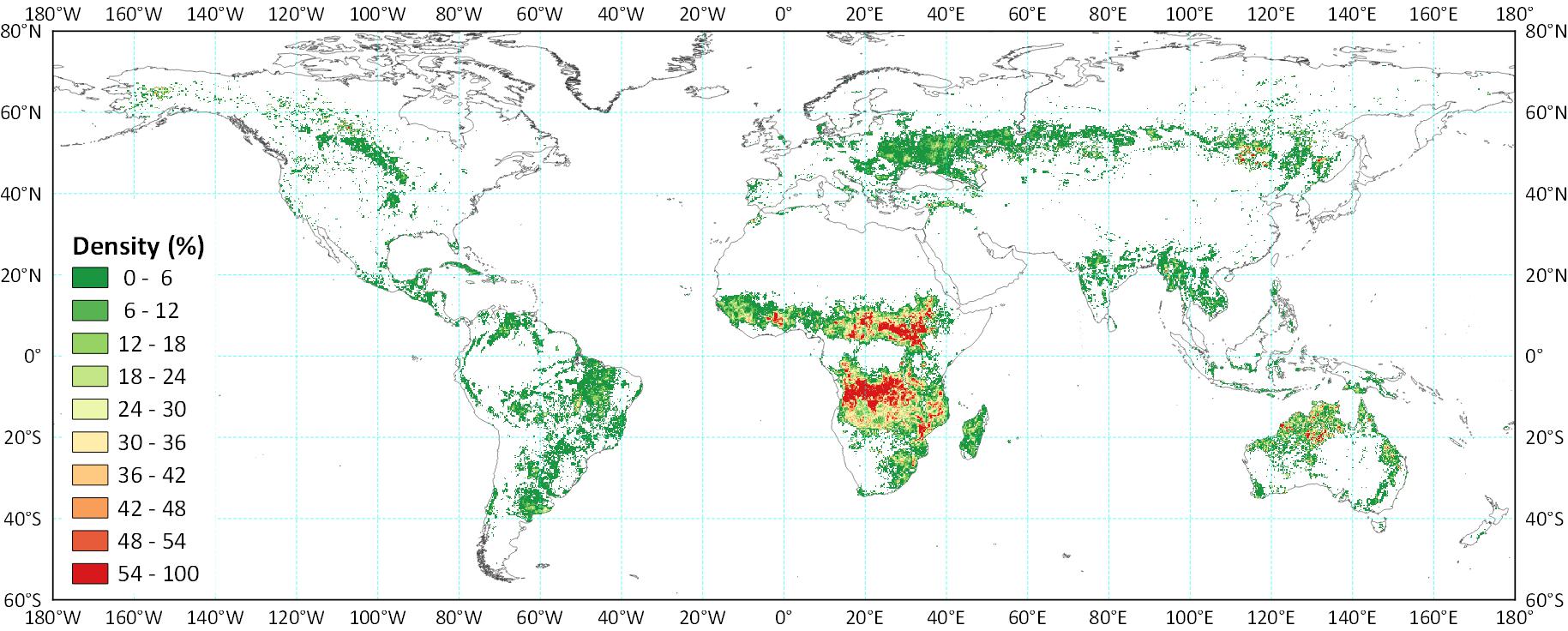}
		\caption{GABAM 2015}
		\label{fig:grid-gabam}%
	\end{subfigure}
	\begin{subfigure}[t]{0.99\columnwidth}
		\centering
		\includegraphics[width=\textwidth]{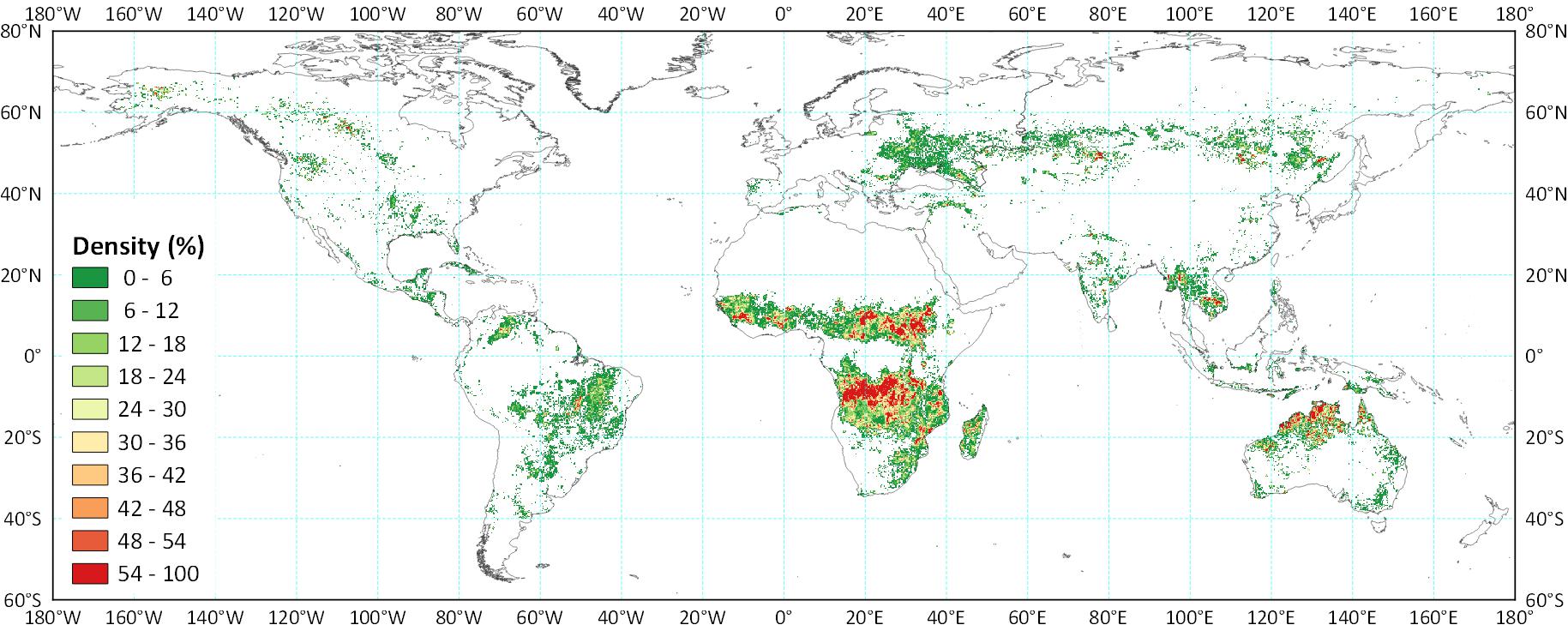}
		\caption{Fire\_cci 2015}
		\label{fig:grid-fcci}%
	\end{subfigure}
	\caption[]{Global distribution of burned area density (percentage of burned pixels in every $0.25^{\circ}\times0.25^{\circ}$ grid) of GABAM and Fire\_cci product within 2015. ~\ref{fig:grid-gabam} is the annual grid composition of BA of GABAM, and~\ref{fig:grid-fcci} is that of the Fire\_cci product.} 
	\label{fig:grid}
\end{figure}

 Figure~\ref{fig:grid} illustrates the GABAM and Fire\_cci annual grid composition of BA, consisting of percentage of burned pixels in each $0.25^{\circ}\times0.25^{\circ}$ grid. Figure~\ref{fig:grid-fcci} and Figure~\ref{fig:grid-gabam} show similar global distributions of BA density. 

\subsubsection{Regression analysis}
\par
\begin{figure}[!ht]%
	\centering
	\begin{subfigure}[t]{0.32\columnwidth}
		\includegraphics[width=\textwidth]{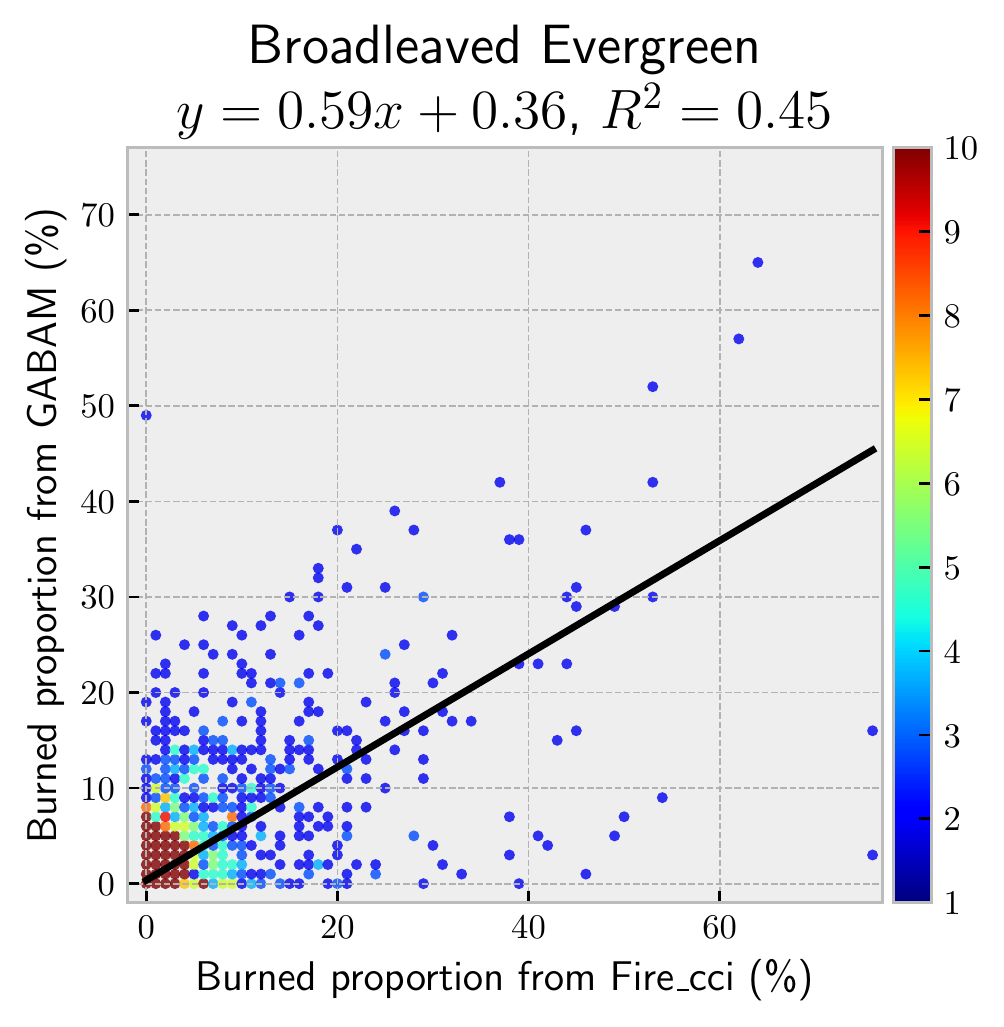}%
		\caption{}
		\label{fig:fcci-zone1}%
	\end{subfigure}%
	\hspace{1pt}
	\begin{subfigure}[t]{0.32\columnwidth}
		\includegraphics[width=\textwidth]{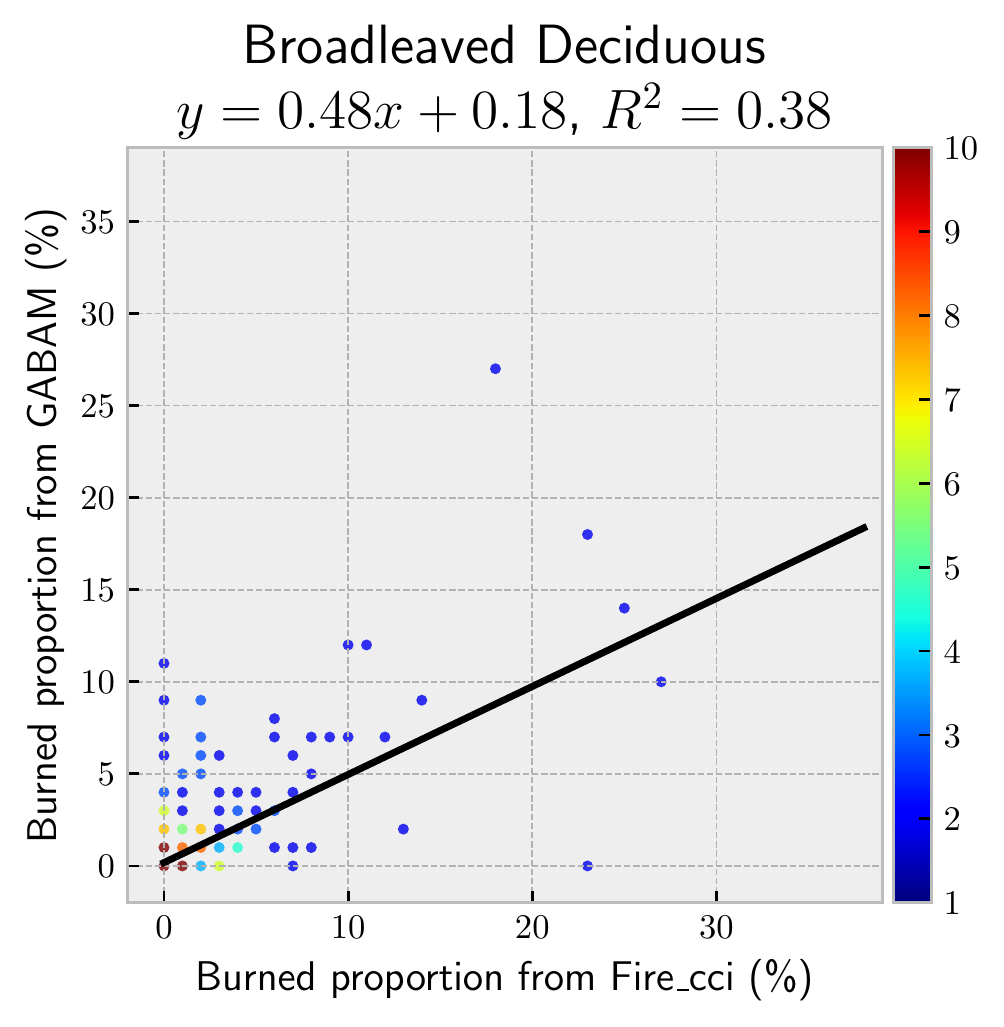}%
		\caption{}
		\label{fig:fcci-zone2}%
	\end{subfigure}
	\begin{subfigure}[t]{0.32\columnwidth}
		\includegraphics[width=\textwidth]{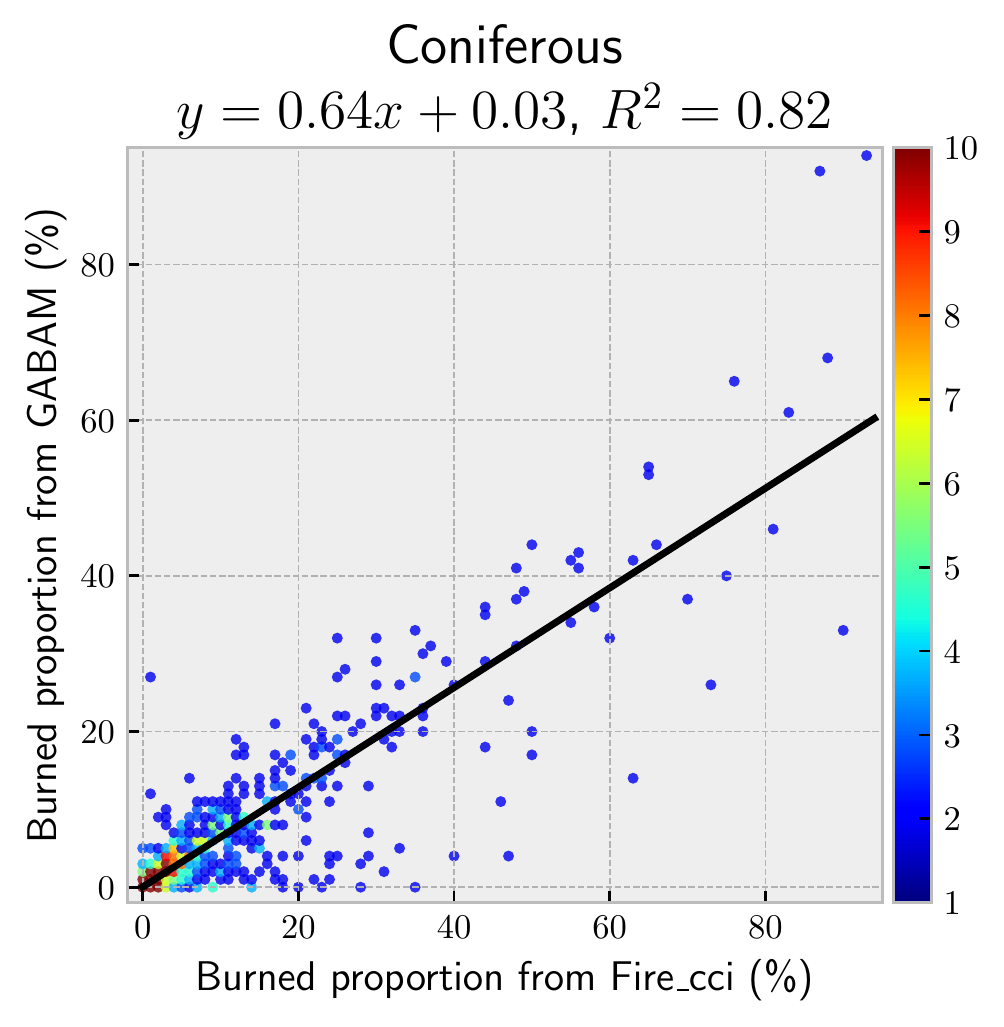}%
		\caption{}
		\label{fig:fcci-zone3}%
	\end{subfigure}
	\begin{subfigure}[t]{0.32\columnwidth}
		\includegraphics[width=\textwidth]{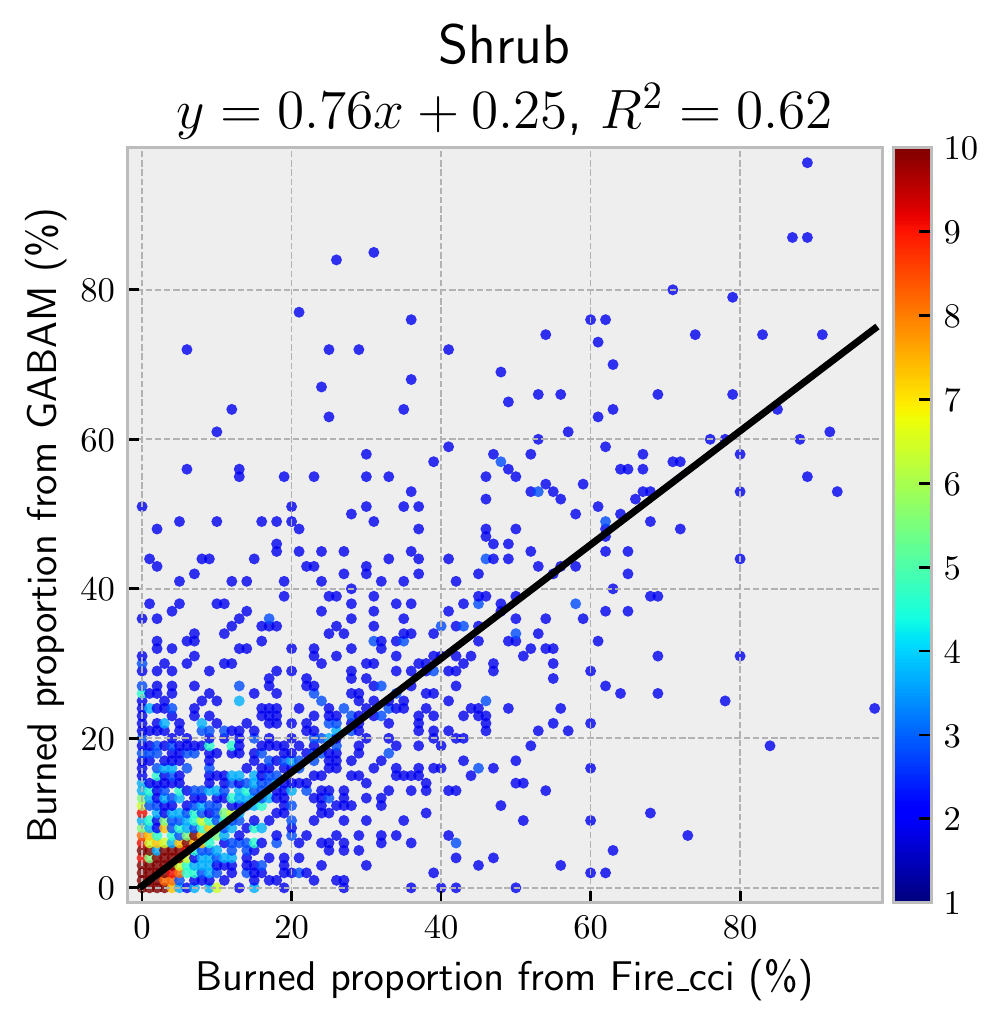}%
		\caption{}
		\label{fig:fcci-zone4}%
	\end{subfigure}
	\begin{subfigure}[t]{0.32\columnwidth}
		\includegraphics[width=\textwidth]{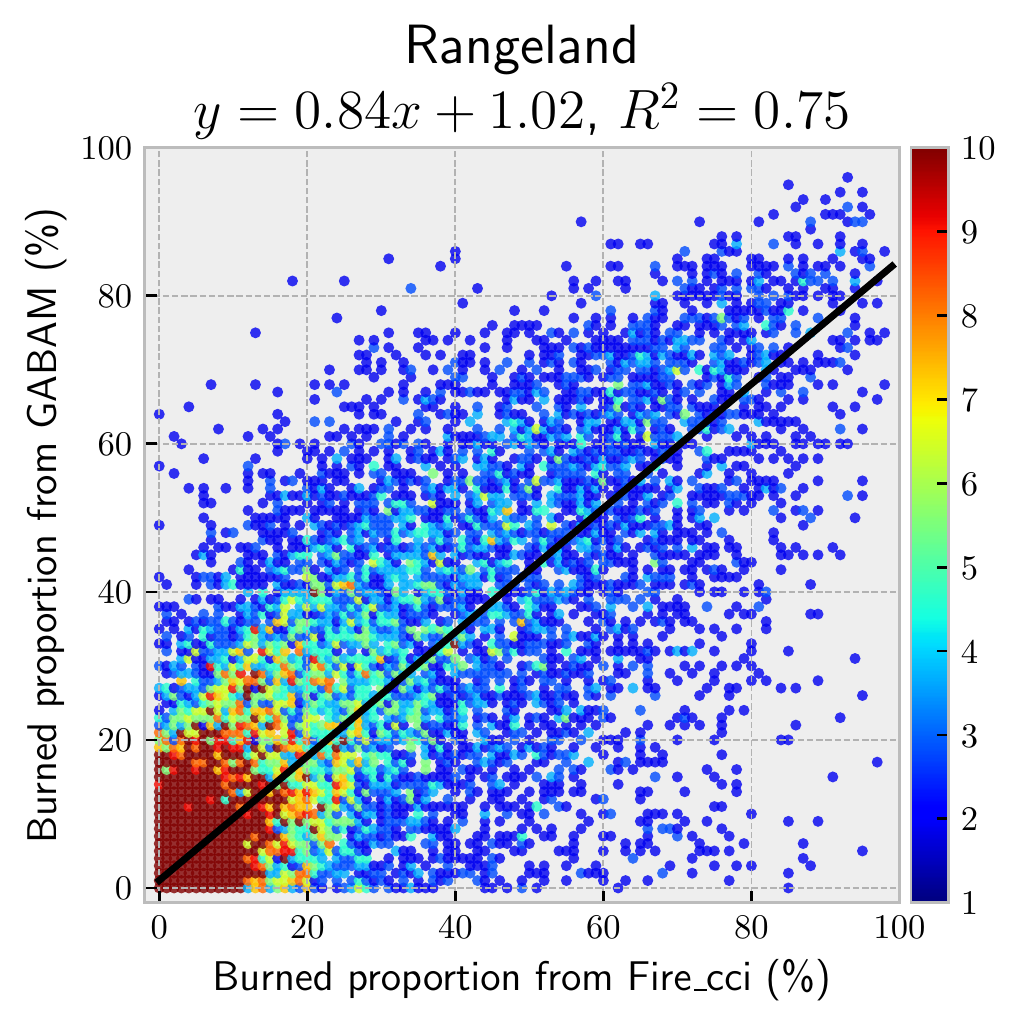}%
		\caption{}
		\label{fig:fcci-zone5}%
	\end{subfigure}
	\begin{subfigure}[t]{0.32\columnwidth}
		\includegraphics[width=\textwidth]{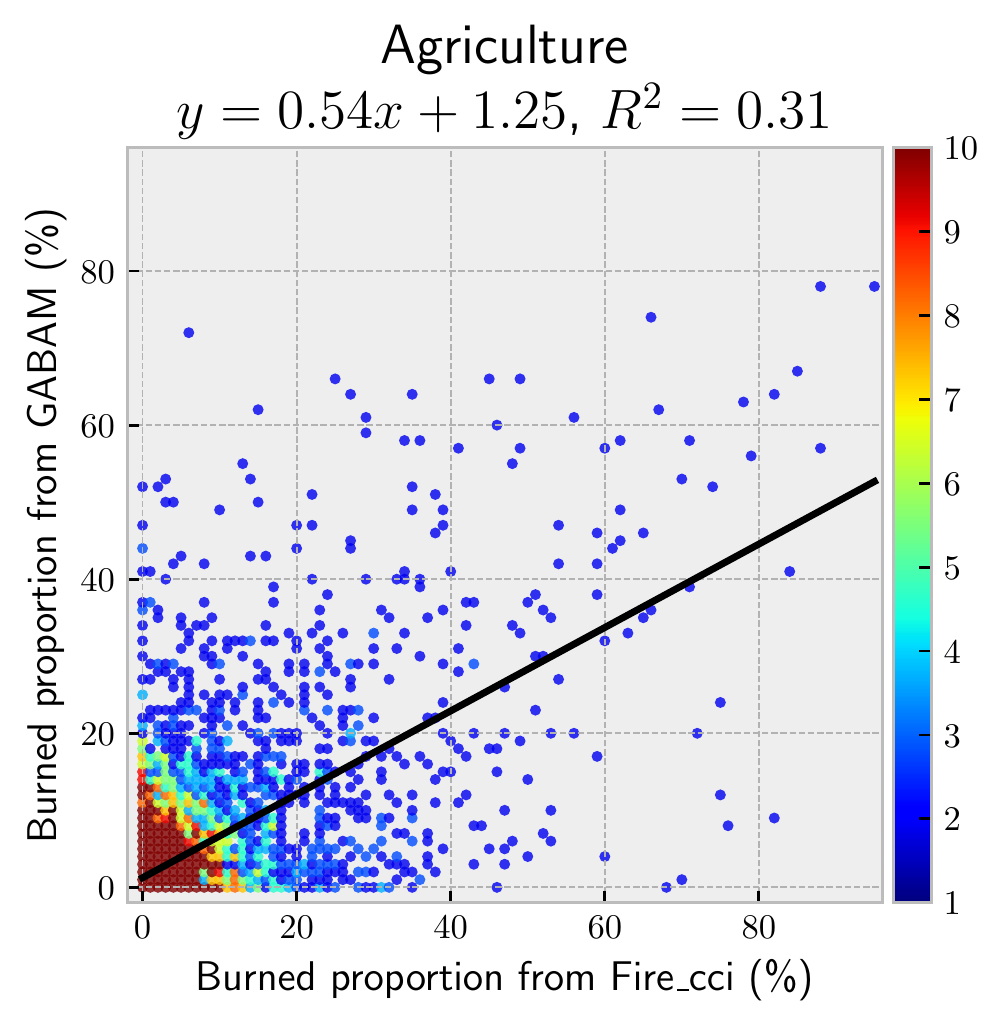}%
		\caption{}
		\label{fig:fcci-zone6}%
	\end{subfigure}
	\begin{subfigure}[t]{0.32\columnwidth}
		\includegraphics[width=\textwidth]{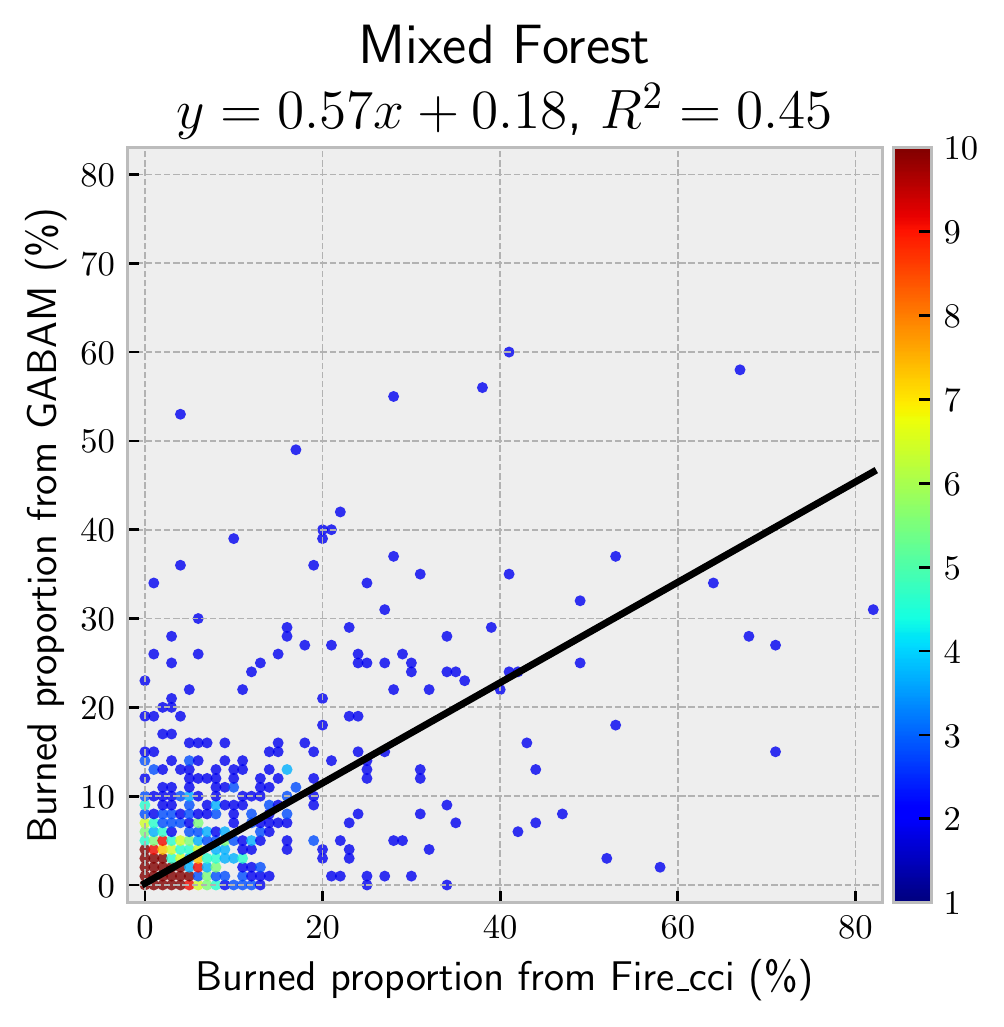}%
		\caption{}
		\label{fig:fcci-zone7}%
	\end{subfigure}
	\begin{subfigure}[t]{0.32\columnwidth}
		\includegraphics[width=\textwidth]{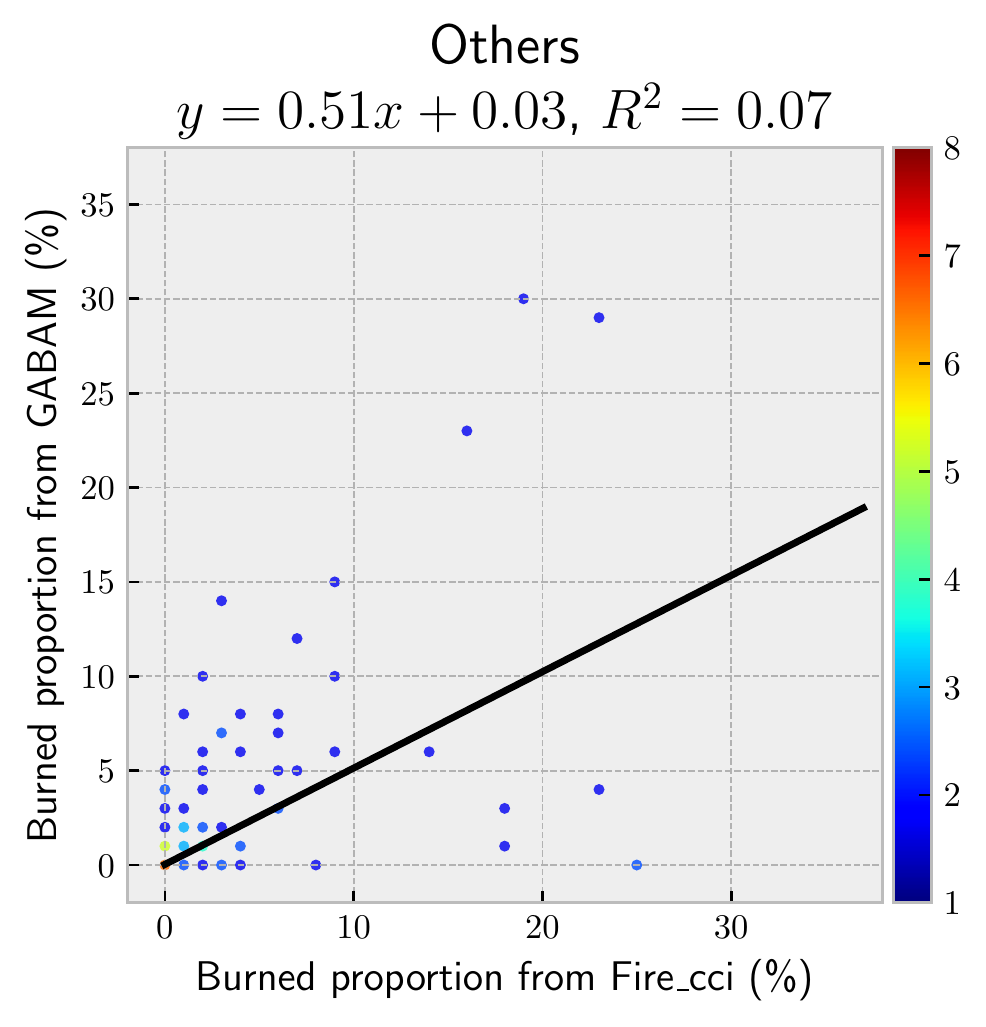}%
		\caption{}
		\label{fig:fcci-zone8}%
	\end{subfigure}
	\begin{subfigure}[t]{0.32\columnwidth}
		\includegraphics[width=\textwidth]{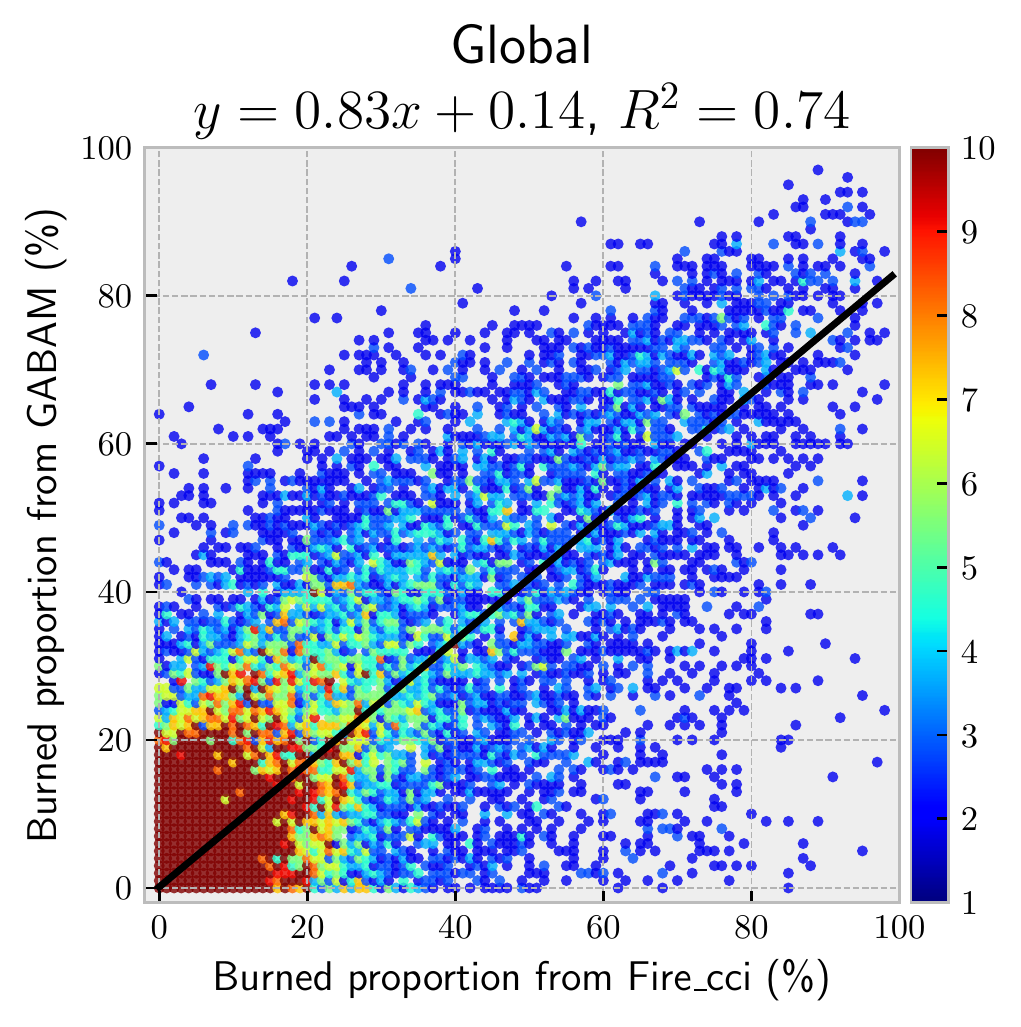}%
		\caption{}
		\label{fig:fcci-global}%
	\end{subfigure}
	\caption[]{
		Scatter graphs and regression lines between GABAM and Fire\_cci. \ref{fig:fcci-zone1}--\ref{fig:fcci-zone8} are the results in different land cover categories; \ref{fig:fcci-global} shows the global result in all kinds of land covers. The color scheme illustrates the number of grid cells having the same proportion values.
	}\label{fig:fcci}%
\end{figure}

 Figure~\ref{fig:fcci} shows the proportion of BA in $0.25^{\circ}\times0.25^{\circ}$ grids of different land cover categories in Table~\ref{tbl:categories}, for Fire\_cci product (x-axis) and GABAM 2015 (y-axis), and regression analysis was also performed between the two products, providing a regression line (expressed as the slope and the intercept coefficient estimates) and the coefficient of determination ($R^2$) for each land cover category (Figure~\ref{fig:fcci-zone1}--\ref{fig:fcci-zone8}) and for global scale (Figure~\ref{fig:fcci-global}). Moreover, as many points overlapped in the scatter graphs, we also rendered the scatters with different colors according to the number of grid cells (1 to 10 or more) having the same proportion values.

\par According to Figure~\ref{fig:fcci}, the intercept values of the estimated regression lines were close to 0 while the slopes were lower than 1, showing that GABAM 2015 generally underestimated~\citep{chuvieco2011esa} burned scars compared with Fire\_cci product. Moreover, the distribution and color of scatters in Figure~\ref{fig:fcci-global} also show that a large number of grids were considered to have higher burned proportion by Fire\_cci product than by GABAM. The main reason for the inconsistence can be attributed to the difference in spatial resolution of data sources, and less pixels were commonly classified as BA in Landsat images, e.g. Figure~\ref{fig:fcci-compare}. Specifically, only a few $0.25^\circ \times 0.25^\circ$ grids were occupied by more than 90\% BA in GABAM while grids with high proportion of BA were more common in Fire\_cci product.
\par Considering the coefficients of determination of estimated regression lines, the two products showed highest linear relationship strengths in coniferous forest ($R^2=0.82$), rangeland ($R^2=0.75$) and shrub ($R^2=0.62$), and lowest strengths in agriculture land ($R^2=0.31$) and ``Others'' category ($R^2=0.07$). In ``Others'' category, which is considered to be not prone to fire, the two products only included a few grids containing BA (with low burned proportions), thus they were not likely to be correlated; the low correlation in agriculture land is owing to the uncertainty of both products, which will be further discussed in \cref{sec:discussion}.
\par The quantity and color of scatters in Figure~\ref{fig:fcci} indicate that most of burned areas were located in rangeland, and the global relationship (Figure~\ref{fig:fcci-global}) of the GABAM and Fire\_cci product was mainly determined by that in rangeland (Figure~\ref{fig:fcci-zone5}), i.e. woody savannas, savannas and grasslands.

\subsection{Validation}
\subsubsection{Data sources}
\par Accuracy assessment was carried out according to the 80 validation sites which were created in \cref{subsec:sampling}, and the reference data were selected in these sites from multiple data sources, including fire perimeter datasets and satellite images. Commonly, when satellite data are used as reference data, they should have higher spatial resolution than the data used to generate the BA product~\citep{boschetti2009international}. For Landsat BA product, however, access to global higher-resolution time-series satellite data is difficult, and a thorough validation of Landsat science products can be completed with independent Landsat-derived reference data while strengthened by the use of complementary sources of high-resolution data~\citep{Vanderhoof_2017}. Consequently, in this study, some publicly available satellite images of higher-resolution were included in the validation scheme, while Landsat was the majority of validation data source. Specifically, Landsat-8 (LC8) images were employed to generate reference data independently for most of the validation sites except those located in the United States (U.S.), South America and China. In U.S., MTBS perimeters of 2015 were used as the supplemental reference data of LC8 images, and in South America and China, CBERS-4 MUX (CB4) and Gaofen-1 WFV (GF1) satellite images were used to create perimeters of burned area, respectively. The characteristics of CB4 and GF1 are illustrated in Table~\ref{tbl:cbers-gf1}. Note that the size of validation site varied by the type of data source, i.e. a WRS-II frame (about $185$km $\times$ $185$km) for Landsat images, a scene for CB4 images (about $120$km $\times$ $120$km) and a box of $100$km $\times$ $100$km for GF1 images. Using Landsat frames or image scenes as a unit of validation site is convenient for data downloading and processing; we chose a smaller box for GF1 to improve the data availability considering the extents of GF1 frames or scenes are not fixed due to the long orbital return period.
\begin{table}[!ht]
	\caption{Characteristics of CBERS-4 MUX and Gaofen-1 WFV.}
	\label{tbl:cbers-gf1}
	\centering
	\begin{threeparttable}
		\begin{tabular}{ccccccc}
			\toprule
			\multirow{2}{*}{\textbf{Sensors}} &\textbf{Spatial resolution} & \textbf{Swath width }  & \multicolumn{4}{c}{\textbf{Spectral bands ($\mu m$)}} \\
			& \textbf{at nadir (m)}      & \textbf{at nadir (km)} &  \textbf{blue}	& \textbf{green} & \textbf{red} & \textbf{NIR}\\
			\midrule
			CBERS-4 MUX		& 20 & 120 & \multirow{2}{*}{0.45-0.52}	& \multirow{2}{*}{0.52-0.59}	& \multirow{2}{*}{0.63-0.69}	& \multirow{2}{*}{0.77-0.89}\\
			Gaofen-1 WFV	& 16 & 192 & 	&  &  & \\
			\bottomrule
		\end{tabular}
	\end{threeparttable}
\end{table}

\subsubsection{Reference data generation}
\par In each validation site, all the available image scenes (LC8\footnote{\url{https://earthexplorer.usgs.gov
}}, CB4\footnote{\url{http://www.dgi.inpe.br/catalogo/}} or GF1\footnote{\url{http://218.247.138.119:7777/DSSPlatform/productSearch.html}}) acquired in 2015 were used. LC8 images were ortho-rectified surface reflectance products, CB4 images were ortho products, and GF1 images were not geometrically rectified. The procedure of generating reference BA can be summarized as following steps.
\begin{enumerate}
	\item \textbf{Preprocessing}
	\par All the images utilized to generate BA reference data were spatially aligned with mean squared error less than 1 pixel. The ortho-rectified LC8 and CB4 images met the requirement of geometric accuracy, yet the GF1 images did not. Accordingly, an automated method~\citep{Long_2016} was applied to ortho-rectify the time-series GF1 images, taking the LC8 panchromatic images (spatial resolution is 15 meters) as geo-references. 
	\item \textbf{BA Detection}
	\par BA perimeters were generated from the time-series images via a semi-automatic approach. Firstly, image pairs (pre- and post-fire) were manually selected from the time-series image by checking whether any new burned scars appeared in the newer images. For LC8 images, SWIR2, NIR and Green bands were composited in a Red, Green, Blue (RGB) combination; for CB4 and GF1 images, Red, NIR and Green bands were composited in an RGB combination. The identification of BA might be difficult for CB4 and GF1 images due to the lack of shortwave infrared bands, thus Fire\_cci BA product was used to verify the BA identification. Secondly, burned and unburned samples were manually collected from each selected image pair. The burned samples included only the newly burned scars, which appeared burned in the newer image but unburned in the older image; the unburned samples consisted of unburned pixels, partially recovered BA pixels, and also pixels covered by cloud or cloud shadows in either images. Afterwards, the support vector machines (SVM) classifier in ENVI\textsuperscript{TM} software were used to classify each image pair into burned and unburned pixels, and the detected burned pixels in all the image pairs were integrated to create a composited annual BA map. Note that the sensitive features in \cref{subsec:feature} were utilized in SVM for each LC8 image pair; but for CB4 and GF1 images, features used for classification consisted of the digital number (DN) values in four bands of an image pair (totally 8 DN values), as most of the burned-sensitive spectral indices cannot be derived from the RGB-NIR bands. Finally, the BA perimeters of 2015 were generated from the annual BA composition using the vectorization tool in ArcGIS\textsuperscript{TM} software.
	\item \textbf{Reviewing and manually revision}
	\par The result of supervised classifier (SVM) and automated vectorization algorithm might not be perfect, thus BA perimeters were further edited visually by experienced experts, via overlapping the vector layer of BA perimeters with the satellite image layers.
\end{enumerate}
\par Additionally, in U.S., MTBS perimeters of 2015 were directly used as the main reference data, supplemented by the interpreted results of LC8 time-series images, which could help to avoid missing of small fires.

\subsubsection{Validation results}
\par To assess the accuracy of GABAM 2015, a cross tabulation~\citep{Pontius_2011} between the pixels assigned by in our BA product and in the reference data was computed to produce the confusion matrix for each validation site. Afterwards, the global cross tabulation (Table~\ref{tbl:cross-tabulation}) was generated by averaging all the cross tabulations.

\begin{table}[!ht]
	\caption{Cross tabulation between GABAM 2015 and the reference data.}
	\label{tbl:cross-tabulation}
	\centering
	\begin{threeparttable}
		\begin{tabular}{lllll}
			\toprule
			\multicolumn{2}{c}{\multirow{2}{*}{}} &\multicolumn{3}{c}{\textbf{Reference data (pixel)}} \\
			& & \textbf{Burned}      & \textbf{Unburned} &  \textbf{Total}\\
			\midrule
			\multirow{3}{*}{\textbf{GABAM 2015 (pixel)}}& \textbf{Burned} & 5473720 ($X_{11}$) & 823170  \hspace{7pt} ($X_{12}$) & 6296890\\
			& \textbf{Unburned}  & 2360096  ($X_{21}$) & 43661559 ($X_{22}$)  & 46021655\\
			& \textbf{Total}       & 7833816 & 44484729 & 52318545\\
			\bottomrule
		\end{tabular}
	\end{threeparttable}
\end{table}

\par Finally, three statistics, i.e. commission error, omission error and overall accuracy, can be derived from the confusion matrix:
\begin{itemize}
	\item \textbf{Commission Error} ($E_c$)\textbf{:} $X_{12}/(X_{11}+X_{12})$, the ratio between the false BA positives (detected burned areas that were not in fact burned) and the total area classified as burned by GABAM 2015.
	\item \textbf{Omission Error} ($E_o$)\textbf{:} $X_{21}/(X_{11}+X_{21})$, the ratio between the false BA negatives (actual burned areas not detected) and the total area classified as burned by the reference data.
	\item \textbf{Overall Accuracy} ($A_o$)\textbf{:} $(X_{11}+X_{12})/(X_{11}+X_{12}+X_{21}+X_{22})$, the ratio between the area classified correctly and the total area to evaluate.
\end{itemize}

\par According to Table~\ref{tbl:cross-tabulation}, $E_c$ and $E_o$ of GABAM 2015 were 13.17\% and 30.13\%, respectively, while $A_o$ was 93.92\%. Generally, GABAM 2015 was expected to have a lower $E_c$ but a higher $E_o$. High omission error might result from several reasons: 
\begin{enumerate}
	\item In the validation sites located in tropical zones, clear burned evidences were frequently missed by Landsat sensor due to the quick recovery of the vegetation surface. This point will be further discussed in \cref{sec:discussion}.
	\item Some pixels located within a burned area, but not showing strong burned appearance, might be excluded by GABAM 2015 (e.g. Figure~\ref{fig:fcci-compare-gabam}), while they were considered as a part of a complete burned scar in the reference data. Particularly, high $E_o$ was found in those validation sites using MTBS perimeters, e.g. $E_c$ and $E_o$ of the validation site in Figure~\ref{fig:mtbs} were 1.45\% and 67.97\%.
\end{enumerate}
\par Table~\ref{tbl:validation-landcover} shows the average accuracy of GABAM 2015 in various land cover categories, and more details of validation can be found in \cref{app:a}, which includes 5 examples of validation sites from various regions, with different data sources as reference data.

\begin{table}[!ht]
	\caption{Information of site validation examples.}
	\label{tbl:validation-landcover}
	\centering
	\begin{threeparttable}
		\begin{tabular}{lrrr}
			\toprule
			\textbf{Land cover type}  & \textbf{$E_c$ (\%)} & \textbf{$E_o$ (\%)} & \textbf{$A_o$ (\%)}\\
			\midrule
			Broadleaved Evergreen & 8.64 & 10.95 & 90.99\\
			Broadleaved Deciduous & 23.59 & 34.85 & 99.03\\
			Coniferous & 7.41 & 18.27 & 99.77\\
			Mixed Forest & 8.73 & 34.33 & 98.36\\
			Shrub & 13.00 & 3.78 & 99.49\\
			Rangeland & 11.91 & 23.06 & 91.79\\
			Agriculture & 10.91 & 45.38& 94.41\\
			\bottomrule
		\end{tabular}
	\end{threeparttable}
\end{table}

\subsection{Discussion}\label{sec:discussion}
\par Different from the satellite images of coarse spatial resolution, the temporal resolution of Landsat images is not high enough to capture the short-term events on the earth. Specifically, the general revisit period of Landsat image is more than 10 days, hence active fire will be observed by Landsat satellite with probability less than 10\% (considering the cloud coverage). In addition, the gaps between Landsat images of adjacent time phases and the occurrence of cloud also increase the uncertainty to analyze the time-series patterns of land surface. Without using the evidence of active fire, it is not easy to identify the burned scars at global-scale with high confidence due to the wide variety of vegetation types, phenological characters and burned-like landcovers, and spectral characteristics within a burned scar (char, scorched leaves or grass, or even green leaves when the fire is not very severe\citep{Bastarrika_2011}). In this work, MODIS Vegetation Continuous Fields (VCF) product was applied to discriminate tree-dominated and grass-dominated regions, but the VCF product is neither precise in spatial resolution nor available before 2000 year and, moreover, two categories is far from enough to separate different burning types. Actually, much prior knowledge can be utilized to improve the accuracy of GABAM, if the globe is carefully divided into intensive regions according to the fire behaviour, land cover types and climate. For instance, most biomass burning in the tropics is limited to a burning season, around 10\% of the savanna biome burns every year, burning cropland after a harvest is extremely prevalent, and so on. Consequently, region-specified algorithms should be helpful to improve the accuracy of high-resolution global annual burned area mapping. Furthermore, despite of the high correlation between GABAM and Fire\_cci, the area of detected BA was generally smaller in GABAM than that in Fire\_cci, since some pixels located within a burned area but not showing strong burned appearance, were not included in GABAM. This situation can be considered as underestimation of BA or omission error if only taking into account the connectivity and completeness of burned patches; on the other hand, however, the detailed perimeter of BA from GABAM can be useful to statistics the area of biomes actually burned, and therefore to improve the simulation of carbon emissions from biomass burning.
At present form, however, GABAM suffers limitations in the following aspects.

\subsubsection{BA in Agriculture land} \label{subsubsec:agriculture}
\par It is difficult to detect BA in cropland with high confidence (low commission error and low omission error) from satellite images:
\begin{itemize}
	\item A lot of croplands have comparable spectral characteristics to burned areas when harvested or ploughed.
	\item The temporal behaviour of harvest or burning of cropland is similar to that of grassland fire, e.g. sudden decline and gradual recovery of NDVI, and periodic variation of NBR values year after year. 
	\item Different from the wildfires in rangeland and forest, most of the fires in croplands are human-intended stubble burning, and they are commonly small and of short duration, difficult to be captured by satellite sensors. In this sense, the traditional burned area detection algorithms, which are frequently used to generate BA products from data source of the middle resolution (e.g. MODIS, AVHRR, MERIS) are likely to have high omission error in croplands for small cropland fire. 
\end{itemize}

\par
\begin{figure}%
	\centering
	\begin{subfigure}[t]{0.245\columnwidth}
		\includegraphics[width=\textwidth]{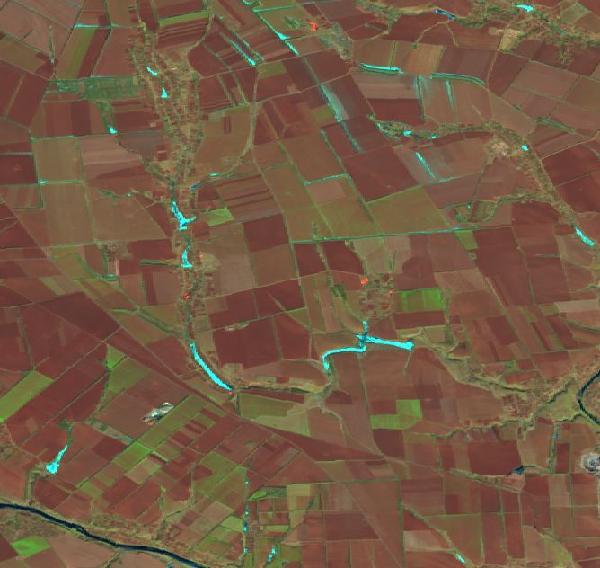}%
		\caption{2015-02-17}
		\label{fig:cropland-2015-02-17}%
	\end{subfigure}
	\begin{subfigure}[t]{0.245\columnwidth}
		\includegraphics[width=\textwidth]{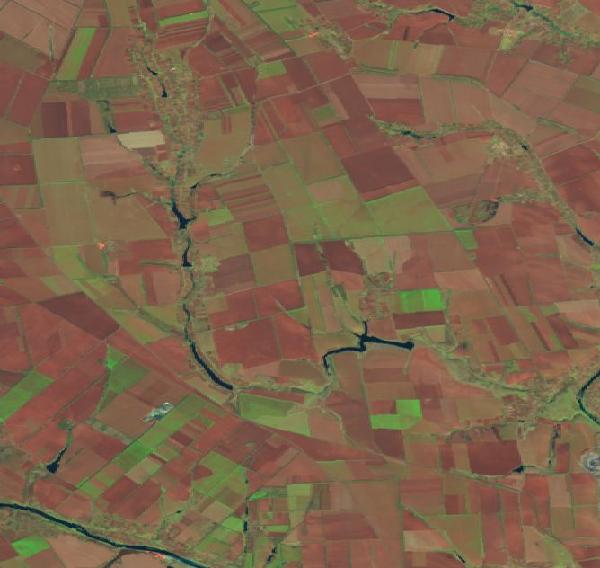}%
		\caption{2015-03-21}
		\label{fig:cropland-2015-03-21}%
	\end{subfigure}
	\begin{subfigure}[t]{0.245\columnwidth}
		\includegraphics[width=\textwidth]{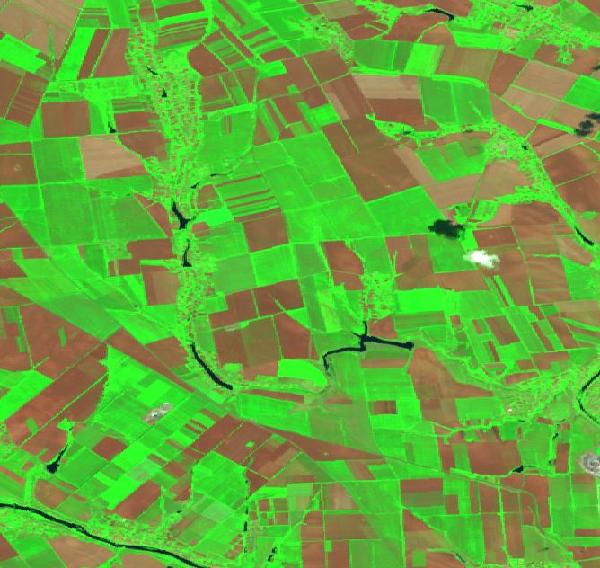}%
		\caption{2015-05-24}
		\label{fig:cropland-2015-05-24}%
	\end{subfigure}
	\begin{subfigure}[t]{0.245\columnwidth}
		\includegraphics[width=\textwidth]{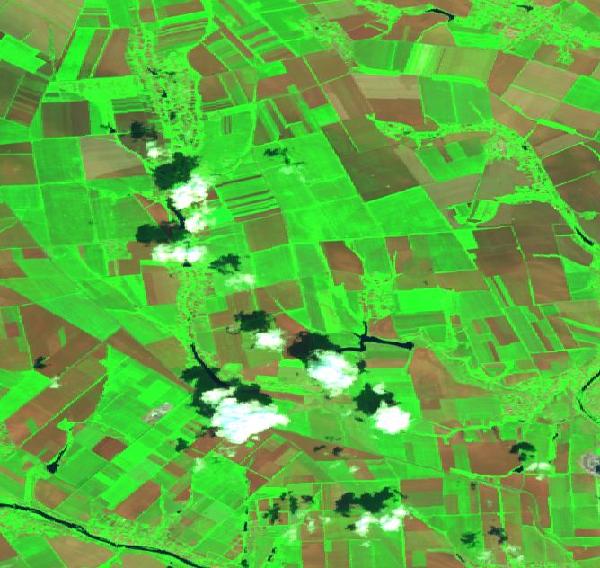}%
		\caption{2015-06-02}
		\label{fig:cropland-2015-06-02}%
	\end{subfigure}
	\begin{subfigure}[t]{0.245\columnwidth}
		\includegraphics[width=\textwidth]{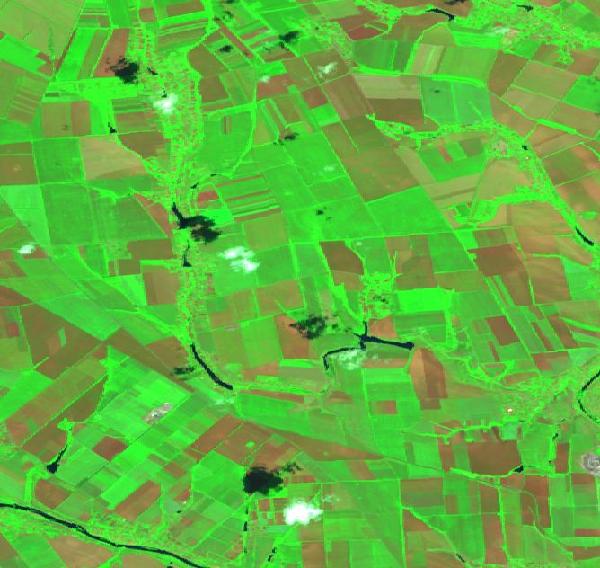}%
		\caption{2015-06-09}
		\label{fig:cropland-2015-06-09}%
	\end{subfigure}
	\begin{subfigure}[t]{0.245\columnwidth}
		\includegraphics[width=\textwidth]{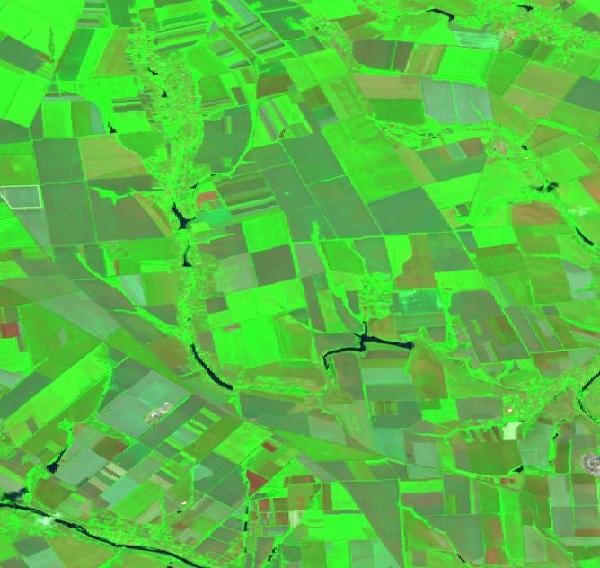}%
		\caption{2015-06-25}
		\label{fig:cropland-2015-06-25}%
	\end{subfigure}
	\begin{subfigure}[t]{0.245\columnwidth}
		\includegraphics[width=\textwidth]{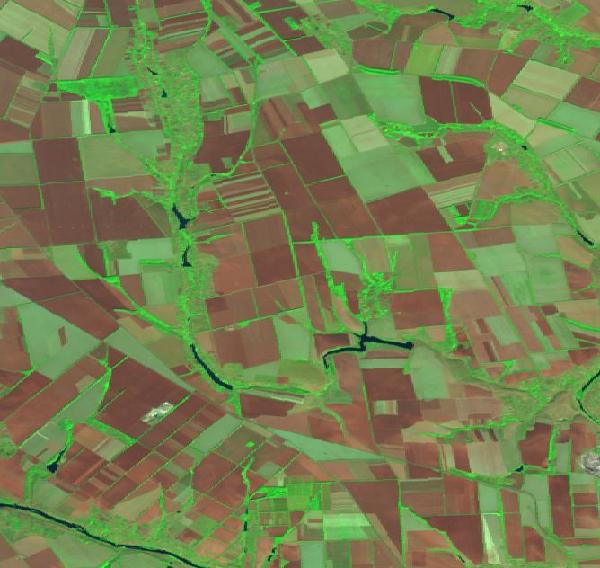}%
		\caption{2015-09-22}
		\label{fig:cropland-2015-09-22}%
	\end{subfigure}	
	\begin{subfigure}[t]{0.245\columnwidth}
		\includegraphics[width=\textwidth]{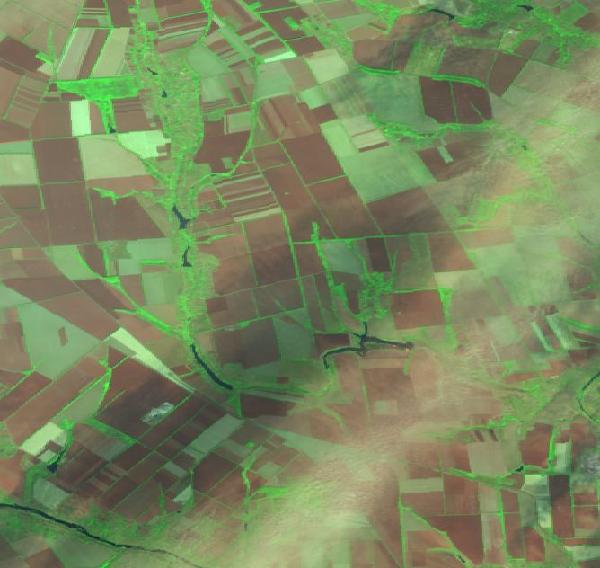}%
		\caption{2015-09-29}
		\label{fig:cropland-2015-09-29}%
	\end{subfigure}	
	\begin{subfigure}[t]{0.245\columnwidth}
		\includegraphics[width=\textwidth]{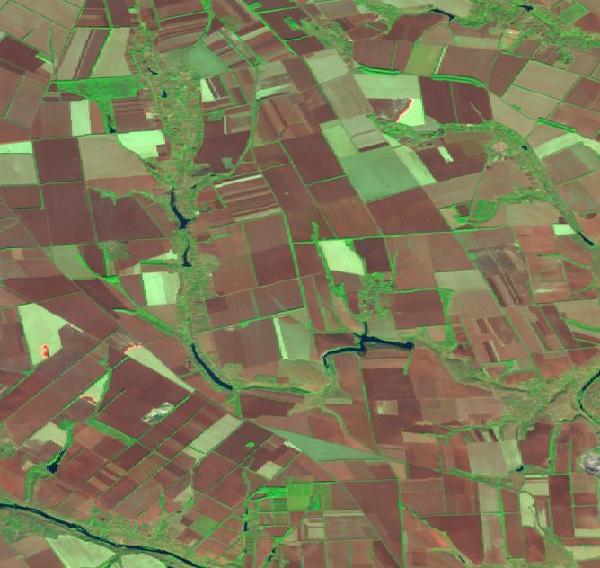}%
		\caption{2015-10-15}
		\label{fig:cropland-2015-10-15}%
	\end{subfigure}	
	\begin{subfigure}[t]{0.245\columnwidth}
		\includegraphics[width=\textwidth]{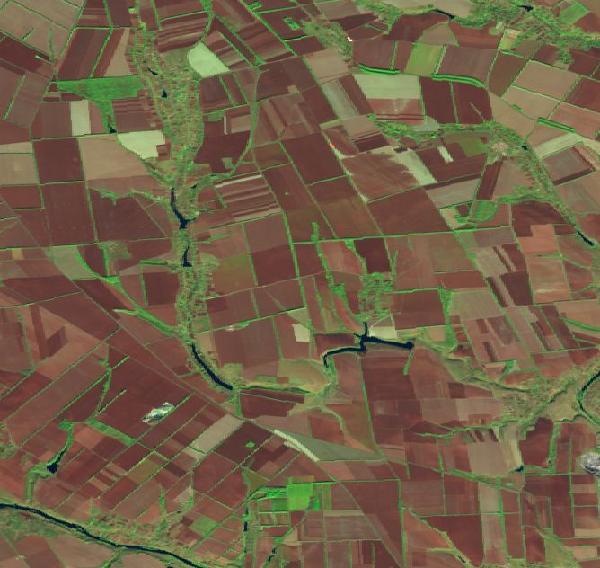}%
		\caption{2015-10-31}
		\label{fig:cropland-2015-10-31}%
	\end{subfigure}
	\begin{subfigure}[t]{0.245\columnwidth}
		\includegraphics[width=\textwidth]{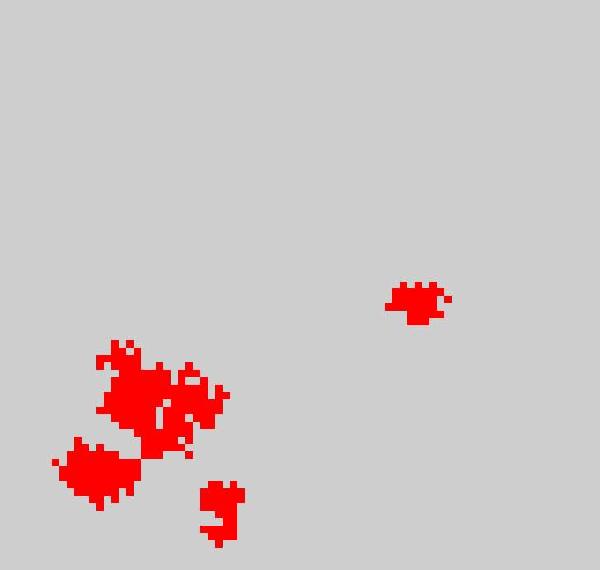}%
		\caption{BA from Fire\_cci}
		\label{fig:cropland-fcci}%
	\end{subfigure}	
	\begin{subfigure}[t]{0.245\columnwidth}
		\includegraphics[width=\textwidth]{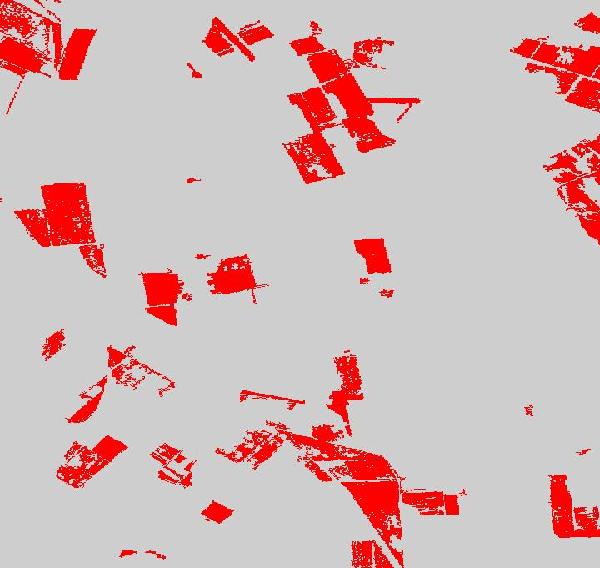}%
		\caption{BA from GABAM}
		\label{fig:cropland-gabam}%
	\end{subfigure}
	\caption[]{
		Burned area map example of croplands in Mykolayiv, Ukraine. \ref{fig:cropland-2015-02-17}--\ref{fig:cropland-2015-10-31} shows the Landsat-8 images displayed in false color composition (red: SWIR2 band, green: NIR band and blue: GREEN band); \ref{fig:cropland-fcci} and \ref{fig:cropland-gabam} show the BA from Fire\_cci product and GABAM 2015, respectively.
	}\label{fig:cropland}%
\end{figure}

 Figure~\ref{fig:cropland} shows an example of cropland in Mykolayiv, Ukraine, including the Landsat-8 time series (Figure~\ref{fig:cropland-2015-02-17}--\ref{fig:cropland-2015-10-31}) and the burned scars mapped by Fire\_cci (Figure~\ref{fig:cropland-fcci}) and GABAM  (Figure~\ref{fig:cropland-gabam}). Small fire spots, showing light orange color, can be visually observed from Figure~\ref{fig:cropland-2015-02-17}, \ref{fig:cropland-2015-03-21} and \ref{fig:cropland-2015-10-15}, but burned scars surrounding these fire spots were not included in Fire\_cci product. On the other hand, without fire evidence or field validation, it is also difficult to tell whether the burned-like surfaces detected by GABAM were false alarms.
\par Due to these difficulties, discriminating true-burned areas from croplands is not a trivial task, and cropland masks can be employed to remove potential confusions.

\subsubsection{Omission of observations} \label{subsubsec:omission}
\par
\begin{figure}[htbp]%
	\centering
	\begin{subfigure}[t]{0.2\columnwidth}
		\centering
		\includegraphics[width=\textwidth]{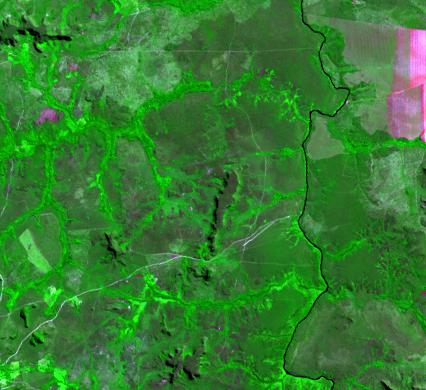}%
		\caption{2015-05-16 (LC8)}
		\label{fig:l8-2015-05-16}%
	\end{subfigure}
	\begin{subfigure}[t]{0.2\columnwidth}
		\centering
		\includegraphics[width=\textwidth]{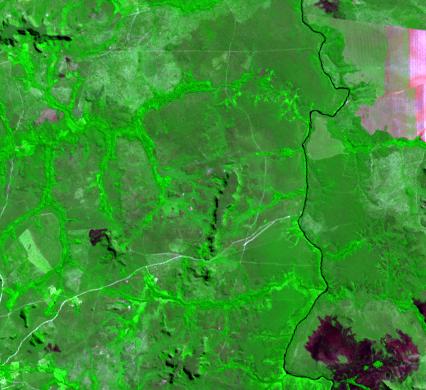}%
		\caption{2015-06-01 (LC8)}
		\label{fig:l8-2015-06-01}%
	\end{subfigure}
	\begin{subfigure}[t]{0.2\columnwidth}
		\centering
		\includegraphics[width=\textwidth]{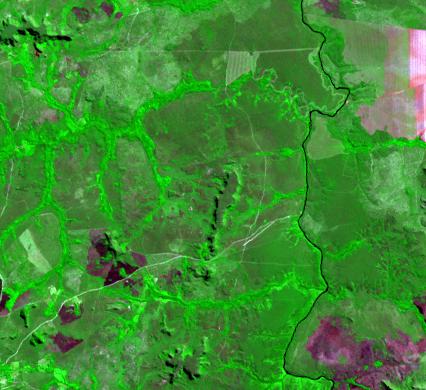}%
		\caption{2015-06-17 (LC8)}
		\label{fig:l8-2015-06-17}%
	\end{subfigure}
	\begin{subfigure}[t]{0.2\columnwidth}
		\centering
		\includegraphics[width=\textwidth]{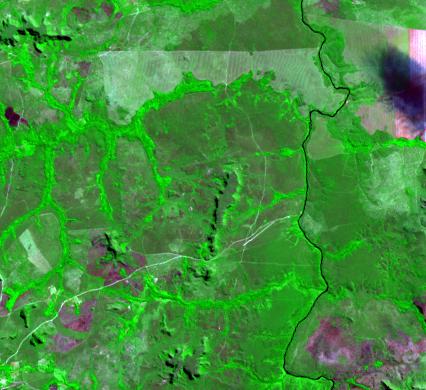}%
		\caption{2015-07-03 (LC8)}
		\label{fig:l8-2015-07-03}%
	\end{subfigure}
	\begin{subfigure}[t]{0.2\columnwidth}
		\centering
		\includegraphics[width=\textwidth]{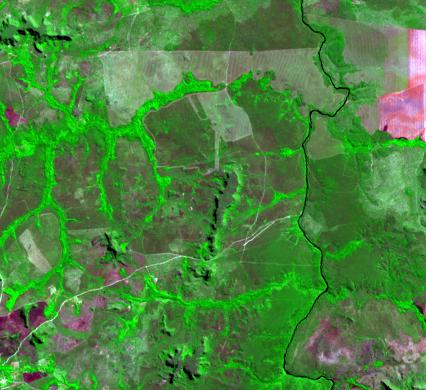}%
		\caption{2015-07-19 (LC8)}
		\label{fig:l8-2015-07-19}%
	\end{subfigure}	
	\begin{subfigure}[t]{0.2\columnwidth}
		\centering
		\includegraphics[width=\textwidth]{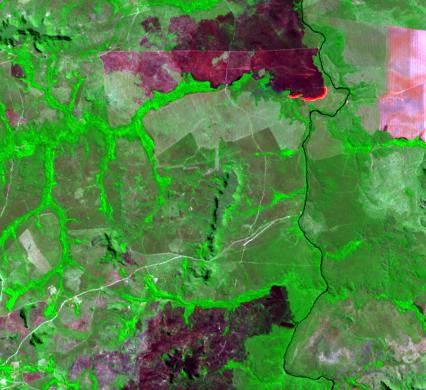}%
		\caption{2015-08-04 (LC8)}
		\label{fig:l8-2015-08-04}%
	\end{subfigure}
	\begin{subfigure}[t]{0.2\columnwidth}
		\centering
		\includegraphics[width=\textwidth]{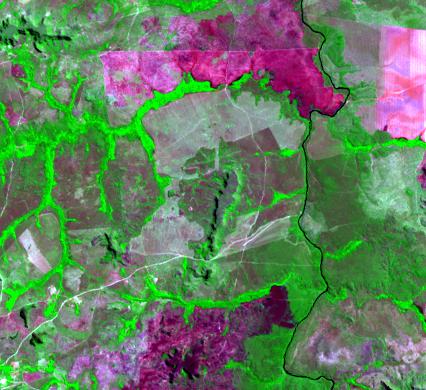}%
		\caption{2015-08-20 (LC8)}
		\label{fig:l8-2015-08-20}%
	\end{subfigure}	
	\begin{subfigure}[t]{0.2\columnwidth}
		\centering
		\includegraphics[width=\textwidth]{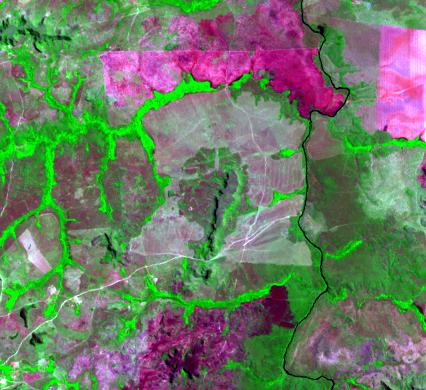}%
		\caption{2015-09-05 (LC8)}
		\label{fig:l8-2015-09-05}%
	\end{subfigure}
	\begin{subfigure}[t]{0.2\columnwidth}
		\centering
		\includegraphics[width=\textwidth]{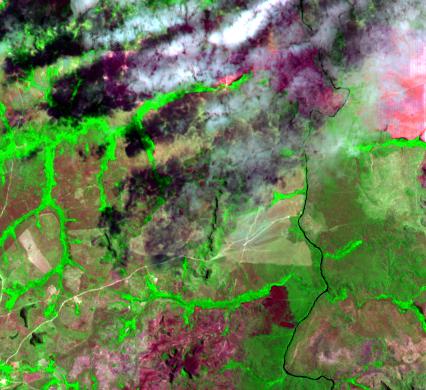}%
		\caption{2015-09-21 (LC8)}
		\label{fig:l8-2015-09-21}%
	\end{subfigure}	
	\begin{subfigure}[t]{0.2\columnwidth}
		\centering
		\includegraphics[width=\textwidth]{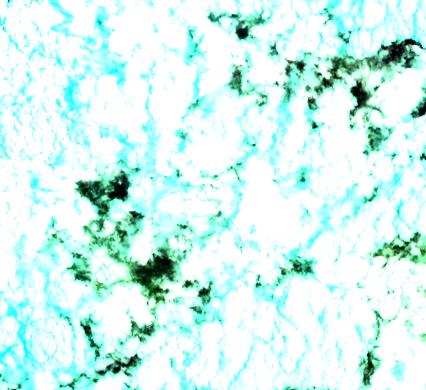}%
		\caption{2015-10-07 (LC8)}
		\label{fig:l8-2015-10-07}%
	\end{subfigure}
	\begin{subfigure}[t]{0.2\columnwidth}
		\centering
		\includegraphics[width=\textwidth]{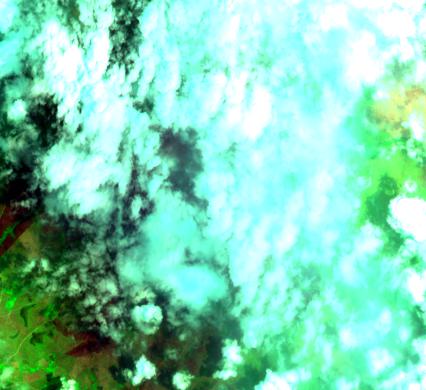}%
		\caption{2015-10-23 (LC8)}
		\label{fig:l8-2015-10-23}%
	\end{subfigure}
	\begin{subfigure}[t]{0.2\columnwidth}
		\centering
		\includegraphics[width=\textwidth]{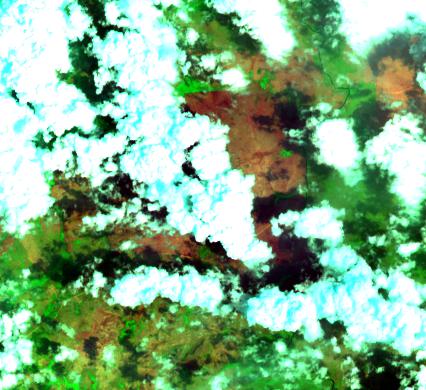}%
		\caption{2015-11-08 (LC8)}
		\label{fig:l8-2015-11-08}%
	\end{subfigure}
	\begin{subfigure}[t]{0.2\columnwidth}
		\centering
		\includegraphics[width=\textwidth]{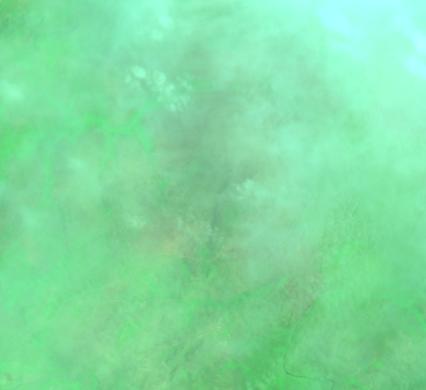}%
		\caption{2015-11-24 (LC8)}
		\label{fig:l8-2015-11-24}%
	\end{subfigure}
	\begin{subfigure}[t]{0.2\columnwidth}
		\centering
		\includegraphics[width=\textwidth]{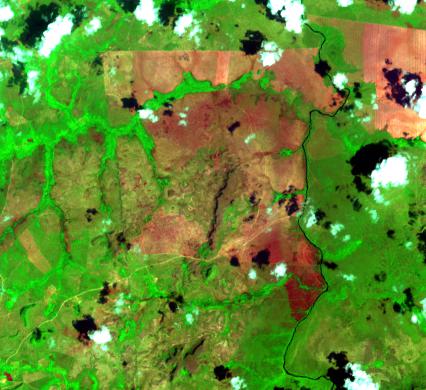}%
		\caption{2015-12-10 (LC8)}
		\label{fig:l8-2015-12-10}%
	\end{subfigure}
	\begin{subfigure}[t]{0.2\columnwidth}
		\centering
		\includegraphics[width=\textwidth]{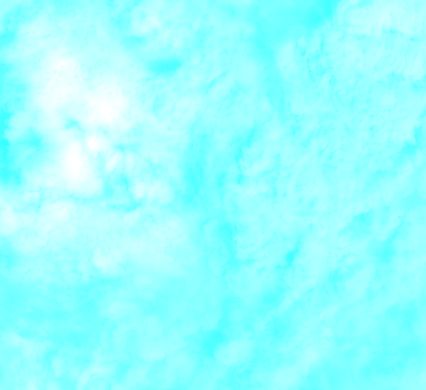}%
		\caption{2015-12-26 (LC8)}
		\label{fig:l8-2015-12-26}%
	\end{subfigure}
	\begin{subfigure}[t]{0.2\columnwidth}
		\centering
		\includegraphics[width=\textwidth]{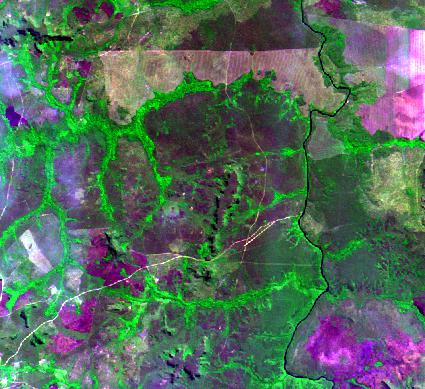}%
		\caption{2015-06-30 (CB4)}
		\label{fig:CBERS4-2015-06-30}%
	\end{subfigure}
	\begin{subfigure}[t]{0.2\columnwidth}
		\centering
		\includegraphics[width=\textwidth]{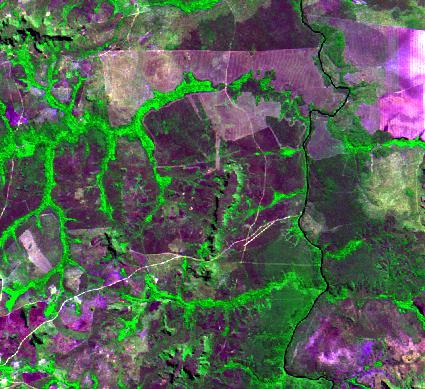}%
		\caption{2015-07-26 (CB4)}
		\label{fig:CBERS4-2015-07-26}%
	\end{subfigure}
	\begin{subfigure}[t]{0.2\columnwidth}
		\centering
		\includegraphics[width=\textwidth]{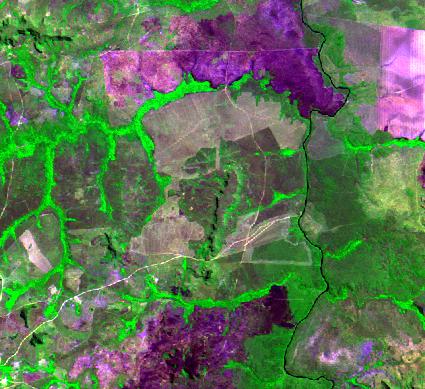}%
		\caption{2015-08-21 (CB4)}
		\label{fig:CBERS4-2015-08-21}%
	\end{subfigure}
	\begin{subfigure}[t]{0.2\columnwidth}
		\centering
		\includegraphics[width=\textwidth]{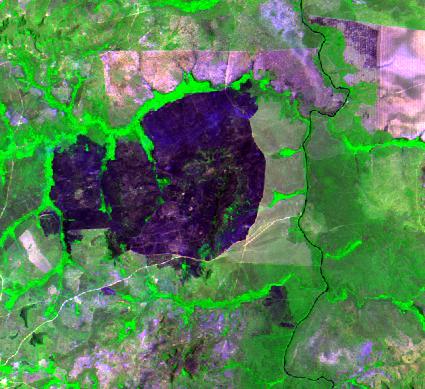}%
		\caption{2015-10-12 (CB4)}
		\label{fig:CBERS4-2015-10-12}%
	\end{subfigure}
	\begin{subfigure}[t]{0.2\columnwidth}
		\centering
		\includegraphics[width=\textwidth]{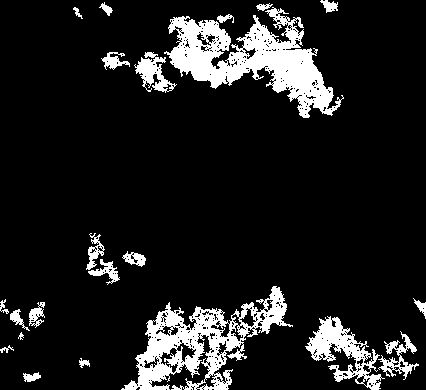}%
		\caption{Detected BA}
		\label{fig:missing-ba}%
	\end{subfigure}
	\caption[]{
		Example of omission error of GABAM 2015. \ref{fig:l8-2015-05-16}--\ref{fig:l8-2015-12-26} are Landsat-8 image patches displayed in false color composition (red: SWIR2 band, green: NIR band and blue: GREEN band), \ref{fig:CBERS4-2015-06-30}--\ref{fig:CBERS4-2015-10-12} are CBERS-4 image patches displayed in false color composition (red: NIR band, green: RED band and blue: GREEN band), and ~\ref{fig:missing-ba} shows the detected BA.
	}\label{fig:missing}%
\end{figure}

 Using Landsat images as input data for GABAM, the number of valid observations is a limiting factor for detecting fires, since the active- or post-fire evidence may be omitted or weaken due to the temporal gaps caused by temporal resolution as well as cloud contamination. Especially in Tropical regions, where vegetation recovery is quite quick after fire, temporal gaps usually result in high omission error. Figure~\ref{fig:missing} shows an example of omission error in South America. From the CBERS-4 images (Figure~\ref{fig:CBERS4-2015-06-30}--\ref{fig:CBERS4-2015-10-12}), a new burned scar, which occurred during August 21 to October 12, can be identified at the center of image patch. However, all the Landsat-8 images (Figure~\ref{fig:l8-2015-05-16}--\ref{fig:l8-2015-12-26}) acquired between the date interval from September 21 to November 24 were contaminated by cloud, thus the region covering this burned scar in these images were masked by QA band during the process of BA detecting.

\subsubsection{Validation} \label{subsubsec:validation}
\par For satellite data product validation, a commonly used method is to employ higher spatial resolution satellite data. For example, in order to validate MODIS derived data product (1 km spatial resolution), Landsat satellite data is commonly used. In this study, however, Landsat images were used as the main reference source to validate Landsat derived burned area product. Although the validation process was conducted by independent experienced experts with great caution, relying on Landsat for both product generation and validation limits our ability to assess inaccuracies imposed by the satellite sensor itself, such as radiometric calibration accuracy, spectral band settings, geolocation and mixed pixels~\citep{strahler2006global}. Accordingly, extensive validation of GABAM is expected to be further performed by professional users.

\section{Conclusions}
An automated pipeline for generating 30m resolution global-scale annual burned area map utilizing Google Earth Engine was proposed in this study. Different from the previous coarse resolution global burned area products, GABAM 2015, a novel 30-m resolution global annual burned area map of 2015 year, was derived from all available Landsat-8 images, and its commission error and omission error are 13.17\% and 30.13\%, respectively, according to global validation. 
Comparison with Fire\_cci product showed a similar spatial distribution and strong correlation between the burned areas from the two products, particularly in coniferous forests. 
The automated pipeline makes it possible to efficiently generate GABAM from huge catalog of Landsat images, and our future effort will be concentrated to produce long time-series 30m resolution GABAMs.

\section*{Acknowledgments}%
\addtocontents{toc}{\protect\vspace{6pt}}%
\addcontentsline{toc}{section}{Acknowledgments}%
This research has been supported by The National Key Research and Development Program of China (2016YFA0600302 and 2016YFB0501502), and National Natural Science Foundation of China (61401461 and 61701495).

\appendix 
\renewcommand{\thesubsection}{\Alph{subsection}}
\section{Examples of validation sites} \label{app:a}

\begin{figure}[H]%
	\centering
	\begin{subfigure}[t]{0.32\columnwidth}
		\centering
		\includegraphics[width=\textwidth]{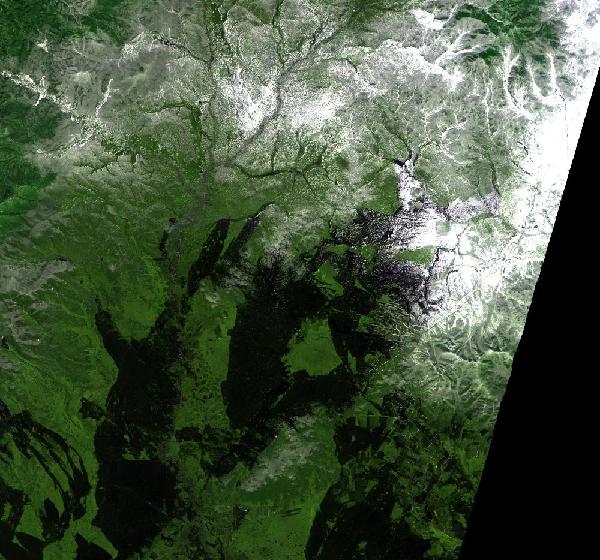}%
		\caption{2015-04-01 (GF1)}
		\label{fig:gf1-2015-04-01}%
	\end{subfigure}
	\begin{subfigure}[t]{0.32\columnwidth}
		\centering
		\includegraphics[width=\textwidth]{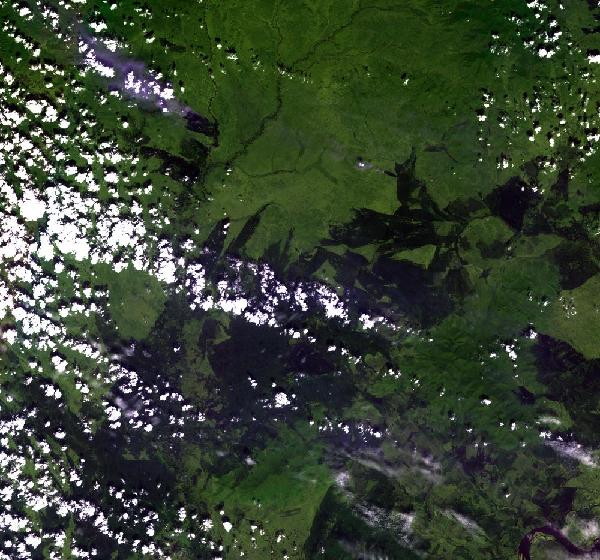}%
		\caption{2015-05-03 (GF1)}
		\label{fig:gf1-2015-05-03}%
	\end{subfigure}
	\begin{subfigure}[t]{0.32\columnwidth}
		\centering
		\includegraphics[width=\textwidth]{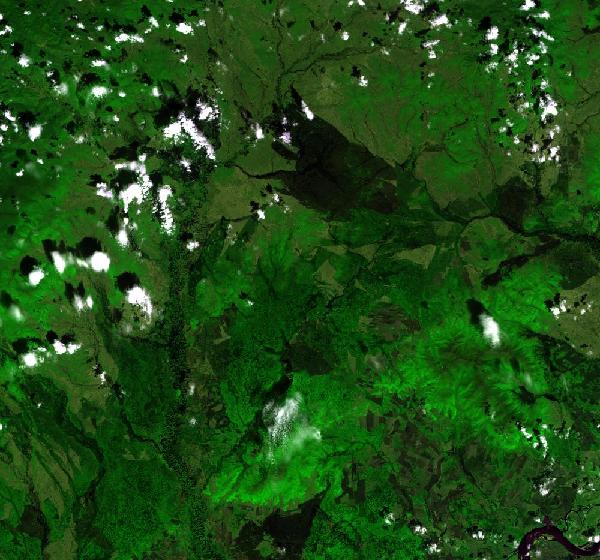}%
		\caption{2015-05-20 (GF1)}
		\label{fig:gf1-2015-05-20}%
	\end{subfigure}
	\begin{subfigure}[t]{0.32\columnwidth}
		\centering
		\includegraphics[width=\textwidth]{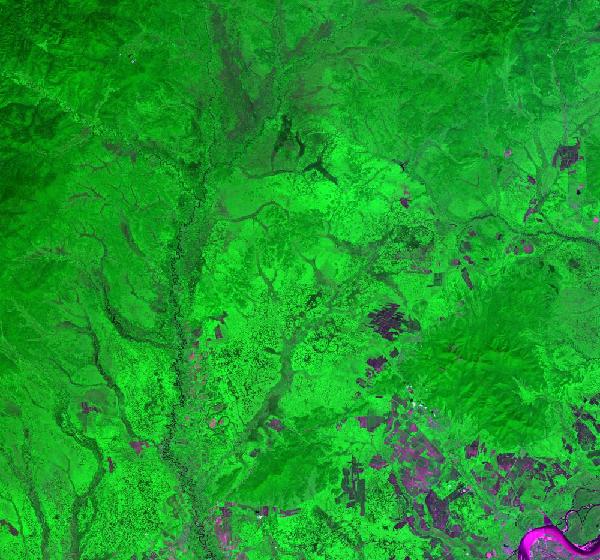}%
		\caption{2015-07-16 (GF1)}
		\label{fig:gf1-2015-07-16}%
	\end{subfigure}	
	\begin{subfigure}[t]{0.32\columnwidth}
		\centering
		\includegraphics[width=\textwidth]{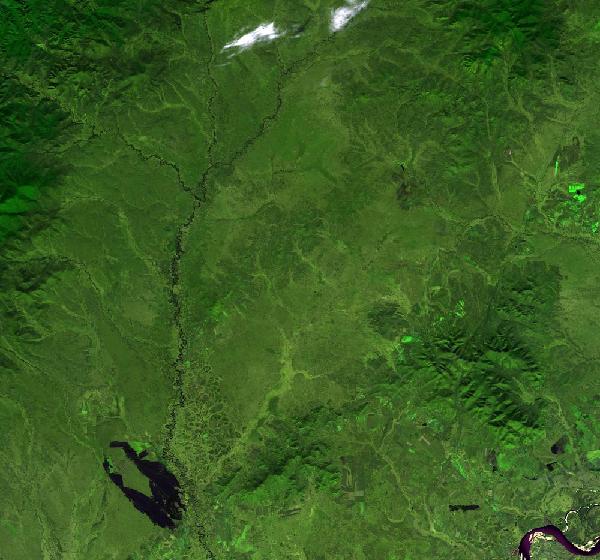}%
		\caption{2015-10-15 (GF1)}
		\label{fig:gf1-2015-10-15}%
	\end{subfigure}	
	\begin{subfigure}[t]{0.32\columnwidth}
		\centering
		\includegraphics[width=\textwidth]{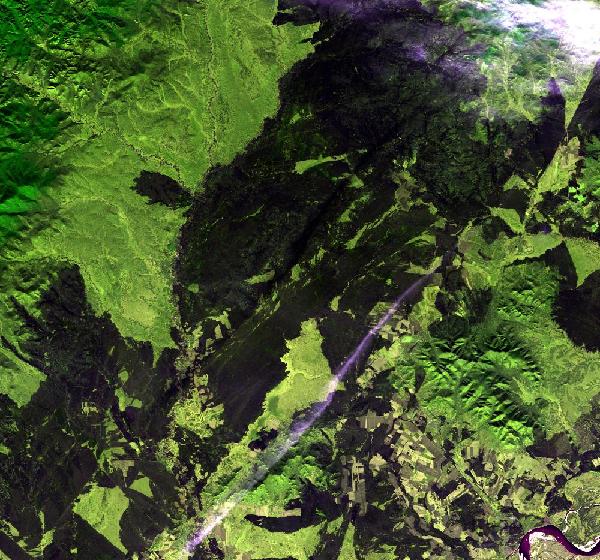}%
		\caption{2015-11-08 (GF1)}
		\label{fig:gf1-2015-11-08}%
	\end{subfigure}
	\begin{subfigure}[t]{0.32\columnwidth}
		\centering
		\includegraphics[width=\textwidth]{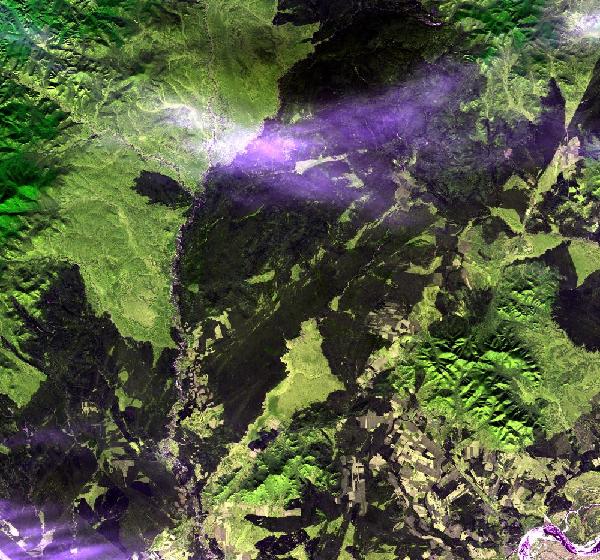}%
		\caption{2015-11-20 (GF1)}
		\label{fig:gf1-2015-11-20}%
	\end{subfigure}
	\begin{subfigure}[t]{0.32\columnwidth}
		\centering
		\includegraphics[width=\textwidth]{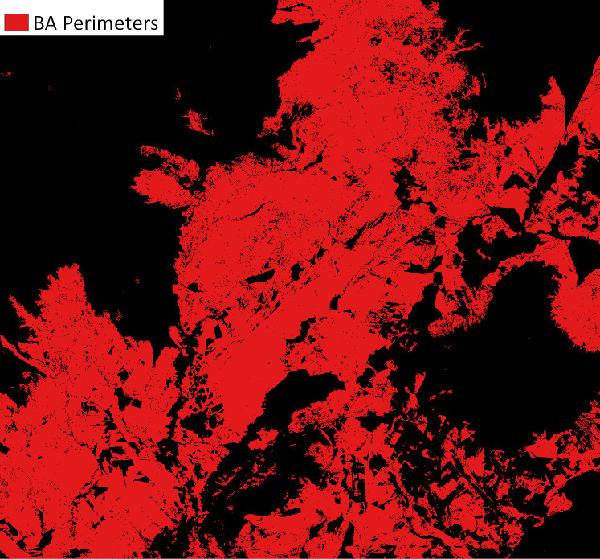}%
		\caption{BA (GF1)}
		\label{fig:gf1-ba-gf1}%
	\end{subfigure}
	\begin{subfigure}[t]{0.32\columnwidth}
		\centering
		\includegraphics[width=\textwidth]{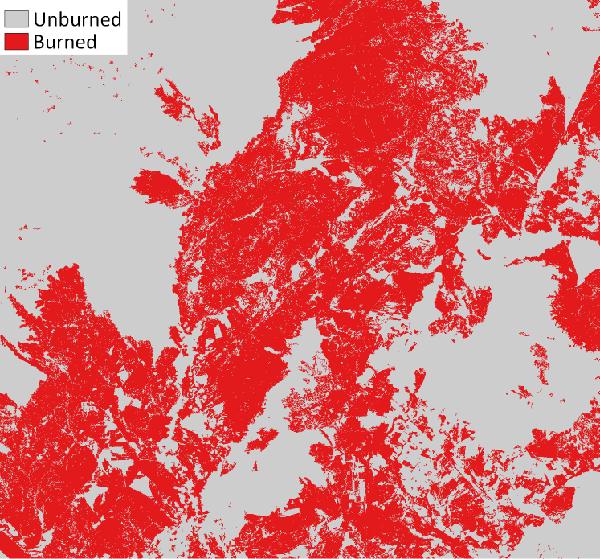}%
		\caption{Detected BA}
		\label{fig:gf1-ba-landsat}%
	\end{subfigure}
	\caption[]{
		Example of validation using GF-1 images. \ref{fig:gf1-2015-04-01}--\ref{fig:gf1-2015-11-20} show the GF-1 images used to generate reference map, displayed in false color composition (red: NIR band, green: RED band and blue: GREEN band), \ref{fig:gf1-ba-gf1} is reference BA map generated from GF-1 images, and~\ref{fig:gf1-ba-landsat} is detected BA by proposed method.
	}\label{fig:GF1}%
\end{figure}

\begin{figure}[H]%
	\centering
	\begin{subfigure}[t]{0.32\columnwidth}
		\centering
		\includegraphics[width=\textwidth]{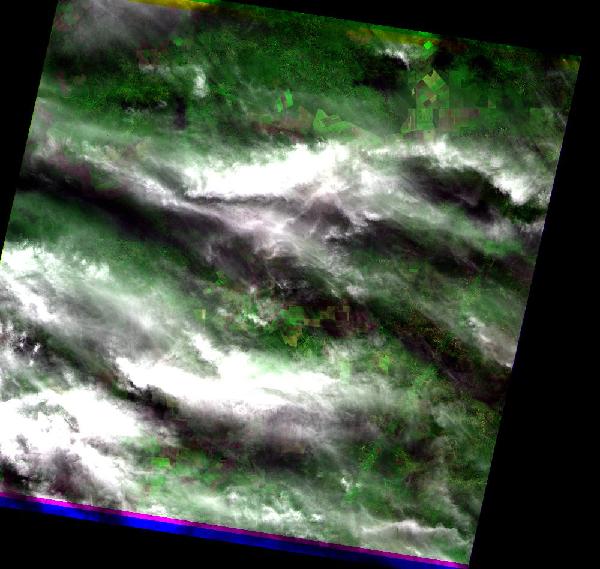}%
		\caption{2015-06-01 (CB4)}
		\label{fig:cb4-2015-06-01}%
	\end{subfigure}
	\begin{subfigure}[t]{0.32\columnwidth}
		\centering
		\includegraphics[width=\textwidth]{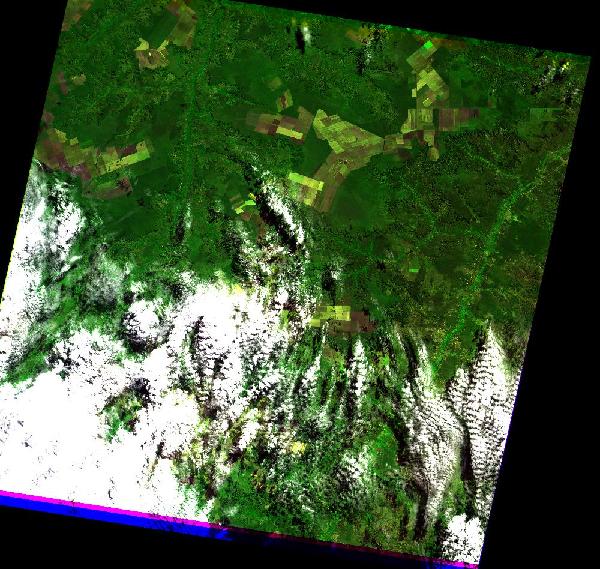}%
		\caption{2015-07-06 (CB4)}
		\label{fig:cb4-2015-07-06}%
	\end{subfigure}
	\begin{subfigure}[t]{0.32\columnwidth}
		\centering
		\includegraphics[width=\textwidth]{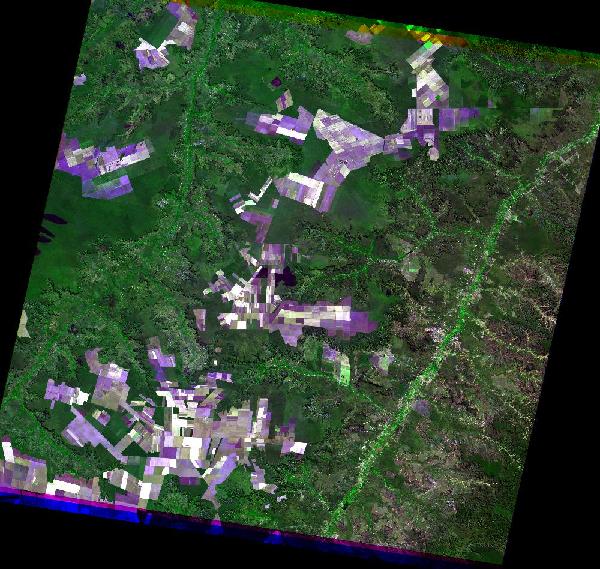}%
		\caption{2015-08-01 (CB4)}
		\label{fig:cb4-2015-08-01}%
	\end{subfigure}
	\begin{subfigure}[t]{0.32\columnwidth}
		\centering
		\includegraphics[width=\textwidth]{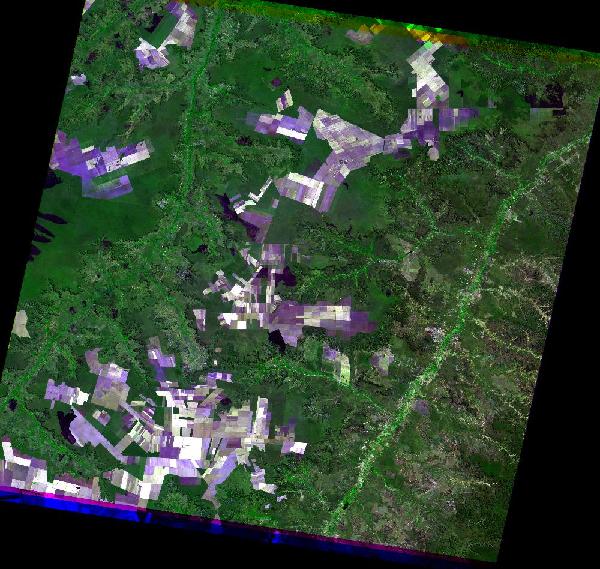}%
		\caption{2015-08-27 (cb4)}
		\label{fig:cb4-2015-08-27}%
	\end{subfigure}	
	\begin{subfigure}[t]{0.32\columnwidth}
		\centering
		\includegraphics[width=\textwidth]{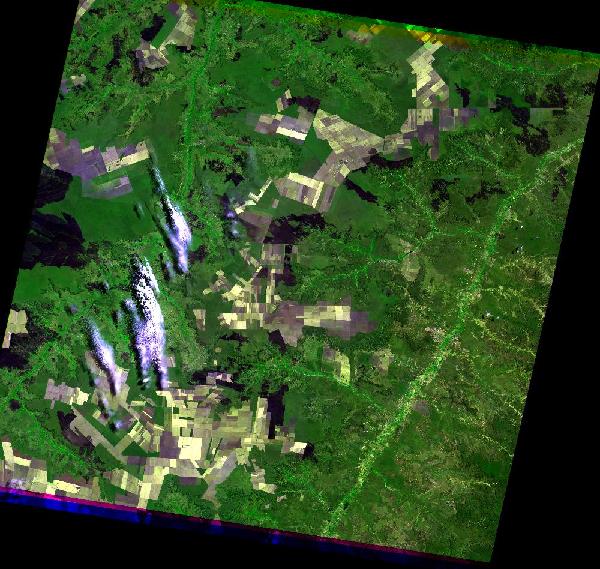}%
		\caption{2015-09-22 (CB4)}
		\label{fig:c4b-2015-09-22}%
	\end{subfigure}	
	\begin{subfigure}[t]{0.32\columnwidth}
		\centering
		\includegraphics[width=\textwidth]{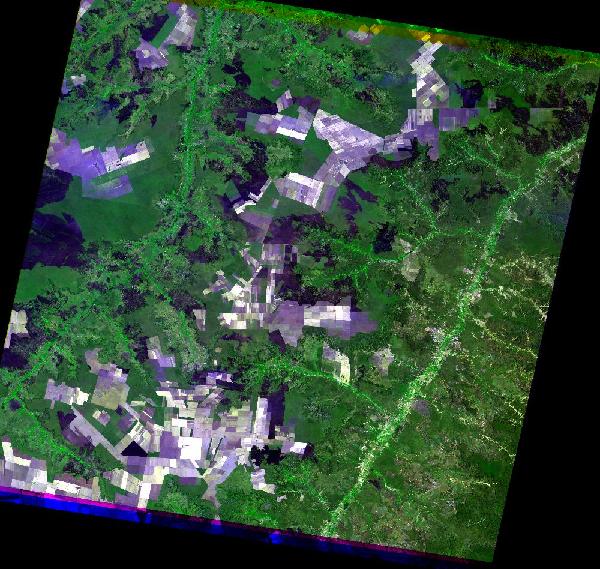}%
		\caption{2015-11-18 (CB4)}
		\label{fig:c4b-2015-10-18}%
	\end{subfigure}
	\begin{subfigure}[t]{0.32\columnwidth}
		\centering
		\includegraphics[width=\textwidth]{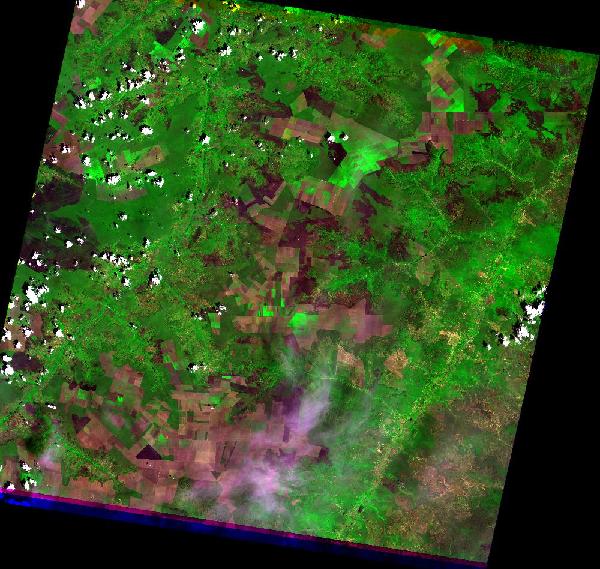}%
		\caption{2015-12-09 (CB4)}
		\label{fig:cb4-2015-12-09}%
	\end{subfigure}
	\begin{subfigure}[t]{0.32\columnwidth}
		\centering
		\includegraphics[width=\textwidth]{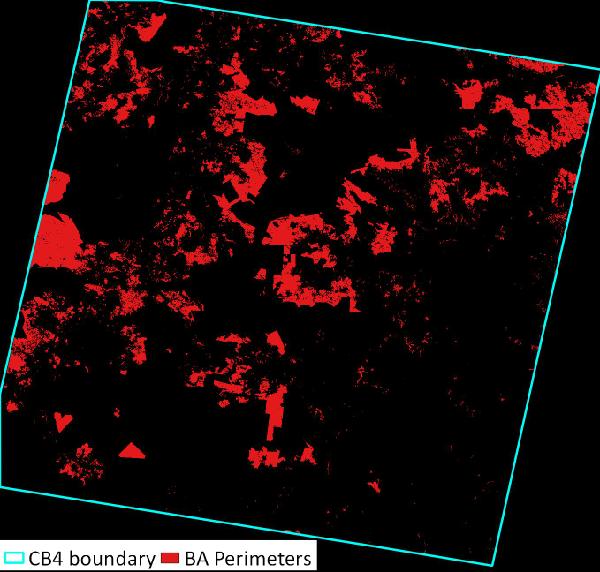}%
		\caption{BA (CB4)}
		\label{fig:cb4-ba-cb4}%
	\end{subfigure}
	\begin{subfigure}[t]{0.32\columnwidth}
		\centering
		\includegraphics[width=\textwidth]{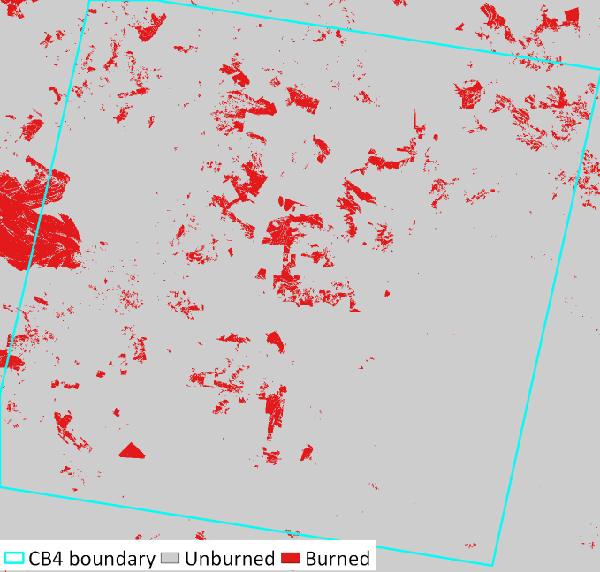}%
		\caption{Detected BA}
		\label{fig:cb4-ba-landsat}%
	\end{subfigure}
	\caption[]{
		Example of validation using CBERS-4 images. \ref{fig:cb4-2015-06-01}--\ref{fig:cb4-2015-12-09} show the CBERS-4 images used to generate reference map, displayed in false color composition (red: NIR band, green: RED band and blue: GREEN band), \ref{fig:cb4-ba-cb4} is reference BA map generated from CBERS-4 images, and~\ref{fig:cb4-ba-landsat} is detected BA by proposed method.
	}\label{fig:CB4}%
\end{figure}

\begin{figure}[H]%
	\centering
	\begin{subfigure}[t]{0.245\columnwidth}
		\centering
		\includegraphics[width=\textwidth]{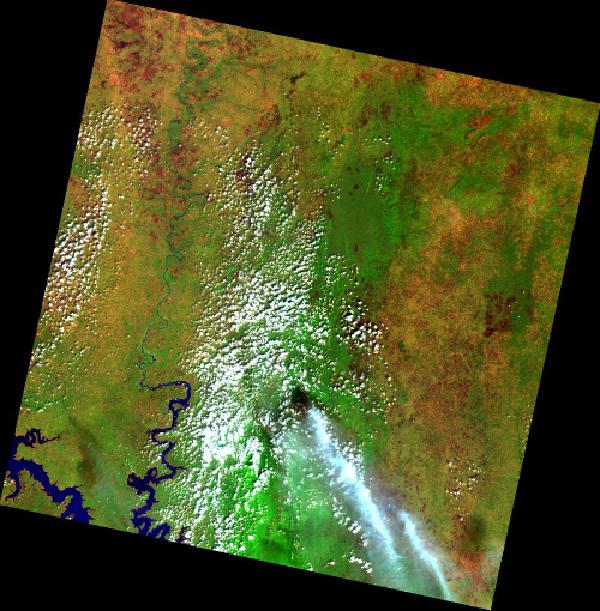}%
		\caption{2014-12-19 (LC8)}
		\label{fig:lc8-20141219}%
	\end{subfigure}
	\begin{subfigure}[t]{0.245\columnwidth}
		\centering
		\includegraphics[width=\textwidth]{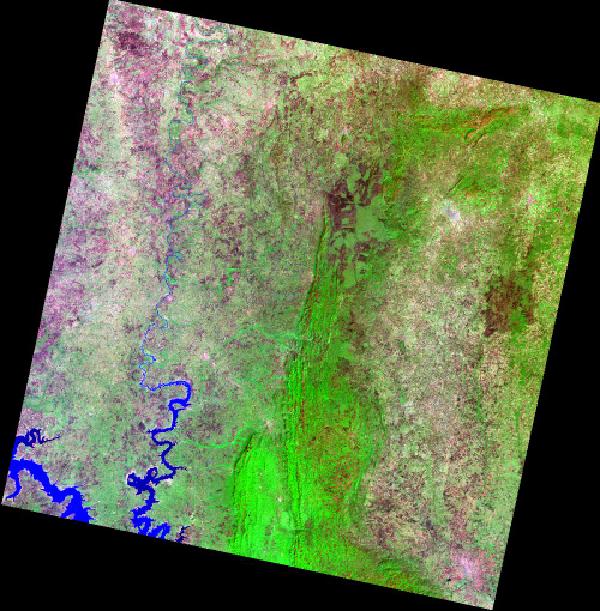}%
		\caption{2015-01-04 (LC8)}
		\label{fig:lc8-20150104}%
	\end{subfigure}
	\begin{subfigure}[t]{0.245\columnwidth}
		\centering
		\includegraphics[width=\textwidth]{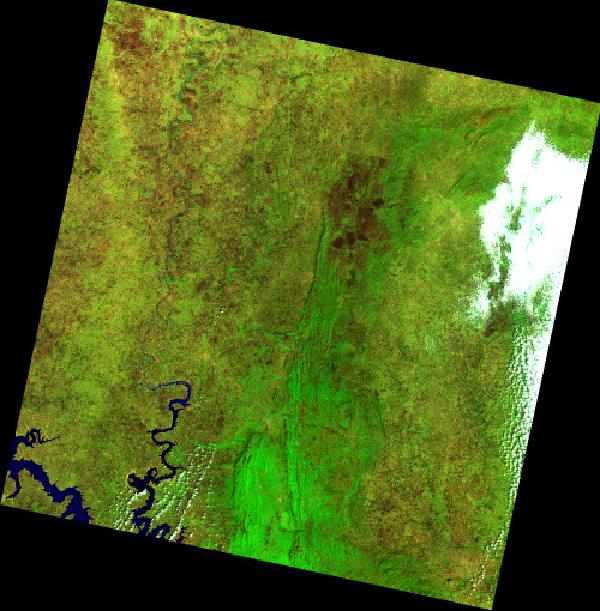}%
		\caption{2015-01-20 (LC8)}
		\label{fig:lc8-20150120}%
	\end{subfigure}
	\begin{subfigure}[t]{0.245\columnwidth}
		\centering
		\includegraphics[width=\textwidth]{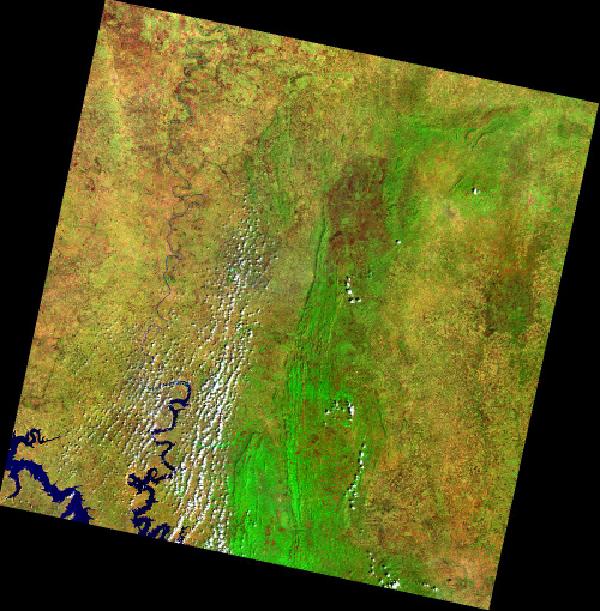}%
		\caption{2015-02-05 (LC8)}
		\label{fig:lc8-20150205}%
	\end{subfigure}
	\begin{subfigure}[t]{0.245\columnwidth}
		\centering
		\includegraphics[width=\textwidth]{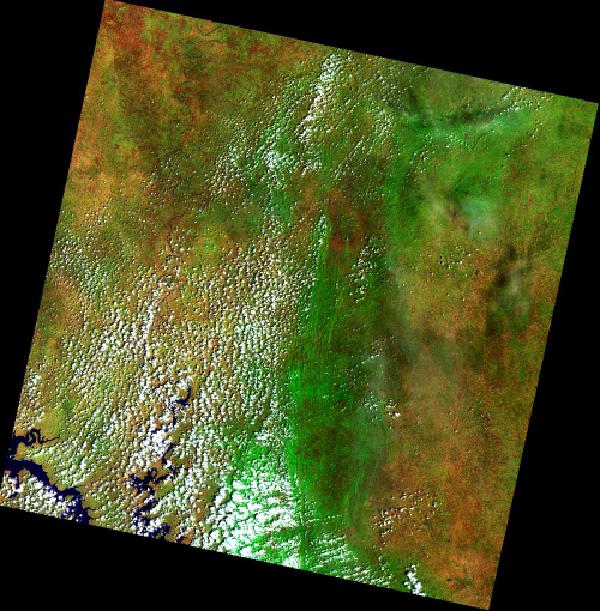}%
		\caption{2015-02-21 (LC8)}
		\label{fig:lc8-20150221}%
	\end{subfigure}
	\begin{subfigure}[t]{0.245\columnwidth}
		\centering
		\includegraphics[width=\textwidth]{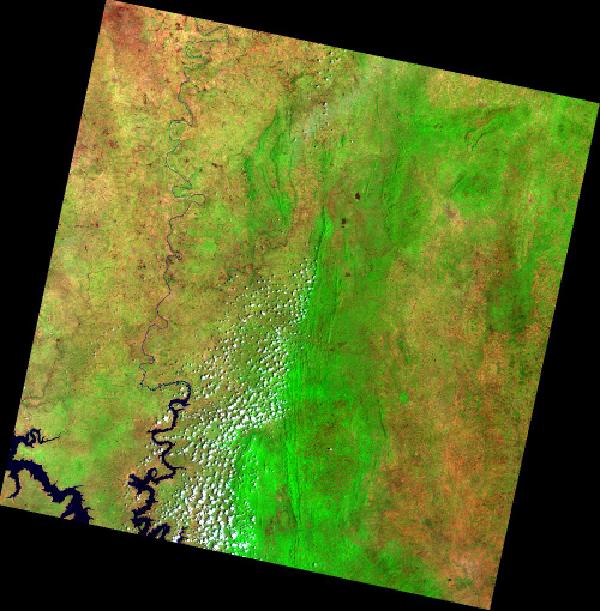}%
		\caption{2015-03-09 (LC8)}
		\label{fig:lc8-20150309}%
	\end{subfigure}
	\begin{subfigure}[t]{0.245\columnwidth}
		\centering
		\includegraphics[width=\textwidth]{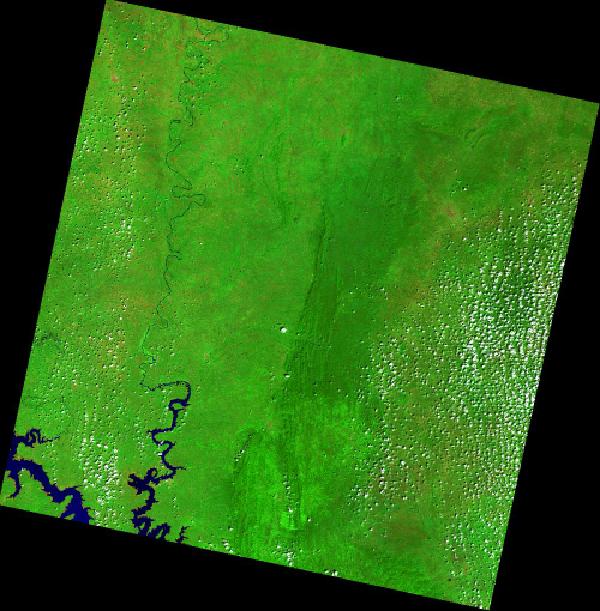}%
		\caption{2015-11-04 (LC8)}
		\label{fig:lc8-20151104}%
	\end{subfigure}
	\begin{subfigure}[t]{0.245\columnwidth}
		\centering
		\includegraphics[width=\textwidth]{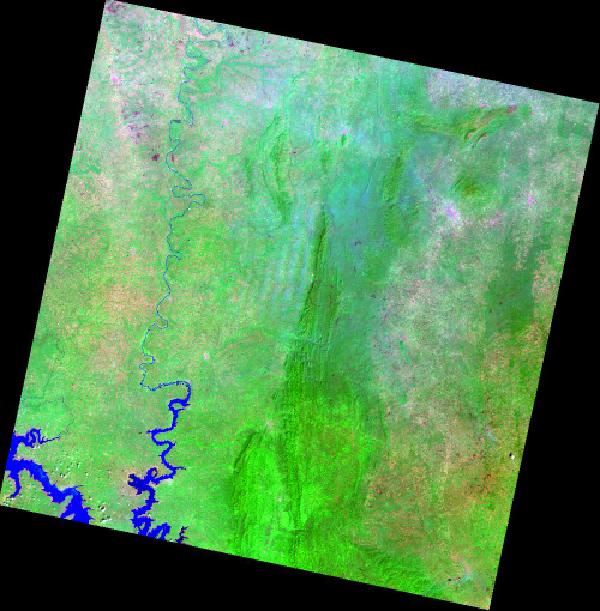}%
		\caption{2015-11-20 (LC8)}
		\label{fig:lc8-20151120}%
	\end{subfigure}
	\begin{subfigure}[t]{0.245\columnwidth}
		\centering
		\includegraphics[width=\textwidth]{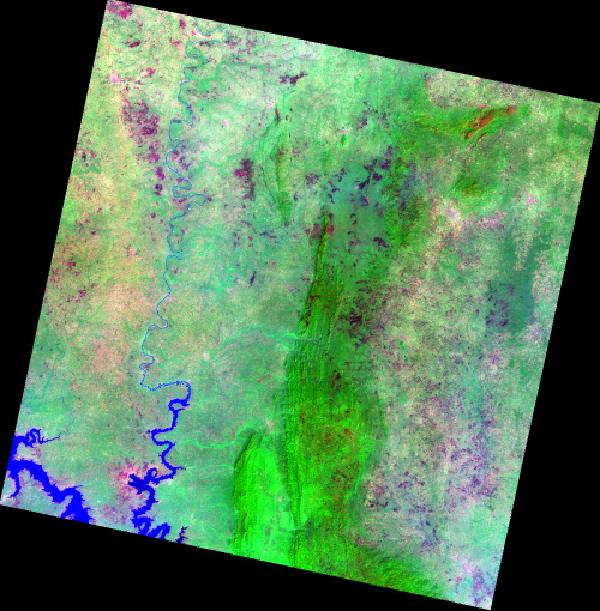}%
		\caption{2015-12-06 (LC8)}
		\label{fig:lc8-20151206}%
	\end{subfigure}
	\begin{subfigure}[t]{0.245\columnwidth}
		\centering
		\includegraphics[width=\textwidth]{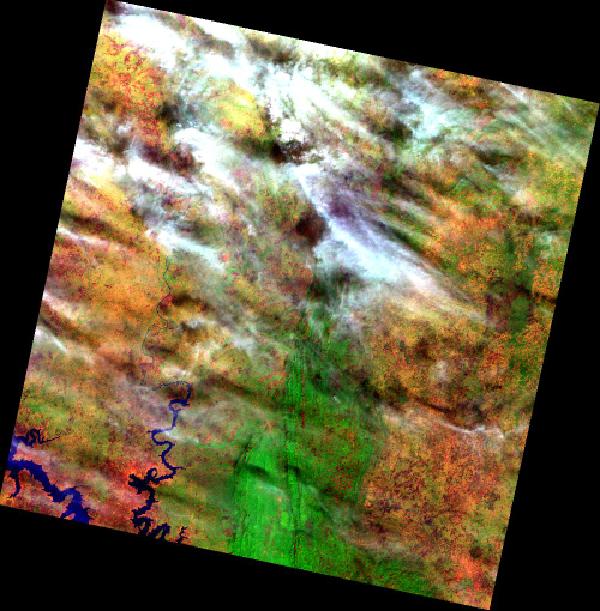}%
		\caption{2015-12-22 (LC8)}
		\label{fig:lc8-20151222}%
	\end{subfigure}
	\begin{subfigure}[t]{0.245\columnwidth}
		\centering
		\includegraphics[width=\textwidth]{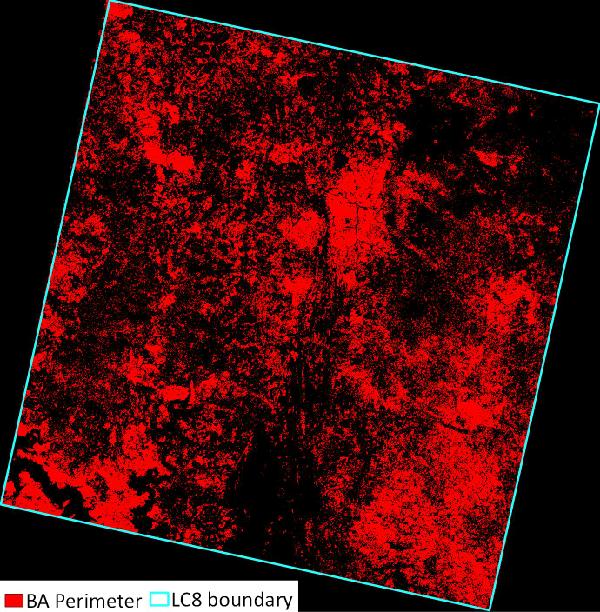}%
		\caption{BA (LC8)}
		\label{fig:lc8-africa-reference}%
	\end{subfigure}
	\begin{subfigure}[t]{0.245\columnwidth}
		\centering
		\includegraphics[width=\textwidth]{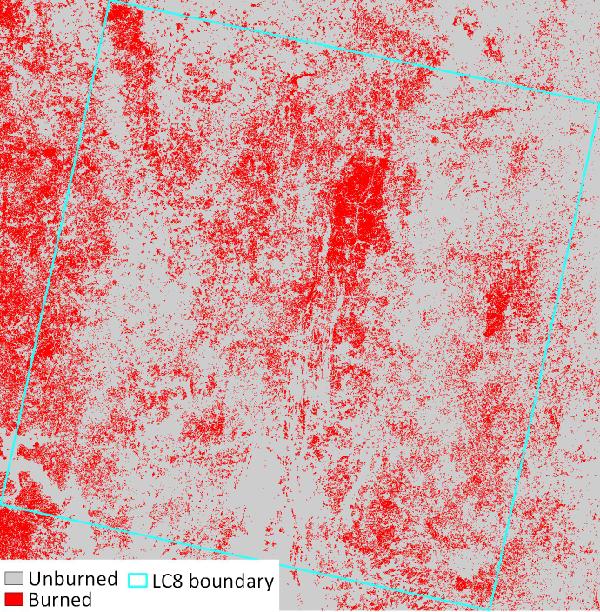}%
		\caption{Detected BA}
		\label{fig:lc8-africa-gabam}%
	\end{subfigure}
	\caption[]{
		Example of validation using Landsat-8 images (path/row:193/054) in Africa. \ref{fig:lc8-20141219}--\ref{fig:lc8-20151222} show the Landsat-8 images used to generate reference map, displayed in false color composition (red: SWIR2 band, green: NIR band and blue: GREEN band), \ref{fig:lc8-africa-reference} is reference BA map generated from Landsat-8 images, and~\ref{fig:lc8-africa-gabam} is detected BA by proposed method.
	}\label{fig:lc8-africa}%
\end{figure}

\begin{figure}[H]%
	\centering
	\begin{subfigure}[t]{0.30\columnwidth}
		\centering
		\includegraphics[width=\textwidth]{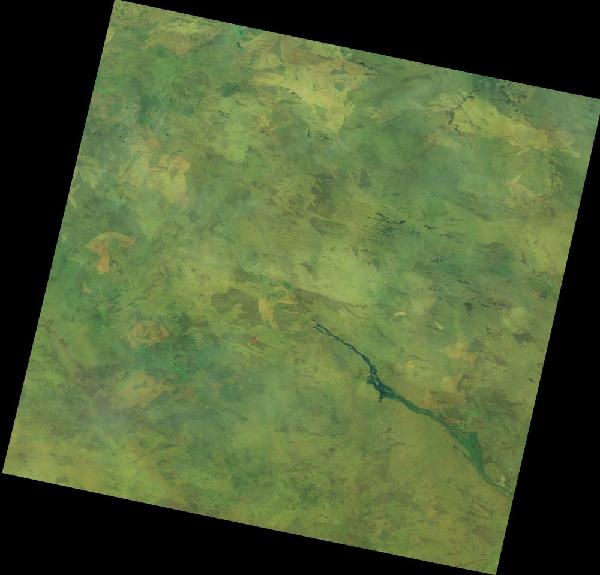}%
		\caption{2015-01-21 (LC8)}
		\label{fig:lc8-20150121}%
	\end{subfigure}
	\begin{subfigure}[t]{0.30\columnwidth}
		\centering
		\includegraphics[width=\textwidth]{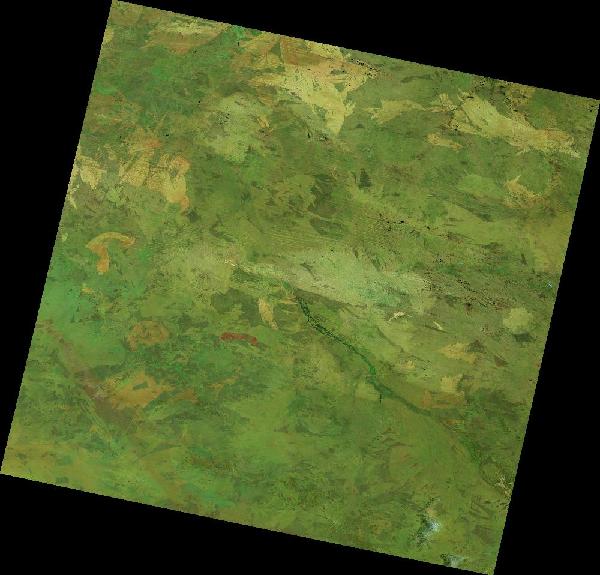}%
		\caption{2015-02-06 (LC8)}
		\label{fig:lc8-20150206}%
	\end{subfigure}
	\begin{subfigure}[t]{0.30\columnwidth}
		\centering
		\includegraphics[width=\textwidth]{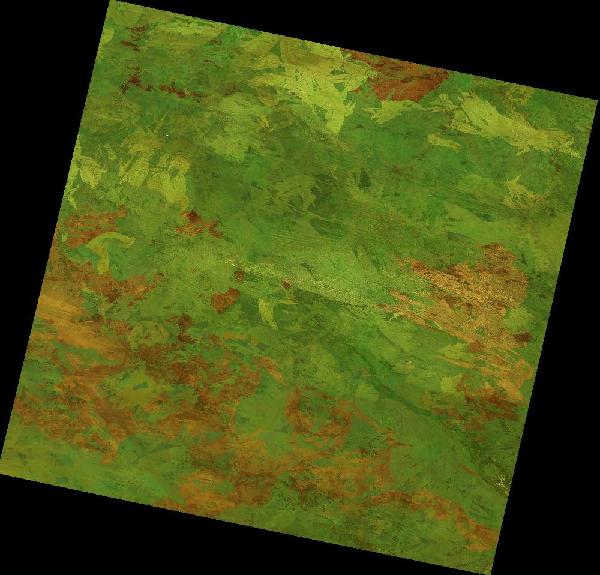}%
		\caption{2015-06-30 (LC8)}
		\label{fig:lc8-20150630}%
	\end{subfigure}
	\begin{subfigure}[t]{0.30\columnwidth}
		\centering
		\includegraphics[width=\textwidth]{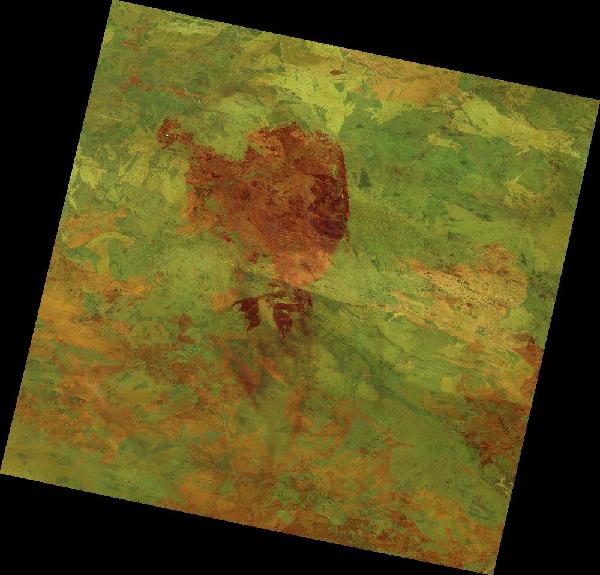}%
		\caption{2015-10-20 (LC8)}
		\label{fig:lc8-20151020}%
	\end{subfigure}
	\begin{subfigure}[t]{0.30\columnwidth}
		\centering
		\includegraphics[width=\textwidth]{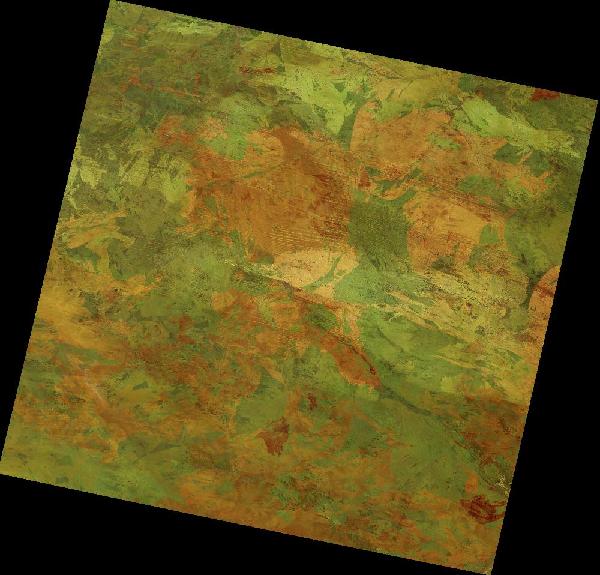}%
		\caption{2015-11-21 (LC8)}
		\label{fig:lc8-20151121}%
	\end{subfigure}
	\begin{subfigure}[t]{0.30\columnwidth}
		\centering
		\includegraphics[width=\textwidth]{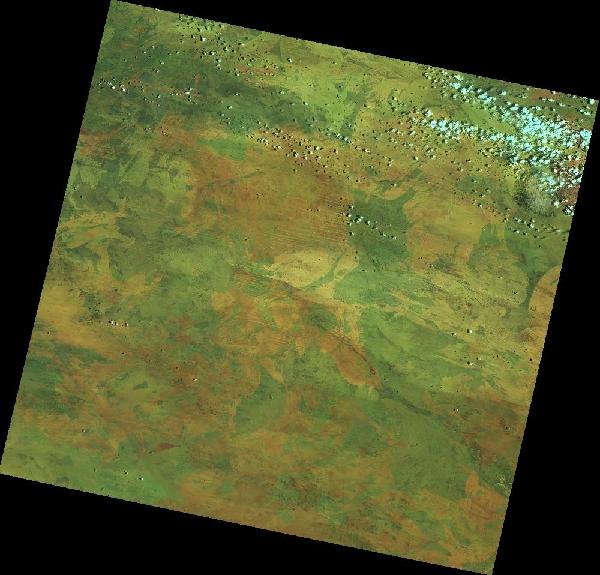}%
		\caption{2015-12-23 (LC8)}
		\label{fig:lc8-20151223}%
	\end{subfigure}
	\begin{subfigure}[t]{0.30\columnwidth}
		\centering
		\includegraphics[width=\textwidth]{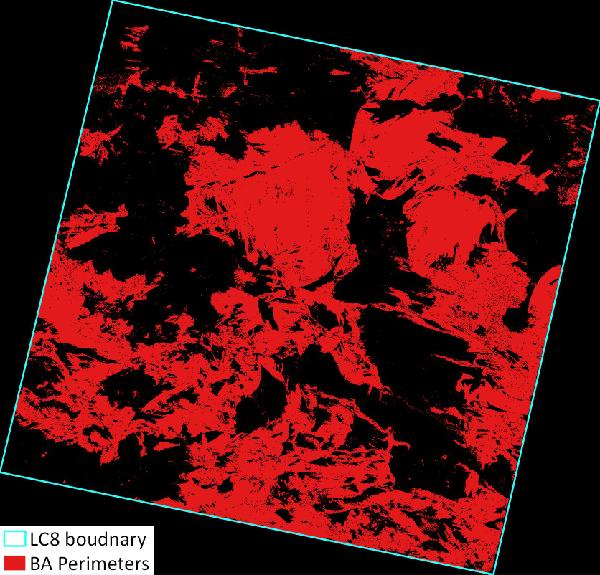}%
		\caption{BA (LC8)}
		\label{fig:lc8-australia-reference}%
	\end{subfigure}
	\begin{subfigure}[t]{0.30\columnwidth}
		\centering
		\includegraphics[width=\textwidth]{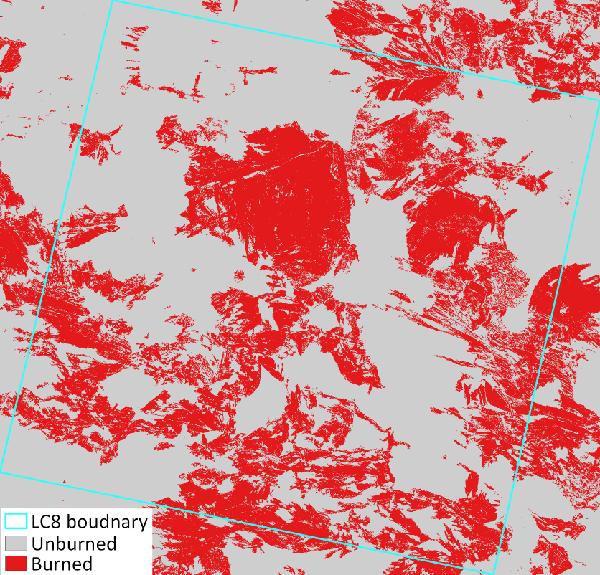}%
		\caption{Detected BA}
		\label{fig:lc8-australia-gabam}%
	\end{subfigure}
	\caption[]{
		Example of validation using Landsat-8 images (path/row:104/074) in Australia. \ref{fig:lc8-20141219}--\ref{fig:lc8-20151222} show the Landsat-8 images used to generate reference map, displayed in false color composition (red: SWIR2 band, green: NIR band and blue: GREEN band), \ref{fig:lc8-africa-reference} is reference BA map generated from Landsat-8 images, and~\ref{fig:lc8-australia-gabam} is detected BA by proposed method.
	}\label{fig:lc8-australia}%
\end{figure}

\begin{figure}[H]%
	\centering
	\begin{subfigure}[t]{0.32\columnwidth}
		\centering
		\includegraphics[width=\textwidth]{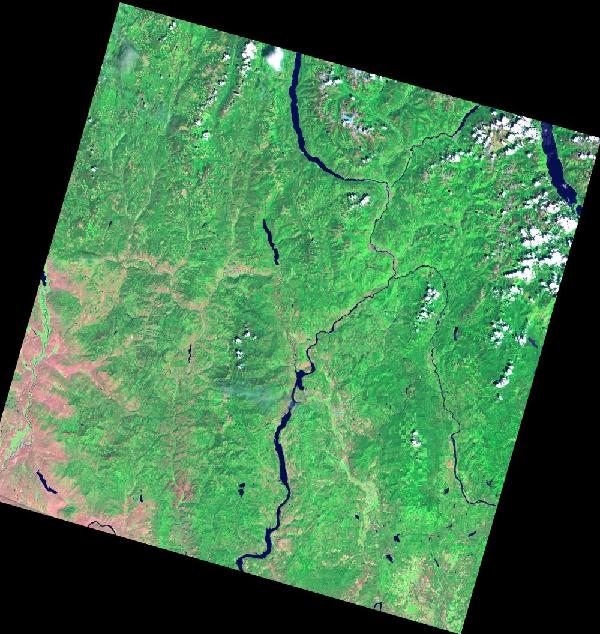}%
		\caption{2015-06-25 (LC8)}
		\label{fig:mtbs-landsat-20150625}%
	\end{subfigure}
	\begin{subfigure}[t]{0.32\columnwidth}
		\centering
		\includegraphics[width=\textwidth]{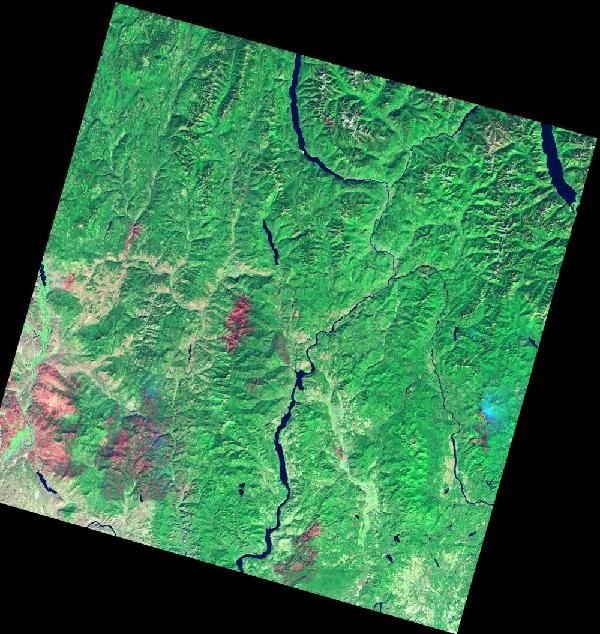}%
		\caption{2015-09-29 (LC8)}
		\label{fig:mtbs-landsat-20150929}%
	\end{subfigure}\\
	\begin{subfigure}[t]{0.32\columnwidth}
		\centering
		\includegraphics[width=\textwidth]{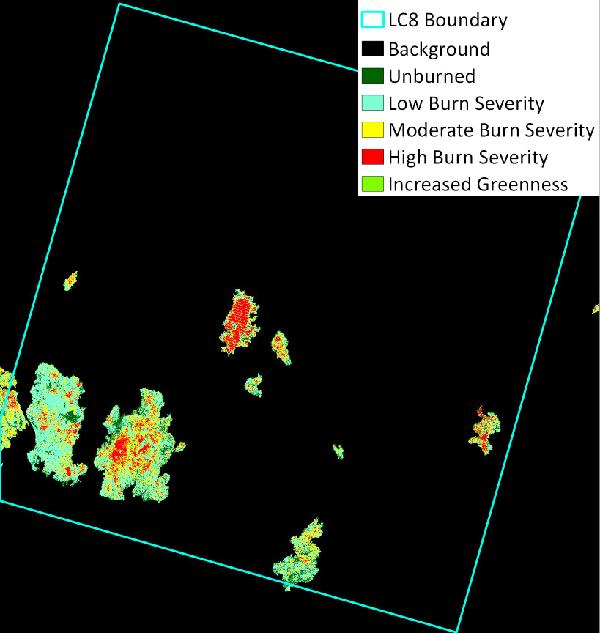}%
		\caption{MTBS BA perimeters}
		\label{fig:mtbs-mtbs}%
	\end{subfigure}
	\begin{subfigure}[t]{0.32\columnwidth}
		\centering
		\includegraphics[width=\textwidth]{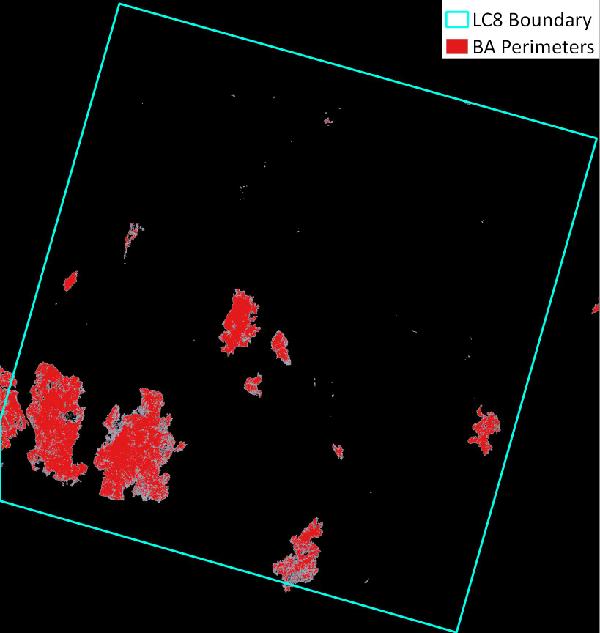}%
		\caption{Reference BA perimeters}
		\label{fig:mtbs-reference}%
	\end{subfigure}
	\begin{subfigure}[t]{0.32\columnwidth}
		\centering
		\includegraphics[width=\textwidth]{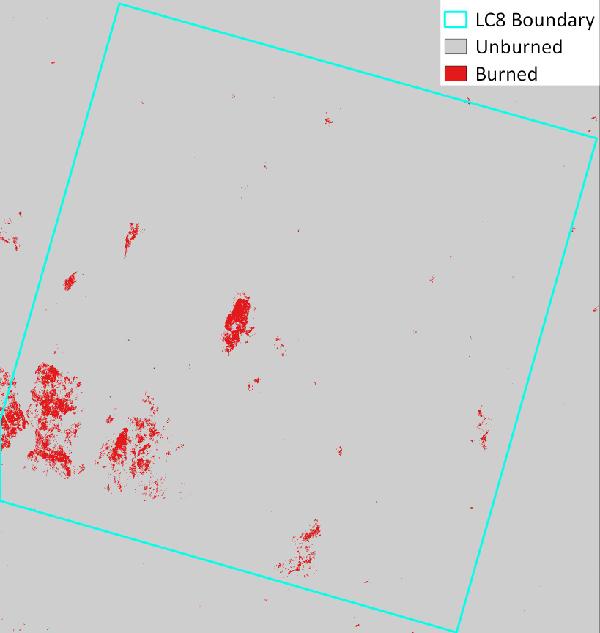}%
		\caption{Detected BA}
		\label{fig:mtbs-gabam}%
	\end{subfigure}
	\caption[]{
		Comparison between MTBS and detected BA. ~\ref{fig:mtbs-landsat-20150625} and ~\ref{fig:mtbs-landsat-20150929} are the Landsat-8 images (path/row:044/026) displayed in false color composition (red: SWIR2 band, green: NIR band and blue: GREEN band), \ref{fig:mtbs-mtbs} is the MTBS perimeters of 2015, \ref{fig:mtbs-reference} shows reference BA perimeters generated from Landsat-8 images and MTBS perimeters of 2015, and \ref{fig:mtbs-gabam} shows burned areas generated by proposed method.
	}\label{fig:mtbs}%
\end{figure}

Figure~\ref{fig:GF1}--\ref{fig:mtbs} show some examples of site validation, and Table~\ref{tbl:validation-show} summarizes the information of these validation sites, including the location, source of reference data, commission error, omission error and overall accuracy.
\begin{table}[!ht]
	\caption{Information of site validation examples.}
	\label{tbl:validation-show}
	\centering
	\begin{threeparttable}
		\begin{tabular}{cccrrrc}
			\toprule
			\textbf{ID} & \textbf{Location} & \textbf{Reference data}  & \textbf{$E_c$ (\%)} & \textbf{$E_o$ (\%)} & \textbf{$A_o$ (\%)} & \textbf{Figure}\\
			\midrule
			1 & China       & GF1 & 8.64 & 10.95 & 90.99 & Figure~\ref{fig:GF1}\\
			2 & South America & CB4 & 13.95 & 33.25 & 94.88 & Figure~\ref{fig:CB4}\\
			3 & Africa & LC8 & 41.23 & 57.41 & 71.29 & Figure~\ref{fig:lc8-africa}\\
			4 & Australia & LC8 & 0.77 & 20.88 & 90.22 & Figure~\ref{fig:lc8-australia}\\
			5 & U.S.       & LC8 \& MTBS & 1.45 & 67.97& 95.87 & Figure~\ref{fig:mtbs}\\
			\bottomrule
		\end{tabular}
	\end{threeparttable}
\end{table}

\section*{Reference}
\bibliography{ref}

\begin{thebibliography}{54}
\expandafter\ifx\csname natexlab\endcsname\relax\def\natexlab#1{#1}\fi
\providecommand{\url}[1]{\texttt{#1}}
\providecommand{\href}[2]{#2}
\providecommand{\path}[1]{#1}
\providecommand{\DOIprefix}{doi:}
\providecommand{\ArXivprefix}{arXiv:}
\providecommand{\URLprefix}{URL: }
\providecommand{\Pubmedprefix}{pmid:}
\providecommand{\doi}[1]{\href{http://dx.doi.org/#1}{\path{#1}}}
\providecommand{\Pubmed}[1]{\href{pmid:#1}{\path{#1}}}
\providecommand{\bibinfo}[2]{#2}
\ifx\xfnm\relax \def\xfnm[#1]{\unskip,\space#1}\fi
\bibitem[{Alonso-Canas and Chuvieco(2015)}]{Alonso_Canas_2015}
\bibinfo{author}{Alonso-Canas, I.}, \bibinfo{author}{Chuvieco, E.},
  \bibinfo{year}{2015}.
\newblock \bibinfo{title}{Global burned area mapping from {ENVISAT}-{MERIS} and
  {MODIS} active fire data}.
\newblock \bibinfo{journal}{Remote Sensing of Environment}
  \bibinfo{volume}{163}, \bibinfo{pages}{140--152}.
\newblock \URLprefix \url{https://doi.org/10.1016%2Fj.rse.2015.03.011},
  \DOIprefix\doi{10.1016/j.rse.2015.03.011}.
\bibitem[{Bastarrika et~al.(2014)Bastarrika, Alvarado, Artano, Martinez,
  Mesanza, Torre, Ramo and Chuvieco}]{Bastarrika_2014}
\bibinfo{author}{Bastarrika, A.}, \bibinfo{author}{Alvarado, M.},
  \bibinfo{author}{Artano, K.}, \bibinfo{author}{Martinez, M.},
  \bibinfo{author}{Mesanza, A.}, \bibinfo{author}{Torre, L.},
  \bibinfo{author}{Ramo, R.}, \bibinfo{author}{Chuvieco, E.},
  \bibinfo{year}{2014}.
\newblock \bibinfo{title}{{BAMS}: A tool for supervised burned area mapping
  using landsat data}.
\newblock \bibinfo{journal}{Remote Sensing} \bibinfo{volume}{6},
  \bibinfo{pages}{12360--12380}.
\newblock \URLprefix \url{https://doi.org/10.3390%2Frs61212360},
  \DOIprefix\doi{10.3390/rs61212360}.
\bibitem[{Bastarrika et~al.(2011)Bastarrika, Chuvieco and
  Mart{\'{\i}}n}]{Bastarrika_2011}
\bibinfo{author}{Bastarrika, A.}, \bibinfo{author}{Chuvieco, E.},
  \bibinfo{author}{Mart{\'{\i}}n, M.P.}, \bibinfo{year}{2011}.
\newblock \bibinfo{title}{Mapping burned areas from landsat {TM}/{ETM+} data
  with a two-phase algorithm: Balancing omission and commission errors}.
\newblock \bibinfo{journal}{Remote Sensing of Environment}
  \bibinfo{volume}{115}, \bibinfo{pages}{1003--1012}.
\newblock \URLprefix \url{https://doi.org/10.1016%2Fj.rse.2010.12.005},
  \DOIprefix\doi{10.1016/j.rse.2010.12.005}.
\bibitem[{Boschetti et~al.(2009)Boschetti, Roy and
  Justice}]{boschetti2009international}
\bibinfo{author}{Boschetti, L.}, \bibinfo{author}{Roy, D.},
  \bibinfo{author}{Justice, C.}, \bibinfo{year}{2009}.
\newblock \bibinfo{title}{International global burned area satellite product
  validation protocol (part i--production and standardization of validation
  reference data)}, in: \bibinfo{booktitle}{CEOS-CalVal, (Ed.)}.
  \bibinfo{publisher}{USA: Committee on Earth Observation Satellites}, pp.
  \bibinfo{pages}{1--11}.
\bibitem[{Boschetti et~al.(2015)Boschetti, Roy, Justice and
  Humber}]{Boschetti_2015}
\bibinfo{author}{Boschetti, L.}, \bibinfo{author}{Roy, D.P.},
  \bibinfo{author}{Justice, C.O.}, \bibinfo{author}{Humber, M.L.},
  \bibinfo{year}{2015}.
\newblock \bibinfo{title}{{MODIS}{\textendash}landsat fusion for large area 30m
  burned area mapping}.
\newblock \bibinfo{journal}{Remote Sensing of Environment}
  \bibinfo{volume}{161}, \bibinfo{pages}{27--42}.
\newblock \URLprefix \url{https://doi.org/10.1016%2Fj.rse.2015.01.022},
  \DOIprefix\doi{10.1016/j.rse.2015.01.022}.
\bibitem[{Boschetti et~al.(2016)Boschetti, Stehman and Roy}]{Boschetti_2016}
\bibinfo{author}{Boschetti, L.}, \bibinfo{author}{Stehman, S.V.},
  \bibinfo{author}{Roy, D.P.}, \bibinfo{year}{2016}.
\newblock \bibinfo{title}{A stratified random sampling design in space and time
  for regional to global scale burned area product validation}.
\newblock \bibinfo{journal}{Remote Sensing of Environment}
  \bibinfo{volume}{186}, \bibinfo{pages}{465--478}.
\newblock \URLprefix \url{https://doi.org/10.1016%2Fj.rse.2016.09.016},
  \DOIprefix\doi{10.1016/j.rse.2016.09.016}.
\bibitem[{Boschetti et~al.(2010)Boschetti, Stroppiana and
  Brivio}]{Boschetti_2010}
\bibinfo{author}{Boschetti, M.}, \bibinfo{author}{Stroppiana, D.},
  \bibinfo{author}{Brivio, P.A.}, \bibinfo{year}{2010}.
\newblock \bibinfo{title}{Mapping burned areas in a mediterranean environment
  using soft integration of spectral indices from high-resolution satellite
  images}.
\newblock \bibinfo{journal}{Earth Interactions} \bibinfo{volume}{14},
  \bibinfo{pages}{1--20}.
\newblock \URLprefix \url{https://doi.org/10.1175%2F2010ei349.1},
  \DOIprefix\doi{10.1175/2010ei349.1}.
\bibitem[{Carmona-Moreno et~al.(2005)Carmona-Moreno, Belward, Malingreau,
  Hartley, Garcia-Alegre, Antonovskiy, Buchshtaber and
  Pivovarov}]{Carmona_Moreno_2005}
\bibinfo{author}{Carmona-Moreno, C.}, \bibinfo{author}{Belward, A.},
  \bibinfo{author}{Malingreau, J.P.}, \bibinfo{author}{Hartley, A.},
  \bibinfo{author}{Garcia-Alegre, M.}, \bibinfo{author}{Antonovskiy, M.},
  \bibinfo{author}{Buchshtaber, V.}, \bibinfo{author}{Pivovarov, V.},
  \bibinfo{year}{2005}.
\newblock \bibinfo{title}{Characterizing interannual variations in global fire
  calendar using data from earth observing satellites}.
\newblock \bibinfo{journal}{Global Change Biology} \bibinfo{volume}{11},
  \bibinfo{pages}{1537--1555}.
\newblock \URLprefix \url{https://doi.org/10.1111%2Fj.1365-2486.2005.01003.x},
  \DOIprefix\doi{10.1111/j.1365-2486.2005.01003.x}.
\bibitem[{Chuvieco et~al.(2011)Chuvieco, Padilla, Hantson, Theis and
  Snadow}]{chuvieco2011esa}
\bibinfo{author}{Chuvieco, E.}, \bibinfo{author}{Padilla, M.},
  \bibinfo{author}{Hantson, S.}, \bibinfo{author}{Theis, R.},
  \bibinfo{author}{Snadow, C.}, \bibinfo{year}{2011}.
\newblock \bibinfo{title}{Esa cci ecv fire disturbance-product validation plan
  (v3. 1)}.
\newblock \bibinfo{journal}{ESA Fire-CCI project (http://www. esa-fire-cci.
  org/)} .
\bibitem[{Chuvieco et~al.(2016)Chuvieco, Yue, Heil, Mouillot, Alonso-Canas,
  Padilla, Pereira, Oom and Tansey}]{Chuvieco_2016}
\bibinfo{author}{Chuvieco, E.}, \bibinfo{author}{Yue, C.},
  \bibinfo{author}{Heil, A.}, \bibinfo{author}{Mouillot, F.},
  \bibinfo{author}{Alonso-Canas, I.}, \bibinfo{author}{Padilla, M.},
  \bibinfo{author}{Pereira, J.M.}, \bibinfo{author}{Oom, D.},
  \bibinfo{author}{Tansey, K.}, \bibinfo{year}{2016}.
\newblock \bibinfo{title}{A new global burned area product for climate
  assessment of fire impacts}.
\newblock \bibinfo{journal}{Global Ecology and Biogeography}
  \bibinfo{volume}{25}, \bibinfo{pages}{619--629}.
\newblock \URLprefix \url{https://doi.org/10.1111%2Fgeb.12440},
  \DOIprefix\doi{10.1111/geb.12440}.
\bibitem[{DAAC(2015)}]{NASA_2015}
\bibinfo{author}{DAAC, N.L.}, \bibinfo{year}{2015}.
\newblock \bibinfo{title}{Modis vegetation continuous fields (vcf) product.
  version 5.1}.
\newblock
  \bibinfo{howpublished}{\url{https://lpdaac.usgs.gov/dataset_discovery/modis/modis_products_table/mod44b}}.
\newblock \DOIprefix\doi{10.4225/13/511C71F8612C3}. \bibinfo{note}{nASA EOSDIS
  Land Processes DAAC, USGS Earth Resources Observation and Science (EROS)
  Center, Sioux Falls, South Dakota (https://lpdaac.usgs.gov)}.
\bibitem[{Eidenshink et~al.(2007)Eidenshink, Schwind, Brewer, Zhu, Quayle and
  Howard}]{Eidenshink_2007}
\bibinfo{author}{Eidenshink, J.}, \bibinfo{author}{Schwind, B.},
  \bibinfo{author}{Brewer, K.}, \bibinfo{author}{Zhu, Z.L.},
  \bibinfo{author}{Quayle, B.}, \bibinfo{author}{Howard, S.},
  \bibinfo{year}{2007}.
\newblock \bibinfo{title}{A project for monitoring trends in burn severity}.
\newblock \bibinfo{journal}{Fire Ecology} \bibinfo{volume}{3},
  \bibinfo{pages}{3--21}.
\newblock \URLprefix \url{https://doi.org/10.4996%2Ffireecology.0301003},
  \DOIprefix\doi{10.4996/fireecology.0301003}.
\bibitem[{Friedl and Sulla-Menashe(2015)}]{MCD12C1_2015}
\bibinfo{author}{Friedl, M.}, \bibinfo{author}{Sulla-Menashe, D.},
  \bibinfo{year}{2015}.
\newblock \bibinfo{title}{Mcd12c1 modis/terra+aqua land cover type yearly l3
  global 0.05deg cmg}.
\newblock
  \bibinfo{howpublished}{\url{https://lpdaac.usgs.gov/dataset_discovery/modis/modis_products_table/mcd12c1}}.
\newblock \DOIprefix\doi{10.5067/MODIS/MCD12C1.006}. \bibinfo{note}{nASA EOSDIS
  Land Processes DAAC (https://lpdaac.usgs.gov)}.
\bibitem[{Giglio et~al.(2013)Giglio, Randerson and van~der Werf}]{Giglio_2013}
\bibinfo{author}{Giglio, L.}, \bibinfo{author}{Randerson, J.T.},
  \bibinfo{author}{van~der Werf, G.R.}, \bibinfo{year}{2013}.
\newblock \bibinfo{title}{Analysis of daily, monthly, and annual burned area
  using the fourth-generation global fire emissions database ({GFED}4)}.
\newblock \bibinfo{journal}{Journal of Geophysical Research: Biogeosciences}
  \bibinfo{volume}{118}, \bibinfo{pages}{317--328}.
\newblock \URLprefix \url{https://doi.org/10.1002%2Fjgrg.20042},
  \DOIprefix\doi{10.1002/jgrg.20042}.
\bibitem[{Giglio et~al.(2010)Giglio, Randerson, van~der Werf, Kasibhatla,
  Collatz, Morton and DeFries}]{Giglio_2010}
\bibinfo{author}{Giglio, L.}, \bibinfo{author}{Randerson, J.T.},
  \bibinfo{author}{van~der Werf, G.R.}, \bibinfo{author}{Kasibhatla, P.S.},
  \bibinfo{author}{Collatz, G.J.}, \bibinfo{author}{Morton, D.C.},
  \bibinfo{author}{DeFries, R.S.}, \bibinfo{year}{2010}.
\newblock \bibinfo{title}{Assessing variability and long-term trends in burned
  area by merging multiple satellite fire products}.
\newblock \bibinfo{journal}{Biogeosciences} \bibinfo{volume}{7},
  \bibinfo{pages}{1171--1186}.
\newblock \URLprefix \url{https://doi.org/10.5194%2Fbg-7-1171-2010},
  \DOIprefix\doi{10.5194/bg-7-1171-2010}.
\bibitem[{Giglio et~al.(2016)Giglio, Schroeder and Justice}]{Giglio_2016}
\bibinfo{author}{Giglio, L.}, \bibinfo{author}{Schroeder, W.},
  \bibinfo{author}{Justice, C.O.}, \bibinfo{year}{2016}.
\newblock \bibinfo{title}{The collection 6 {MODIS} active fire detection
  algorithm and fire products}.
\newblock \bibinfo{journal}{Remote Sensing of Environment}
  \bibinfo{volume}{178}, \bibinfo{pages}{31--41}.
\newblock \URLprefix \url{https://doi.org/10.1016%2Fj.rse.2016.02.054},
  \DOIprefix\doi{10.1016/j.rse.2016.02.054}.
\bibitem[{Goodwin and Collett(2014)}]{Goodwin_2014}
\bibinfo{author}{Goodwin, N.R.}, \bibinfo{author}{Collett, L.J.},
  \bibinfo{year}{2014}.
\newblock \bibinfo{title}{Development of an automated method for mapping fire
  history captured in landsat {TM} and {ETM+} time series across queensland,
  australia}.
\newblock \bibinfo{journal}{Remote Sensing of Environment}
  \bibinfo{volume}{148}, \bibinfo{pages}{206--221}.
\newblock \URLprefix \url{https://doi.org/10.1016%2Fj.rse.2014.03.021},
  \DOIprefix\doi{10.1016/j.rse.2014.03.021}.
\bibitem[{Gorelick et~al.(2017)Gorelick, Hancher, Dixon, Ilyushchenko, Thau and
  Moore}]{Gorelick_2017}
\bibinfo{author}{Gorelick, N.}, \bibinfo{author}{Hancher, M.},
  \bibinfo{author}{Dixon, M.}, \bibinfo{author}{Ilyushchenko, S.},
  \bibinfo{author}{Thau, D.}, \bibinfo{author}{Moore, R.},
  \bibinfo{year}{2017}.
\newblock \bibinfo{title}{Google earth engine: Planetary-scale geospatial
  analysis for everyone}.
\newblock \bibinfo{journal}{Remote Sensing of Environment}
  \bibinfo{volume}{202}, \bibinfo{pages}{18--27}.
\newblock \URLprefix \url{https://doi.org/10.1016%2Fj.rse.2017.06.031},
  \DOIprefix\doi{10.1016/j.rse.2017.06.031}.
\bibitem[{Hawbaker et~al.(2017a)Hawbaker, Vanderhoof, Beal, Takacs, Schmidt,
  Falgout, Williams, Fairaux, Caldwell, Picotte, Howard, Stitt and
  Dwyer}]{Hawbaker_2017}
\bibinfo{author}{Hawbaker, T.J.}, \bibinfo{author}{Vanderhoof, M.K.},
  \bibinfo{author}{Beal, Y.J.}, \bibinfo{author}{Takacs, J.D.},
  \bibinfo{author}{Schmidt, G.L.}, \bibinfo{author}{Falgout, J.T.},
  \bibinfo{author}{Williams, B.}, \bibinfo{author}{Fairaux, N.M.},
  \bibinfo{author}{Caldwell, M.K.}, \bibinfo{author}{Picotte, J.J.},
  \bibinfo{author}{Howard, S.M.}, \bibinfo{author}{Stitt, S.},
  \bibinfo{author}{Dwyer, J.L.}, \bibinfo{year}{2017}a.
\newblock \bibinfo{title}{Landsat burned area essential climate variable
  products for the conterminous united states (1984 - 2015)}.
\newblock \bibinfo{journal}{Remote Sensing of Environment}
  \bibinfo{volume}{198}, \bibinfo{pages}{504--522}.
\newblock \URLprefix \url{https://doi.org/10.5066/F73B5X76},
  \DOIprefix\doi{10.1016/j.rse.2017.06.027}.
\bibitem[{Hawbaker et~al.(2017b)Hawbaker, Vanderhoof, Beal, Takacs, Schmidt,
  Falgout, Williams, Fairaux, Caldwell, Picotte, Howard, Stitt and
  Dwyer}]{Hawbaker_2017_dense}
\bibinfo{author}{Hawbaker, T.J.}, \bibinfo{author}{Vanderhoof, M.K.},
  \bibinfo{author}{Beal, Y.J.}, \bibinfo{author}{Takacs, J.D.},
  \bibinfo{author}{Schmidt, G.L.}, \bibinfo{author}{Falgout, J.T.},
  \bibinfo{author}{Williams, B.}, \bibinfo{author}{Fairaux, N.M.},
  \bibinfo{author}{Caldwell, M.K.}, \bibinfo{author}{Picotte, J.J.},
  \bibinfo{author}{Howard, S.M.}, \bibinfo{author}{Stitt, S.},
  \bibinfo{author}{Dwyer, J.L.}, \bibinfo{year}{2017}b.
\newblock \bibinfo{title}{Mapping burned areas using dense time-series of
  landsat data}.
\newblock \bibinfo{journal}{Remote Sensing of Environment}
  \bibinfo{volume}{198}, \bibinfo{pages}{504--522}.
\newblock \URLprefix \url{https://doi.org/10.1016%2Fj.rse.2017.06.027},
  \DOIprefix\doi{10.1016/j.rse.2017.06.027}.
\bibitem[{Huete(1988)}]{Huete_1988}
\bibinfo{author}{Huete, A.}, \bibinfo{year}{1988}.
\newblock \bibinfo{title}{A soil-adjusted vegetation index ({SAVI})}.
\newblock \bibinfo{journal}{Remote Sensing of Environment}
  \bibinfo{volume}{25}, \bibinfo{pages}{295--309}.
\newblock \URLprefix \url{https://doi.org/10.1016%2F0034-4257%2888%2990106-x},
  \DOIprefix\doi{10.1016/0034-4257(88)90106-x}.
\bibitem[{Key and Benson(1999)}]{key1999normalized}
\bibinfo{author}{Key, C.H.}, \bibinfo{author}{Benson, N.C.},
  \bibinfo{year}{1999}.
\newblock \bibinfo{title}{The normalized burn ratio (nbr): A landsat tm
  radiometric measure of burn severity}.
\newblock \bibinfo{journal}{United States Geological Survey, Northern Rocky
  Mountain Science Center.(Bozeman, MT)} .
\bibitem[{Koutsias and Karteris(2000)}]{Koutsias_2000}
\bibinfo{author}{Koutsias, N.}, \bibinfo{author}{Karteris, M.},
  \bibinfo{year}{2000}.
\newblock \bibinfo{title}{Burned area mapping using logistic regression
  modeling of a single post-fire landsat-5 thematic mapper image}.
\newblock \bibinfo{journal}{International Journal of Remote Sensing}
  \bibinfo{volume}{21}, \bibinfo{pages}{673--687}.
\newblock \URLprefix \url{https://doi.org/10.1080%2F014311600210506},
  \DOIprefix\doi{10.1080/014311600210506}.
\bibitem[{Laris(2005)}]{Laris_2005}
\bibinfo{author}{Laris, P.S.}, \bibinfo{year}{2005}.
\newblock \bibinfo{title}{Spatiotemporal problems with detecting and mapping
  mosaic fire regimes with coarse-resolution satellite data in savanna
  environments}.
\newblock \bibinfo{journal}{Remote Sensing of Environment}
  \bibinfo{volume}{99}, \bibinfo{pages}{412--424}.
\newblock \URLprefix \url{https://doi.org/10.1016%2Fj.rse.2005.09.012},
  \DOIprefix\doi{10.1016/j.rse.2005.09.012}.
\bibitem[{Lhermitte et~al.(2011)Lhermitte, Verbesselt, Verstraeten, Veraverbeke
  and Coppin}]{Lhermitte_2011}
\bibinfo{author}{Lhermitte, S.}, \bibinfo{author}{Verbesselt, J.},
  \bibinfo{author}{Verstraeten, W.}, \bibinfo{author}{Veraverbeke, S.},
  \bibinfo{author}{Coppin, P.}, \bibinfo{year}{2011}.
\newblock \bibinfo{title}{Assessing intra-annual vegetation regrowth after fire
  using the pixel based regeneration index}.
\newblock \bibinfo{journal}{{ISPRS} Journal of Photogrammetry and Remote
  Sensing} \bibinfo{volume}{66}, \bibinfo{pages}{17--27}.
\newblock \URLprefix \url{https://doi.org/10.1016%2Fj.isprsjprs.2010.08.004},
  \DOIprefix\doi{10.1016/j.isprsjprs.2010.08.004}.
\bibitem[{Liu et~al.(2018)Liu, Heiskanen, Maeda and Pellikka}]{Liu_2018}
\bibinfo{author}{Liu, J.}, \bibinfo{author}{Heiskanen, J.},
  \bibinfo{author}{Maeda, E.E.}, \bibinfo{author}{Pellikka, P.K.},
  \bibinfo{year}{2018}.
\newblock \bibinfo{title}{Burned area detection based on landsat time series in
  savannas of southern burkina faso}.
\newblock \bibinfo{journal}{International Journal of Applied Earth Observation
  and Geoinformation} \bibinfo{volume}{64}, \bibinfo{pages}{210--220}.
\newblock \URLprefix \url{https://doi.org/10.1016%2Fj.jag.2017.09.011},
  \DOIprefix\doi{10.1016/j.jag.2017.09.011}.
\bibitem[{Long et~al.(2016)Long, Jiao, He and Zhang}]{Long_2016}
\bibinfo{author}{Long, T.}, \bibinfo{author}{Jiao, W.}, \bibinfo{author}{He,
  G.}, \bibinfo{author}{Zhang, Z.}, \bibinfo{year}{2016}.
\newblock \bibinfo{title}{A fast and reliable matching method for automated
  georeferencing of remotely-sensed imagery}.
\newblock \bibinfo{journal}{Remote Sensing} \bibinfo{volume}{8},
  \bibinfo{pages}{56}.
\newblock \URLprefix \url{https://doi.org/10.3390%2Frs8010056},
  \DOIprefix\doi{10.3390/rs8010056}.
\bibitem[{Lutes et~al.(2006)Lutes, Keane, Caratti, Key, Benson, Sutherland,
  Gangi et~al.}]{lutes2006firemon}
\bibinfo{author}{Lutes, D.C.}, \bibinfo{author}{Keane, R.E.},
  \bibinfo{author}{Caratti, J.F.}, \bibinfo{author}{Key, C.H.},
  \bibinfo{author}{Benson, N.C.}, \bibinfo{author}{Sutherland, S.},
  \bibinfo{author}{Gangi, L.J.}, et~al., \bibinfo{year}{2006}.
\newblock \bibinfo{title}{Firemon: Fire effects monitoring and inventory
  system}.
\newblock \bibinfo{journal}{Gen. Tech. Rep. RMRS-GTR-164-CD. Fort Collins, CO:
  US Department of Agriculture, Forest Service, Rocky Mountain Research
  Station} \bibinfo{volume}{1}.
\bibitem[{Mart{\'\i}n(1998)}]{martin1998cartografia}
\bibinfo{author}{Mart{\'\i}n, M.}, \bibinfo{year}{1998}.
\newblock \bibinfo{title}{Cartograf{\'\i}a e inventario de incendios forestales
  en la pen{\'\i}nsula ib{\'e}rica a partir de im{\'a}genes noaa-avhrr}.
\newblock \bibinfo{journal}{Departmento de Geograf{\'\i}a. Alcal{\'a} de
  Henares, Universidad de Alcal{\'a}} .
\bibitem[{Miller and Thode(2007)}]{Miller_2007}
\bibinfo{author}{Miller, J.D.}, \bibinfo{author}{Thode, A.E.},
  \bibinfo{year}{2007}.
\newblock \bibinfo{title}{Quantifying burn severity in a heterogeneous
  landscape with a relative version of the delta normalized burn ratio
  ({dNBR})}.
\newblock \bibinfo{journal}{Remote Sensing of Environment}
  \bibinfo{volume}{109}, \bibinfo{pages}{66--80}.
\newblock \URLprefix \url{https://doi.org/10.1016%2Fj.rse.2006.12.006},
  \DOIprefix\doi{10.1016/j.rse.2006.12.006}.
\bibitem[{Padilla et~al.(2017)Padilla, Olofsson, Stehman, Tansey and
  Chuvieco}]{Padilla_2017}
\bibinfo{author}{Padilla, M.}, \bibinfo{author}{Olofsson, P.},
  \bibinfo{author}{Stehman, S.V.}, \bibinfo{author}{Tansey, K.},
  \bibinfo{author}{Chuvieco, E.}, \bibinfo{year}{2017}.
\newblock \bibinfo{title}{Stratification and sample allocation for reference
  burned area data}.
\newblock \bibinfo{journal}{Remote Sensing of Environment}
  \bibinfo{volume}{203}, \bibinfo{pages}{240--255}.
\newblock \URLprefix \url{https://doi.org/10.1016%2Fj.rse.2017.06.041},
  \DOIprefix\doi{10.1016/j.rse.2017.06.041}.
\bibitem[{Padilla et~al.(2014)Padilla, Stehman and Chuvieco}]{Padilla_2014}
\bibinfo{author}{Padilla, M.}, \bibinfo{author}{Stehman, S.V.},
  \bibinfo{author}{Chuvieco, E.}, \bibinfo{year}{2014}.
\newblock \bibinfo{title}{Validation of the 2008 {MODIS}-{MCD}45 global burned
  area product using stratified random sampling}.
\newblock \bibinfo{journal}{Remote Sensing of Environment}
  \bibinfo{volume}{144}, \bibinfo{pages}{187--196}.
\newblock \URLprefix \url{https://doi.org/10.1016%2Fj.rse.2014.01.008},
  \DOIprefix\doi{10.1016/j.rse.2014.01.008}.
\bibitem[{Padilla et~al.(2015)Padilla, Stehman, Ramo, Corti, Hantson, Oliva,
  Alonso-Canas, Bradley, Tansey, Mota, Pereira and Chuvieco}]{Padilla_2015}
\bibinfo{author}{Padilla, M.}, \bibinfo{author}{Stehman, S.V.},
  \bibinfo{author}{Ramo, R.}, \bibinfo{author}{Corti, D.},
  \bibinfo{author}{Hantson, S.}, \bibinfo{author}{Oliva, P.},
  \bibinfo{author}{Alonso-Canas, I.}, \bibinfo{author}{Bradley, A.V.},
  \bibinfo{author}{Tansey, K.}, \bibinfo{author}{Mota, B.},
  \bibinfo{author}{Pereira, J.M.}, \bibinfo{author}{Chuvieco, E.},
  \bibinfo{year}{2015}.
\newblock \bibinfo{title}{Comparing the accuracies of remote sensing global
  burned area products using stratified random sampling and estimation}.
\newblock \bibinfo{journal}{Remote Sensing of Environment}
  \bibinfo{volume}{160}, \bibinfo{pages}{114--121}.
\newblock \URLprefix \url{https://doi.org/10.1016%2Fj.rse.2015.01.005},
  \DOIprefix\doi{10.1016/j.rse.2015.01.005}.
\bibitem[{Pedregosa et~al.(2011)Pedregosa, Varoquaux, Gramfort, Michel,
  Thirion, Grisel, Blondel, Prettenhofer, Weiss, Dubourg, Vanderplas, Passos,
  Cournapeau, Brucher, Perrot and Duchesnay}]{scikit-learn}
\bibinfo{author}{Pedregosa, F.}, \bibinfo{author}{Varoquaux, G.},
  \bibinfo{author}{Gramfort, A.}, \bibinfo{author}{Michel, V.},
  \bibinfo{author}{Thirion, B.}, \bibinfo{author}{Grisel, O.},
  \bibinfo{author}{Blondel, M.}, \bibinfo{author}{Prettenhofer, P.},
  \bibinfo{author}{Weiss, R.}, \bibinfo{author}{Dubourg, V.},
  \bibinfo{author}{Vanderplas, J.}, \bibinfo{author}{Passos, A.},
  \bibinfo{author}{Cournapeau, D.}, \bibinfo{author}{Brucher, M.},
  \bibinfo{author}{Perrot, M.}, \bibinfo{author}{Duchesnay, E.},
  \bibinfo{year}{2011}.
\newblock \bibinfo{title}{Scikit-learn: Machine learning in {P}ython}.
\newblock \bibinfo{journal}{Journal of Machine Learning Research}
  \bibinfo{volume}{12}, \bibinfo{pages}{2825--2830}.
\bibitem[{Pereira(1999)}]{Pereira_1999}
\bibinfo{author}{Pereira, J.}, \bibinfo{year}{1999}.
\newblock \bibinfo{title}{A comparative evaluation of {NOAA}/{AVHRR} vegetation
  indexes for burned surface detection and mapping}.
\newblock \bibinfo{journal}{{IEEE} Transactions on Geoscience and Remote
  Sensing} \bibinfo{volume}{37}, \bibinfo{pages}{217--226}.
\newblock \URLprefix \url{https://doi.org/10.1109%2F36.739156},
  \DOIprefix\doi{10.1109/36.739156}.
\bibitem[{Pettinari and Chuvieco(2018)}]{Pettinari2018}
\bibinfo{author}{Pettinari, M.}, \bibinfo{author}{Chuvieco, E.},
  \bibinfo{year}{2018}.
\newblock \bibinfo{title}{Esa cci ecv fire disturbance: D3.3.3 product user
  guide - modis, version 1.0}.
\newblock \bibinfo{journal}{ESA Fire-CCI project
  (http://www.esa-firecci.org/documents)} .
\bibitem[{Pinty and Verstraete(1992)}]{Pinty_1992}
\bibinfo{author}{Pinty, B.}, \bibinfo{author}{Verstraete, M.M.},
  \bibinfo{year}{1992}.
\newblock \bibinfo{title}{{GEMI}: a non-linear index to monitor global
  vegetation from satellites}.
\newblock \bibinfo{journal}{Vegetatio} \bibinfo{volume}{101},
  \bibinfo{pages}{15--20}.
\newblock \URLprefix \url{https://doi.org/10.1007%2Fbf00031911},
  \DOIprefix\doi{10.1007/bf00031911}.
\bibitem[{Pleniou and Koutsias(2013)}]{Pleniou_2013}
\bibinfo{author}{Pleniou, M.}, \bibinfo{author}{Koutsias, N.},
  \bibinfo{year}{2013}.
\newblock \bibinfo{title}{Sensitivity of spectral reflectance values to
  different burn and vegetation ratios: A multi-scale approach applied in a
  fire affected area}.
\newblock \bibinfo{journal}{{ISPRS} Journal of Photogrammetry and Remote
  Sensing} \bibinfo{volume}{79}, \bibinfo{pages}{199--210}.
\newblock \URLprefix \url{https://doi.org/10.1016%2Fj.isprsjprs.2013.02.016},
  \DOIprefix\doi{10.1016/j.isprsjprs.2013.02.016}.
\bibitem[{Plummer et~al.(2006)Plummer, Arino, Simon and Steffen}]{Plummer_2006}
\bibinfo{author}{Plummer, S.}, \bibinfo{author}{Arino, O.},
  \bibinfo{author}{Simon, M.}, \bibinfo{author}{Steffen, W.},
  \bibinfo{year}{2006}.
\newblock \bibinfo{title}{Establishing a earth observation product service for
  the terrestrial carbon community: The globcarbon initiative}.
\newblock \bibinfo{journal}{Mitigation and Adaptation Strategies for Global
  Change} \bibinfo{volume}{11}, \bibinfo{pages}{97--111}.
\newblock \URLprefix \url{https://doi.org/10.1007%2Fs11027-006-1012-8},
  \DOIprefix\doi{10.1007/s11027-006-1012-8}.
\bibitem[{Pontius and Millones(2011)}]{Pontius_2011}
\bibinfo{author}{Pontius, R.G.}, \bibinfo{author}{Millones, M.},
  \bibinfo{year}{2011}.
\newblock \bibinfo{title}{Death to kappa: birth of quantity disagreement and
  allocation disagreement for accuracy assessment}.
\newblock \bibinfo{journal}{International Journal of Remote Sensing}
  \bibinfo{volume}{32}, \bibinfo{pages}{4407--4429}.
\newblock \URLprefix \url{https://doi.org/10.1080%2F01431161.2011.552923},
  \DOIprefix\doi{10.1080/01431161.2011.552923}.
\bibitem[{Roy et~al.(2005)Roy, Jin, Lewis and Justice}]{Roy_2005}
\bibinfo{author}{Roy, D.}, \bibinfo{author}{Jin, Y.}, \bibinfo{author}{Lewis,
  P.}, \bibinfo{author}{Justice, C.}, \bibinfo{year}{2005}.
\newblock \bibinfo{title}{Prototyping a global algorithm for systematic
  fire-affected area mapping using {MODIS} time series data}.
\newblock \bibinfo{journal}{Remote Sensing of Environment}
  \bibinfo{volume}{97}, \bibinfo{pages}{137--162}.
\newblock \URLprefix \url{https://doi.org/10.1016%2Fj.rse.2005.04.007},
  \DOIprefix\doi{10.1016/j.rse.2005.04.007}.
\bibitem[{S.~Trigg(2001)}]{Flasse_2001}
\bibinfo{author}{S.~Trigg, S.F.}, \bibinfo{year}{2001}.
\newblock \bibinfo{title}{An evaluation of different bi-spectral spaces for
  discriminating burned shrub-savannah}.
\newblock \bibinfo{journal}{International Journal of Remote Sensing}
  \bibinfo{volume}{22}, \bibinfo{pages}{2641--2647}.
\newblock \URLprefix \url{https://doi.org/10.1080%2F01431160119380},
  \DOIprefix\doi{10.1080/01431160119380}.
\bibitem[{Simon(2004)}]{Simon_2004}
\bibinfo{author}{Simon, M.}, \bibinfo{year}{2004}.
\newblock \bibinfo{title}{Burnt area detection at global scale using {ATSR}-2:
  The {GLOBSCAR} products and their qualification}.
\newblock \bibinfo{journal}{Journal of Geophysical Research}
  \bibinfo{volume}{109}.
\newblock \URLprefix \url{https://doi.org/10.1029%2F2003jd003622},
  \DOIprefix\doi{10.1029/2003jd003622}.
\bibitem[{Sobrino and Raissouni(2000)}]{Sobrino_2000}
\bibinfo{author}{Sobrino, J.A.}, \bibinfo{author}{Raissouni, N.},
  \bibinfo{year}{2000}.
\newblock \bibinfo{title}{Toward remote sensing methods for land cover dynamic
  monitoring: Application to morocco}.
\newblock \bibinfo{journal}{International Journal of Remote Sensing}
  \bibinfo{volume}{21}, \bibinfo{pages}{353--366}.
\newblock \URLprefix \url{https://doi.org/10.1080%2F014311600210876},
  \DOIprefix\doi{10.1080/014311600210876}.
\bibitem[{Strahler et~al.(2006)Strahler, Boschetti, Foody, Friedl, Hansen,
  Herold, Mayaux, Morisette, Stehman and Woodcock}]{strahler2006global}
\bibinfo{author}{Strahler, A.H.}, \bibinfo{author}{Boschetti, L.},
  \bibinfo{author}{Foody, G.M.}, \bibinfo{author}{Friedl, M.A.},
  \bibinfo{author}{Hansen, M.C.}, \bibinfo{author}{Herold, M.},
  \bibinfo{author}{Mayaux, P.}, \bibinfo{author}{Morisette, J.T.},
  \bibinfo{author}{Stehman, S.V.}, \bibinfo{author}{Woodcock, C.E.},
  \bibinfo{year}{2006}.
\newblock \bibinfo{title}{Global land cover validation: Recommendations for
  evaluation and accuracy assessment of global land cover maps}.
\newblock \bibinfo{journal}{European Communities, Luxembourg}
  \bibinfo{volume}{51}.
\bibitem[{Stroppiana et~al.(2012)Stroppiana, Bordogna, Carrara, Boschetti,
  Boschetti and Brivio}]{Stroppiana_2012}
\bibinfo{author}{Stroppiana, D.}, \bibinfo{author}{Bordogna, G.},
  \bibinfo{author}{Carrara, P.}, \bibinfo{author}{Boschetti, M.},
  \bibinfo{author}{Boschetti, L.}, \bibinfo{author}{Brivio, P.},
  \bibinfo{year}{2012}.
\newblock \bibinfo{title}{A method for extracting burned areas from landsat
  {TM}/{ETM+} images by soft aggregation of multiple spectral indices and a
  region growing algorithm}.
\newblock \bibinfo{journal}{{ISPRS} Journal of Photogrammetry and Remote
  Sensing} \bibinfo{volume}{69}, \bibinfo{pages}{88--102}.
\newblock \URLprefix \url{https://doi.org/10.1016%2Fj.isprsjprs.2012.03.001},
  \DOIprefix\doi{10.1016/j.isprsjprs.2012.03.001}.
\bibitem[{Stroppiana et~al.(2009)Stroppiana, Boschetti, Zaffaroni and
  Brivio}]{Stroppiana_2009}
\bibinfo{author}{Stroppiana, D.}, \bibinfo{author}{Boschetti, M.},
  \bibinfo{author}{Zaffaroni, P.}, \bibinfo{author}{Brivio, P.},
  \bibinfo{year}{2009}.
\newblock \bibinfo{title}{Analysis and interpretation of spectral indices for
  soft multicriteria burned-area mapping in mediterranean regions}.
\newblock \bibinfo{journal}{{IEEE} Geoscience and Remote Sensing Letters}
  \bibinfo{volume}{6}, \bibinfo{pages}{499--503}.
\newblock \URLprefix \url{https://doi.org/10.1109%2Flgrs.2009.2020067},
  \DOIprefix\doi{10.1109/lgrs.2009.2020067}.
\bibitem[{Tansey(2004)}]{Tansey_2004}
\bibinfo{author}{Tansey, K.}, \bibinfo{year}{2004}.
\newblock \bibinfo{title}{Vegetation burning in the year 2000: Global burned
  area estimates from {SPOT} {VEGETATION} data}.
\newblock \bibinfo{journal}{Journal of Geophysical Research}
  \bibinfo{volume}{109}.
\newblock \URLprefix \url{https://doi.org/10.1029%2F2003jd003598},
  \DOIprefix\doi{10.1029/2003jd003598}.
\bibitem[{Tansey et~al.(2008)Tansey, Gr{\'{e}}goire, Defourny, Leigh, Pekel,
  van Bogaert and Bartholom{\'{e}}}]{Tansey_2008}
\bibinfo{author}{Tansey, K.}, \bibinfo{author}{Gr{\'{e}}goire, J.M.},
  \bibinfo{author}{Defourny, P.}, \bibinfo{author}{Leigh, R.},
  \bibinfo{author}{Pekel, J.F.}, \bibinfo{author}{van Bogaert, E.},
  \bibinfo{author}{Bartholom{\'{e}}, E.}, \bibinfo{year}{2008}.
\newblock \bibinfo{title}{A new, global, multi-annual (2000{\textendash}2007)
  burnt area product at 1 km resolution}.
\newblock \bibinfo{journal}{Geophysical Research Letters} \bibinfo{volume}{35}.
\newblock \URLprefix \url{https://doi.org/10.1029%2F2007gl031567},
  \DOIprefix\doi{10.1029/2007gl031567}.
\bibitem[{Vanderhoof et~al.(2017)Vanderhoof, Fairaux, Beal and
  Hawbaker}]{Vanderhoof_2017}
\bibinfo{author}{Vanderhoof, M.K.}, \bibinfo{author}{Fairaux, N.},
  \bibinfo{author}{Beal, Y.J.G.}, \bibinfo{author}{Hawbaker, T.J.},
  \bibinfo{year}{2017}.
\newblock \bibinfo{title}{Validation of the {USGS} landsat burned area
  essential climate variable ({BAECV}) across the conterminous united states}.
\newblock \bibinfo{journal}{Remote Sensing of Environment}
  \bibinfo{volume}{198}, \bibinfo{pages}{393--406}.
\newblock \URLprefix \url{https://doi.org/10.1016%2Fj.rse.2017.06.025},
  \DOIprefix\doi{10.1016/j.rse.2017.06.025}.
\bibitem[{Veraverbeke et~al.(2012)Veraverbeke, Gitas, Katagis, Polychronaki,
  Somers and Goossens}]{Veraverbeke_2012}
\bibinfo{author}{Veraverbeke, S.}, \bibinfo{author}{Gitas, I.},
  \bibinfo{author}{Katagis, T.}, \bibinfo{author}{Polychronaki, A.},
  \bibinfo{author}{Somers, B.}, \bibinfo{author}{Goossens, R.},
  \bibinfo{year}{2012}.
\newblock \bibinfo{title}{Assessing post-fire vegetation recovery using
  red–near infrared vegetation indices: Accounting for background and
  vegetation variability}.
\newblock \bibinfo{journal}{{ISPRS} Journal of Photogrammetry and Remote
  Sensing} \bibinfo{volume}{68}, \bibinfo{pages}{191}.
\newblock \URLprefix \url{https://doi.org/10.1016%2Fj.isprsjprs.2012.03.003},
  \DOIprefix\doi{10.1016/j.isprsjprs.2012.03.003}.
\bibitem[{Vermote et~al.(2016)Vermote, Justice, Claverie and
  Franch}]{Vermote_2016}
\bibinfo{author}{Vermote, E.}, \bibinfo{author}{Justice, C.},
  \bibinfo{author}{Claverie, M.}, \bibinfo{author}{Franch, B.},
  \bibinfo{year}{2016}.
\newblock \bibinfo{title}{Preliminary analysis of the performance of the
  landsat 8/{OLI} land surface reflectance product}.
\newblock \bibinfo{journal}{Remote Sensing of Environment}
  \bibinfo{volume}{185}, \bibinfo{pages}{46--56}.
\newblock \URLprefix \url{https://doi.org/10.1016%2Fj.rse.2016.04.008},
  \DOIprefix\doi{10.1016/j.rse.2016.04.008}.
\bibitem[{Wilson and Sader(2002)}]{Wilson_2002}
\bibinfo{author}{Wilson, E.H.}, \bibinfo{author}{Sader, S.A.},
  \bibinfo{year}{2002}.
\newblock \bibinfo{title}{Detection of forest harvest type using multiple dates
  of landsat {TM} imagery}.
\newblock \bibinfo{journal}{Remote Sensing of Environment}
  \bibinfo{volume}{80}, \bibinfo{pages}{385--396}.
\newblock \URLprefix \url{https://doi.org/10.1016%2Fs0034-4257%2801%2900318-2},
  \DOIprefix\doi{10.1016/s0034-4257(01)00318-2}.
\bibitem[{Zhu and Woodcock(2014)}]{Zhu_2014}
\bibinfo{author}{Zhu, Z.}, \bibinfo{author}{Woodcock, C.E.},
  \bibinfo{year}{2014}.
\newblock \bibinfo{title}{Automated cloud, cloud shadow, and snow detection in
  multitemporal landsat data: An algorithm designed specifically for monitoring
  land cover change}.
\newblock \bibinfo{journal}{Remote Sensing of Environment}
  \bibinfo{volume}{152}, \bibinfo{pages}{217--234}.
\newblock \URLprefix \url{https://doi.org/10.1016%2Fj.rse.2014.06.012},
  \DOIprefix\doi{10.1016/j.rse.2014.06.012}.

\end{thebibliography}

\end{document}